\RequirePackage[T1]{fontenc}
\documentclass[10pt,journal,compsoc]{IEEEtran}
\usepackage[utf8]{inputenc}
\usepackage[T1]{fontenc}
\usepackage{microtype}
\usepackage{graphicx}
\usepackage{rotating}
\usepackage[caption=false,font=footnotesize,labelfont=sf,textfont=sf]{subfig}
\usepackage{booktabs,multirow,array}
\usepackage{xcolor,pifont}
\usepackage{amsmath,amssymb,amsfonts,mathtools,amsthm}
\usepackage{algorithm,algorithmic}
\usepackage[nocompress]{cite}
\newcommand{\citep}{\cite}

\usepackage[hidelinks]{hyperref}
\usepackage[capitalize,noabbrev]{cleveref}
\newcommand{\cmark}{\ding{51}}
\newcommand{\xmark}{\ding{55}}

\newcommand{\safe}[1]{\textcolor{green!60!black}{#1}}
\newcommand{\best}[1]{\textcolor{green!60!black}{\textbf{#1}}}
\theoremstyle{plain}
\newtheorem{theorem}{Theorem}[section]
\newtheorem{proposition}[theorem]{Proposition}
\newtheorem{lemma}[theorem]{Lemma}
\newtheorem{corollary}[theorem]{Corollary}
\theoremstyle{definition}
\newtheorem{definition}[theorem]{Definition}

\theoremstyle{remark}

\allowdisplaybreaks[1]

\hypersetup{pdftitle={Structural Dual Adaptation for Safe Reinforcement Learning},pdfauthor={Runsheng Yu, Zhuoyang Song, Zengde Deng, Qixin Zhang, Li Shen, Jiaxing Zhang}}

\title{Structural Dual Adaptation for Safe Reinforcement Learning}
\author{Runsheng~Yu, Zhuoyang~Song, Zengde~Deng, Qixin~Zhang, Li~Shen, and Jiaxing~Zhang\thanks{R. Yu, Z. Song, and J. Zhang are with China Merchants Research Institute of Advanced Technology, China Merchants Group, Hong Kong, China (e-mail: runshengyu@gmail.com; \{songzhuoyang, zhangjiaxing\}@cmhk.com).}\thanks{Z. Deng is with The Chinese University of Hong Kong, Hong Kong, China (e-mail: dengzengde@gmail.com).}\thanks{Q. Zhang is with Generative AI Lab, College of Computing and Data Science, Nanyang Technological University, Singapore (e-mail: qixin.zhang@ntu.edu.sg).}\thanks{L. Shen is with the School of Cyber Science and Technology, Shenzhen Campus of Sun Yat-sen University, Shenzhen, China (e-mail: shenli6@mail.sysu.edu.cn).}}
\begin{document}
\IEEEtitleabstractindextext{
\begin{abstract}
Primal--dual methods enforce safety constraints in reinforcement learning through Lagrangian relaxation. For a fixed policy, the exact bounded multiplier response exhibits bang-bang switching at the constraint boundary. Practical gradient updates can oscillate through coupling with policy optimization. We propose Structural DuaL Adaptation (SDLA), which replaces this bang-bang response with a unique continuous response through current-cost centering, trajectory weighting, and quadratic regularization. We characterize activation thresholds and response sensitivity. Ternary search computes multipliers without a dual gradient stepsize. For entropy-regularized tabular policies, we bound equilibrium constraint violations alongside reward gaps. We establish convergence of delayed updates under mild assumptions. Neural PPO/SAC experiments on OmniSafe and HASARD demonstrate competitive rewards with improved constraint tracking. On HASARD, PPO-SDLA achieves higher reported mean reward than the human baseline in six of eighteen settings while meeting the respective mean-cost budgets, including four with lower mean cost. To the best of our knowledge, this is the first report of higher mean reward at lower mean cost than the human baseline across multiple HASARD settings.
\end{abstract}
\begin{IEEEkeywords}
Safe reinforcement learning, constrained Markov decision processes, primal--dual optimization.
\end{IEEEkeywords}}
\maketitle
\IEEEdisplaynontitleabstractindextext

\section{Introduction}
Safe reinforcement learning seeks high reward while respecting cost constraints~\citep{garcia2015comprehensive,gu2024review}. Lagrangian methods introduce a multiplier $\lambda$ that controls the cost penalty during policy improvement~\citep{ray2019benchmarking,chen2024adaptive,paternain2022safe}. For a fixed policy with mean cost $c$, the multiplier objective is $\lambda(d-c)$. Its exact minimizer on $[0,\lambda_{\max}]$ switches from zero to $\lambda_{\max}$ as $c$ crosses the budget $d$. This \emph{fixed-policy bang-bang response} arises even with exact cost estimates.

Quadratic regularization makes the fixed-policy response continuous. Its penalty stays zero for $c\le d$ and activates after a mean-cost violation. This \emph{violation-triggered dual activation} can lag policy changes when cost estimates are delayed. We seek smooth feedback that can activate within the feasible region based on current trajectories. Incremental gradient updates can oscillate through policy coupling. PID uses tuned gains to combine damping and cost-trend feedback~\citep{stooke2020responsivesafe,spoor2025empirical}. Entropy regularization can smooth the dual function when the policy is reoptimized for each multiplier~\citep{ying2022dual,li2024faster}. With the policy held fixed, the multiplier objective remains linear, so its exact minimizer exhibits bang-bang switching. Table~\ref{tab:comparison} compares these methods.

\begin{table}[!b]
\centering
\caption{Single-constraint multiplier responses with policy and return statistics fixed. Inner solve counts scalar optimization steps to multiplier accuracy $\varepsilon_\lambda$. Linear objectives admit an exact closed-form solution~\eqref{eq:fixed-policy-bang-bang}; SDLA uses ternary search. Bang-bang refers to the exact response. For PID, ``No closed form'' denotes feedback updates without a closed-form optimal response; --- denotes not applicable. Entropy-Regularized refers to policy entropy.}
\label{tab:comparison}
\scriptsize
\begin{tabular}{lcc}
\toprule
\textbf{Method} & \textbf{Inner Solve} & \textbf{Avoids Bang-Bang} \\
\midrule
Lagrangian & $O(1)$ (closed form) & \xmark \\
Lagrangian-PID & No closed form & --- \\
Entropy-Regularized & $O(1)$ (closed form) & \xmark \\
\textbf{SDLA (ours)} & $O(\log(\lambda_{\max}/\varepsilon_\lambda))$ & \cmark \\
\bottomrule
\end{tabular}
\end{table}

We propose \emph{Structural DuaL Adaptation (SDLA)} to replace fixed-policy bang-bang switching with structured smooth feedback. Its trajectory-dependent surrogate combines log-sum-exp weighting of penalized returns with current-cost centering. Quadratic regularization and trajectory-weighted cost variance control response sensitivity, yielding a unique continuous multiplier response. Ternary search computes this response without a dual gradient stepsize.

Our main contributions are:

\textbf{Structured multiplier response.}
We construct a strongly convex surrogate whose unique multiplier response replaces fixed-policy bang-bang switching with a continuous transition. Current-cost centering and trajectory weighting determine its activation and saturation regimes. A scalar inner solve computes the response without a dual gradient stepsize.

\textbf{Sensitivity characterization.}
We bound response sensitivity, including clipped endpoints, and identify the interior gain through regularization and trajectory-weighted cost variance. For entropy-regularized tabular policies, we bound fixed-parameter equilibrium feasibility and reward gaps and establish convergence of delayed updates with actor and dual errors under mild assumptions.

\textbf{Practical actor--critic evaluation.} On HASARD, PPO-SDLA achieves higher reported mean reward than the human baseline in 6 of 18 task--difficulty settings, including four with lower mean cost. All six settings meet their respective mean-cost budgets. To the best of our knowledge, this is the first report of simultaneously higher mean reward and lower mean cost than the human baseline across multiple HASARD settings. Ablations examine the surrogate components, while training-time diagnostics quantify multiplier changes and constraint tracking.

\section{Preliminary}
\label{sec:preliminary}
\subsection{Constrained Markov Decision Process}

We consider a single-constraint discounted CMDP $(\mathcal S,\mathcal A,P,r,c,\gamma,d)$~\citep{altman2021constrained}, with $0\le r\le R_{\max}$, $0\le c\le C_{\max}$, $\gamma\in[0,1)$, and a fixed initial distribution. Let $R=\sum_{t\ge0}\gamma^t r(s_t,a_t)$ and $C=\sum_{t\ge0}\gamma^t c(s_t,a_t)$ denote trajectory returns, and write $J^r(\pi)=\mathbb E_\pi R$, $J^c(\pi)=\mathbb E_\pi C$. The objective is
\begin{equation}
\max_\pi J^r(\pi)\qquad\text{s.t.}\qquad J^c(\pi)\le d.
\end{equation}
Returns satisfy $R\le R_{\max}/(1-\gamma)$ and $C\le\bar C:=C_{\max}/(1-\gamma)$. Discounted policy entropy is $H(\pi)=\mathbb E_\pi\sum_{t\ge0}\gamma^t[-\log\pi(a_t\mid s_t)]$.

\subsection{Related Methods}
Gradient primal--dual updates accumulate cost errors, and coupling with policy optimization can cause oscillation and stepsize sensitivity~\citep{lin2020gradient,stooke2020responsivesafe}. PID control adds proportional damping and derivative feedback based on cost trends~\citep{stooke2020responsivesafe}. Augmented penalties~\citep{dai2023augmented}, critic calibration~\citep{wu2024off}, and regularized primal--dual methods~\citep{ding2023last,montenegro2024last} provide other stabilization mechanisms. SDLA derives its response by optimizing a scalar surrogate on a frozen trajectory batch, with explicit activation and sensitivity bounds.

Penalty methods require selecting a penalty strength~\citep{zhang2022penalized}. Constrained or projected methods introduce additional actor optimization steps~\citep{achiam2017constrained,yang2020projection,zhang2020first,yang2022constrained}, while state augmentation requires budget-conditioned policy inputs~\citep{sootla2022saute,sootla2022enhancing}. For constrained multi-objective RL, Gu et al.~\citep{gu2025safe} combine gradient manipulation across reward objectives with cost-reducing policy updates when constraints are violated. Appendix~\ref{app:related-methods} details these tradeoffs. SDLA changes the multiplier response within an actor--critic implementation.

\subsection{Policy Dual Smoothness and Fixed-Policy Responses}
Entropy regularization improves the optimization landscape of CMDPs by smoothing policy optimization. With entropy coefficient $\alpha>0$, the entropy-regularized problem is
\begin{equation}
\max_\pi\{J^r(\pi)+\alpha H(\pi)\}
\quad\text{s.t.}\quad J^c(\pi)\le d.
\end{equation}
\textbf{Lagrangian dual.} The corresponding policy dual is \citep{ying2022dual}
\begin{equation}
g_{\mathrm{std}}(\lambda)
:=\sup_\pi\{J^r(\pi)+\alpha H(\pi)-\lambda(J^c(\pi)-d)\}.
\label{eq:dual_ying}
\end{equation}
In entropy-regularized tabular formulations, policy-dual smoothness enables accelerated optimization \citep{ying2022dual,li2024faster}. These guarantees involve solving, or approximating, the underlying policy optimization problem.

\noindent\textbf{Fixed-policy dual response.}
For a frozen policy $\pi_{\mathrm{cur}}$, both $J^r(\pi_{\mathrm{cur}})$ and $\alpha H(\pi_{\mathrm{cur}})$ are constant with respect to $\lambda$. Writing $c=J^c(\pi_{\mathrm{cur}})$, the exact multiplier response is
\begin{equation}
\operatorname*{arg\,min}_{0\le\lambda\le\lambda_{\max}}\lambda(d-c)
=
\begin{cases}
\{0\}, & c<d,\\
[0,\lambda_{\max}], & c=d,\\
\{\lambda_{\max}\}, & c>d.
\end{cases}
\label{eq:fixed-policy-bang-bang}
\end{equation}
This bang-bang response concerns the fixed-policy subproblem. The policy dual in~\eqref{eq:dual_ying} reoptimizes the policy as $\lambda$ changes. Quadratic regularization smooths the fixed-policy response and retains the activation boundary $c=d$: the multiplier remains zero for $c\le d$ and activates after a mean-cost violation. SDLA builds on this curvature to construct structured smooth feedback from current mean and trajectory-weighted costs. Its response has controlled sensitivity and can also activate within the feasible region.

\section{Structural Dual Adaptation}
\subsection{Surrogate Dual Formulation}
\label{sec:surrogate_duality}
We first define SDLA through expectations under a fixed policy $\pi$. Write $c_\pi=J^c(\pi)$ and fix $\tau,\beta,\lambda_{\max}>0$. Returns satisfy the bounds in Section~\ref{sec:preliminary}.
\begin{definition}[Population Surrogate Dual]
\label{def:surrogate-dual}
The population surrogate dual is
\begin{equation}
g_\pi(\lambda)=\tau\log\mathbb E_\pi e^{(R-\lambda C)/\tau}
+\lambda(d-c_\pi)+\frac{\lambda^2}{2\beta},
\label{eq:release-population}
\end{equation}
for $\lambda\in[0,\lambda_{\max}]$. Its multiplier response is
\begin{equation}
\Lambda(\pi)=\operatorname*{arg\,min}_{0\le\lambda\le\lambda_{\max}}g_\pi(\lambda).
\label{eq:release-response}
\end{equation}
\end{definition}
Write $F_\pi(\lambda)=\tau\log\mathbb E_\pi e^{(R-\lambda C)/\tau}$. Its derivative supplies the trajectory-weighted cost
\begin{equation}
m_\pi(\lambda):=-F_\pi'(\lambda)
=\frac{\mathbb E_\pi[C e^{(R-\lambda C)/\tau}]}
{\mathbb E_\pi e^{(R-\lambda C)/\tau}}.
\label{eq:population-weighted-cost}
\end{equation}
Trajectories with higher penalized returns $R-\lambda C$ receive greater weight. Centering by $\lambda(d-c_\pi)$ combines mean and trajectory-weighted costs into the signal $c_\pi+m_\pi(\lambda)-d$. Quadratic regularization and trajectory-weighted cost variance control the response curvature.

\subsection{Structural Properties and Activation}
\label{sec:structural-properties}
\begin{lemma}[Curvature and continuous activation]
\label{lem:smooth_convex}
For any fixed policy with $0\le C\le\bar C$, the population surrogate satisfies
\begin{align}
g_\pi'(\lambda)&=d-c_\pi-m_\pi(\lambda)+\lambda/\beta,
\label{eq:derivs}\\
g_\pi''(\lambda)&=\frac{\operatorname{Var}_{q_{\pi,\lambda}}(C)}{\tau}+\frac1\beta.
\label{eq:curvature}
\end{align}
Here $q_{\pi,\lambda}$ is the trajectory law reweighted by $e^{(R-\lambda C)/\tau}$. Hence $g_\pi$ is $1/\beta$-strongly convex and $(1/\beta+\bar C^2/(4\tau))$-smooth. Its response $\Lambda(\pi)$ is unique and satisfies
\begin{equation}
\Lambda(\pi)>0\quad\Longleftrightarrow\quad c_\pi+m_\pi(0)>d.
\label{eq:activate}
\end{equation}
Moreover,
$0\le\Lambda(\pi)\le\min\{\lambda_{\max},\beta[c_\pi+m_\pi(0)-d]_+\}$.
\end{lemma}

\textbf{Feasible-region activation.}
Condition~\eqref{eq:activate} permits penalties before the current policy's mean cost exceeds the budget. Proposition~\ref{prop:fixed-parameter-bounds} quantifies equilibrium constraint violations and regularized reward gaps for fixed parameters.

The endpoint derivatives distinguish three response regimes:
\begin{enumerate}
\item[(i)] \textbf{Inactive}: $g_\pi'(0)\ge0$ gives $\Lambda(\pi)=0$.
\item[(ii)] \textbf{Transitional}: $g_\pi'(0)<0<g_\pi'(\lambda_{\max})$ gives $0<\Lambda(\pi)<\lambda_{\max}$ and
$c_\pi+m_\pi(\Lambda(\pi))=d+\Lambda(\pi)/\beta$.
\item[(iii)] \textbf{Saturated}: $g_\pi'(\lambda_{\max})\le0$ gives $\Lambda(\pi)=\lambda_{\max}$.
\end{enumerate}
The lemma implies $\Lambda(\pi)\to0$ as $[c_\pi+m_\pi(0)-d]_+\to0$. Section~\ref{sec:penalty-release} quantifies sensitivity across these regimes; Appendix~\ref{sec:proofs} gives the proofs. Section~\ref{sec:ablations} compares empirical variants with fixed and adaptive centering.

\subsection{Closed-Loop Multiplier Feedback}
\label{sec:penalty-release}
For consecutive policies $\pi_t=\pi_{\theta_t}$, write $g_t=g_{\pi_t}$, $m_t=m_{\pi_t}$, $c_t=J^c(\pi_t)$, and $\ell_t=\Lambda(\pi_t)$. For the actor interpretation, set $r(\theta)=J^r(\pi_\theta)$ and $c(\theta)=J^c(\pi_\theta)$. Keeping $\tau,\beta,\lambda_{\max}$ fixed isolates the response to policy changes.

The following result quantifies how the multiplier increases or decreases as the mean and weighted costs change. Define
\begin{equation}
\mathcal D_t:=(c_{t+1}-c_t)+m_{t+1}(\ell_t)-m_t(\ell_t),
\qquad \delta_t:=\ell_{t+1}-\ell_t.
\label{eq:release-signal}
\end{equation}
\begin{theorem}[Closed-loop multiplier feedback]
\label{thm:penalty-release}
For any two consecutive policies, including responses at either endpoint,
\begin{equation}
\frac{\delta_t^2}{\beta}\le\delta_t\mathcal D_t,
\qquad |\delta_t|\le\beta|\mathcal D_t|.
\label{eq:release-feedback-bound}
\end{equation}
Thus $\mathcal D_t<0$ gives $\ell_{t+1}\le\ell_t$, and $\mathcal D_t>0$ gives $\ell_{t+1}\ge\ell_t$. Both implications are strict when $\ell_t\in(0,\lambda_{\max})$. If both responses are interior, then
\begin{equation}
\ell_{t+1}-\ell_t=\frac{\mathcal D_t}{\beta^{-1}+\overline V_{t+1}},
\label{eq:release-law}
\end{equation}
where $\overline V_{t+1}\ge0$ is the average trajectory-weighted cost variance divided by $\tau$ along the segment from $\ell_t$ to $\ell_{t+1}$.
\end{theorem}
\textit{Proof sketch.} Combining the constrained first-order conditions with strong convexity gives~\eqref{eq:release-feedback-bound}, including clipped endpoints. For interior responses, integrating the curvature between the two stationary points gives~\eqref{eq:release-law}. Appendix~\ref{app:penalty-release} provides the full proof.

Both weighted costs in $\mathcal D_t$ are evaluated at the old multiplier $\ell_t$, so their difference measures the change in trajectory information under a common penalty. A larger multiplier places greater weight on cost reduction in the actor update. The resulting policy supplies the next trajectory batch and dual signal, completing the feedback loop.

Equation~\eqref{eq:release-feedback-bound} bounds the multiplier change in proportion to the combined cost-signal change. In the interior, the response gain is $\beta/(1+\beta\overline V_{t+1})$. Increasing the multiplier downweights higher-cost trajectories, adding the trajectory-weighted cost variance term to the curvature of the quadratic regularizer. For a given feedback signal, this reduces the multiplier change. Clipping can keep the multiplier fixed under a nonzero signal, as captured by~\eqref{eq:release-feedback-bound}.

\section{Theoretical Guarantees}
\label{sec:theory}
\subsection{Convergence}
\label{sec:concrete-actor}
We consider a finite discounted CMDP with tabular policies and $0<\gamma<1$. Let $\Pi=\prod_{s\in\mathcal S}\Delta(\mathcal A)$ be the tabular policy space, with Frobenius norm $\|\cdot\|_F$, and assume the initial distribution satisfies $\rho(s)>0$ for every state. Fix a policy-entropy coefficient $\alpha>0$ and write
\begin{equation}
\begin{aligned}
J_\alpha(\pi)&=J^r(\pi)+\alpha H(\pi),\\
\mathcal L_\alpha(\pi,\lambda)&=J_\alpha(\pi)-\lambda[J^c(\pi)-d],\\
B(\lambda)&=\operatorname*{arg\,max}_{\pi\in\Pi}\mathcal L_\alpha(\pi,\lambda).
\end{aligned}
\label{eq:entropy-actor-response}
\end{equation}
The policy-entropy coefficient $\alpha$ is distinct from the trajectory temperature $\tau$. Write $\Lambda_{\beta,\tau}(\pi)$ and $m_{\pi,\tau}(\lambda)$ for the response and trajectory-weighted cost in Section~\ref{sec:surrogate_duality}, making parameter dependence explicit. Let $K_c$ and $K_m(\tau)$ be Lipschitz constants for $J^c(\pi)$ and $m_{\pi,\tau}(\lambda)$ in Frobenius norm, with the latter uniform over $\lambda\in[0,\lambda_{\max}]$. Set $\kappa=\alpha\min_s\rho(s)>0$ and $L_\Lambda=\beta[K_c+K_m(\tau)]$. Appendix~\ref{app:parameter-convergence} derives these constants and the actor response bound from the soft-policy optimality identity~\citep{ying2022dual}.

Take a full-support initial policy and $\lambda_0\in[0,\lambda_{\max}]$. At outer step $t$, hold $\lambda_t$ fixed and run $K_t\ge1$ actor updates from $\pi_t^0=\pi_t$:
\[
\pi_t^{k+1}(a\mid s)\propto
[\pi_t^k(a\mid s)]^{1-\omega}
\exp(\omega\widehat Q_t^k(s,a)/\alpha).
\]
For $0<\omega\le1$, assume $\|\widehat Q_t^k-Q_{\lambda_t}^{\pi_t^k}\|_\infty\le\xi_t\le\bar\xi<\infty$ and set $\pi_{t+1}=\pi_t^{K_t}$. The cases $\omega=1$ and $0<\omega<1$ correspond to tabular soft policy improvement and an entropy-regularized KL-PPO update, respectively~\citep{haarnoja2018soft,liu2019neural}. The analysis establishes tabular convergence under mild assumptions specified below. Section~\ref{sec:exp} evaluates neural PPO-Clip and SAC implementations empirically.

Write $b=K_c/\kappa$, $L_\Lambda=L_\Lambda(\beta,\tau)$, and $\chi=1-(1-\gamma)\omega$. Appendix~\ref{app:concrete-actor} specifies the updates and derives uniform constants $A_0,A_1$ from Cen et al.~\citep{cen2022fast}, with actor error at most $A_0\chi^{K_t-1}+A_1\xi_t$. At each outer iteration, let $\lambda_{t+1}\in[0,\lambda_{\max}]$ satisfy
\begin{equation}
|\lambda_{t+1}-\Lambda_{\beta,\tau}(\pi_t)|\le\zeta_t,\qquad\zeta_t\ge0.
\label{eq:inexact-dual-update}
\end{equation}

\begin{theorem}[Convergence of tabular SDLA updates]
\label{thm:concrete-actor-convergence}
Under the assumptions and updates above, suppose $q=bL_\Lambda<1$. Then the equilibrium satisfying $\bar\pi=B(\bar\lambda)$ and $\bar\lambda=\Lambda_{\beta,\tau}(\bar\pi)$ is unique. Write $e_t=\|\pi_t-\bar\pi\|_F$.
If $K_t\to\infty$, $\xi_t\to0$, and $\zeta_t\to0$, then $(\pi_t,\lambda_t)\to(\bar\pi,\bar\lambda)$ from every full-support initial policy. If $K_t\ge K$, $\xi_t\le\xi$, and $\zeta_t\le\zeta$, then
\begin{equation}
\begin{aligned}
\limsup_{t\to\infty}e_t&\le\mathcal E_{K,\xi,\zeta}
:=\frac{A_0\chi^{K-1}+A_1\xi+b\zeta}{1-q},\\
\limsup_{t\to\infty}|\lambda_t-\bar\lambda|
&\le\zeta+L_\Lambda\mathcal E_{K,\xi,\zeta}.
\end{aligned}
\label{eq:concrete-error-floor}
\end{equation}
If $\chi^{K_t-1}+\xi_t+\zeta_t=O(r^t)$ for $r\in(0,1)$, then
\begin{equation}
\|\pi_t-\bar\pi\|_F+|\lambda_t-\bar\lambda|=O(\nu^t),
\qquad \max\{\sqrt{q},r\}<\nu<1.
\label{eq:concrete-linear-rate}
\end{equation}
\end{theorem}
The proof uses a two-step error recursion (Supplementary Eq.~\eqref{eq:concrete-delayed-recursion}), yielding an outer-iteration contraction factor of $\sqrt q$. Appendix~\ref{app:concrete-actor} combines the actor bound of Cen et al.~\citep{cen2022fast} with the actor and dual sensitivities from Appendix~\ref{app:parameter-convergence}.

The condition $q=bL_\Lambda<1$ bounds the coupled response gains, following small-gain principles used in mean-field learning~\citep{guo2019learning,xie2021learning}, performative RL~\citep{mandal2023performative}, and actor--critic analysis~\citep{olshevsky2023small}. Since $L_\Lambda=\beta(K_c+K_m)$, stronger quadratic regularization reduces this gain; Appendix~\ref{app:parameter-convergence} gives a nonempty sufficient parameter range.

\subsection{Feasibility and Reward Bounds}
\label{sec:feasibility-bounds}
\begin{proposition}[Fixed-parameter feasibility and reward bounds]
\label{prop:fixed-parameter-bounds}
In the tabular setting above, suppose $J^c(\pi_s)\le d-\sigma$ for some $\sigma>0$. Let $U\ge\sup_\pi J_\alpha(\pi)$, $\Delta=U-J_\alpha(\pi_s)$, and $J_\alpha^\star(d)=\max_{J^c(\pi)\le d}J_\alpha(\pi)$. Fix $\beta,\tau,L=\lambda_{\max}>0$. Any exact equilibrium $\bar\pi=B(\bar\lambda)$, $\bar\lambda=\Lambda_{\beta,\tau}(\bar\pi)$ satisfies
\begin{align}
&[J^c(\bar\pi)-d]_+\le V_{\beta,L},\label{eq:fixed-feasibility-bound}\\
&V_{\beta,L}:=\max\left\{
\frac{\sqrt{\sigma^2+4\Delta/\beta}-\sigma}{2},
\left[\frac{\Delta}{L}-\sigma\right]_+\right\},\nonumber\\
&[J_\alpha^\star(d)-J_\alpha(\bar\pi)]_+
\le\frac{\beta\bar C^2}{4}.
\label{eq:fixed-reward-bound}
\end{align}
If $L\ge\Delta/\sigma$, then $V_{\beta,L}\le\Delta/(\beta\sigma)$.
\end{proposition}
Actor optimality gives $\bar\lambda[J^c(\bar\pi)-d+\sigma]\le\Delta$; combining this with the surrogate endpoint conditions proves the bounds. Appendix~\ref{app:fixed-parameter-bounds} gives the proof and, under Theorem~\ref{thm:concrete-actor-convergence}, the iterative feasibility bound $V_{\beta,L}+K_c\mathcal E_{K,\xi,\zeta}$. The equilibrium bounds hold independently of the convergence condition $q<1$.

\subsection{Optimality}
\label{sec:optimality}
\begin{corollary}[CMDP optimality under parameter matching]
\label{cor:concrete-actor-feasibility}
Under Theorem~\ref{thm:concrete-actor-convergence}, suppose $\mathcal L_\alpha$ has a saddle point $(\pi_\alpha^\star,\lambda_\alpha^\star)$ with $J^c(\pi_\alpha^\star)=d>0$ and $0<\lambda_\alpha^\star<\lambda_{\max}$. Choose $\beta=\beta^\star(\tau)$, where
\begin{equation}
\beta^\star(\tau)=\frac{\lambda_\alpha^\star}{m_{\pi_\alpha^\star,\tau}(\lambda_\alpha^\star)}>0.
\label{eq:optimality-parameter-curve}
\end{equation}
Require $q<1$ at this parameter choice. Then $(\bar\pi,\bar\lambda)=(\pi_\alpha^\star,\lambda_\alpha^\star)$ and $[J^c(\pi_t)-d]_+\le K_c e_t$. As $K_t\to\infty$ and $\xi_t,\zeta_t\to0$, $J_\alpha(\pi_t)\to J_\alpha(\pi_\alpha^\star)$ and constraint violations vanish. Uniform errors give $\limsup_t[J^c(\pi_t)-d]_+\le K_c\mathcal E_{K,\xi,\zeta}$.
\end{corollary}
Corollary~\ref{cor:concrete-actor-feasibility} establishes exact optimality through parameter matching. Parameter conditions expressed through optimal multipliers also occur in constrained optimization, including exact penalty theory~\citep[Exercise~5.16]{boyd2004convex}~\citep{antczak2013lower}. Here $\beta^\star(\tau)$ provides the theoretical characterization, and ablations evaluate practical parameter choices.

\section{Practical Implementation}
\label{sec:implementation}
\subsection{Empirical Surrogate Dual}
\label{sec:empirical-surrogate}
Following standard sample-based practice in safe RL~\citep{achiam2017constrained}, we collect $M$ trajectories under $\pi_{\mathrm{cur}}$ and compute discounted returns $\{(R_i,C_i)\}_{i=1}^M$. Replacing population expectations with sample averages gives
\begin{equation}
\begin{aligned}
\widehat F_M(\lambda)&=\tau\log\left[\frac1M\sum_{i=1}^M e^{(R_i-\lambda C_i)/\tau}\right],\\
\widehat c_M&=\frac1M\sum_{i=1}^M C_i,
\end{aligned}
\label{eq:trajectory-objective}
\end{equation}
and the empirical surrogate
\begin{equation}
\widehat g_M(\lambda)=\widehat F_M(\lambda)+\lambda(d-\widehat c_M)+\frac{\lambda^2}{2\beta}.
\label{eq:surrogate-dual}
\end{equation}
Its trajectory-weighted cost is
\begin{equation}
\widehat m(\lambda)=-\widehat F_M'(\lambda)
=\frac{\sum_i C_i e^{(R_i-\lambda C_i)/\tau}}{\sum_i e^{(R_i-\lambda C_i)/\tau}}.
\label{eq:weighted-cost}
\end{equation}
The batch stays fixed during the solve. The empirical surrogate is $1/\beta$-strongly convex, with a unique minimizer $\widehat\lambda^*$ on $[0,\lambda_{\max}]$. Its activation condition is $\widehat\lambda^*>0$ exactly when $\widehat c_M+\widehat m(0)>d$.

\subsection{Scalar Solver and PPO/SAC Integration}
\label{sec:scalar-integration}
Ternary search computes $\widehat\lambda^*$ to accuracy $\varepsilon_\lambda$ without a dual learning rate, using $O(M\log(\lambda_{\max}/\varepsilon_\lambda))$ arithmetic operations. We use 15 iterations and stabilized log-sum-exp evaluations; Appendix~\ref{app:scalar-solver} gives the interval bound. Trajectory-estimation and scalar-solver errors contribute to $\zeta_t$ in~\eqref{eq:inexact-dual-update} (Supplementary Eq.~\eqref{eq:sample-dual-error}).

SDLA combines a scalar dual solve with SAC~\citep{haarnoja2018soft} or PPO~\citep{schulman2017proximal}. Reward and cost critics estimate $Q^r_{\phi_r}$ and $Q^c_{\phi_c}$; the actor uses $Q^r-\lambda Q^c$. Each dual solve holds the trajectory batch fixed.

Related safe RL methods also combine structural analyses with neural implementations~\citep{chen2024adaptive,liu2022constrained,xu2021crpo}; Appendix~\ref{app:theory-practice} compares their theoretical settings.

PPO-SDLA combines rollout advantages as $\widehat A^r-\lambda\widehat A^c$ and uses the same batch for the dual solve. SAC-SDLA updates actors and critics from replay transitions; recent rollout or evaluation episodes supply trajectory returns and mean cost for the surrogate. Both hold $\lambda$ fixed during actor updates and use the subsequent dual solution in the next iteration.

Supplementary Algorithms~\ref{alg:general-convex}--\ref{alg:sdla-sac} give the general framework and its PPO/SAC implementations.

\begin{table*}[t]
\centering
\caption{Rewards and costs over five SDLA runs and available archived baseline runs. The cost threshold is 5 for Remedy Rush, Collateral Damage, and Detonator's Dilemma, and 50 for all other environments. \textcolor{green!60!black}{Green}: mean cost meets the budget.  \textcolor{green!60!black}{\textbf{Bold Green}}: highest mean reward among methods meeting the mean cost budget.}
\label{tab:baseline_comparison}
\footnotesize
\setlength{\tabcolsep}{4pt}
\begin{tabular}{@{}clcccccccccccc@{}}
\toprule
\multirow{3}{*}{Level} & \multirow{3}{*}{Method} & \multicolumn{2}{c}{Armament} & \multicolumn{2}{c}{Volcanic} & \multicolumn{2}{c}{Remedy} & \multicolumn{2}{c}{Collateral} & \multicolumn{2}{c}{Precipice} & \multicolumn{2}{c}{Detonator's} \\
& & \multicolumn{2}{c}{Burden} & \multicolumn{2}{c}{Venture} & \multicolumn{2}{c}{Rush} & \multicolumn{2}{c}{Damage} & \multicolumn{2}{c}{Plunge} & \multicolumn{2}{c}{Dilemma} \\
\cmidrule(lr){3-4} \cmidrule(lr){5-6} \cmidrule(lr){7-8} \cmidrule(lr){9-10} \cmidrule(lr){11-12} \cmidrule(lr){13-14}
& & $R \uparrow$ & $C \downarrow$ & $R \uparrow$ & $C \downarrow$ & $R \uparrow$ & $C \downarrow$ & $R \uparrow$ & $C \downarrow$ & $R \uparrow$ & $C \downarrow$ & $R \uparrow$ & $C \downarrow$ \\
\midrule
\multirow{7}{*}{1} 
& PPO & 9.68 & 109.30 & 51.64 & 172.93 & 50.78 & 52.44 & 78.61 & 41.06 & 243.42 & 475.62 & 29.67 & 14.28 \\
& PPOCost & \textcolor{green!60!black}{5.47} & \textcolor{green!60!black}{3.67} & \textcolor{green!60!black}{30.73} & \textcolor{green!60!black}{5.44} & 28.21 & 6.75 & 68.71 & 24.04 & \textcolor{green!60!black}{237.72} & \textcolor{green!60!black}{0.69} & 27.60 & 8.45 \\
& PPOLag & 7.51 & 52.41 & 42.40 & 52.00 & 36.37 & 5.25 & 29.09 & 5.61 & \textcolor{green!60!black}{147.24} & \textcolor{green!60!black}{44.96} & 21.49 & 5.62 \\
& PPOSaute & \textcolor{green!60!black}{2.33} & \textcolor{green!60!black}{32.87} & 32.68 & 62.00 & 20.50 & 9.18 & 26.08 & 7.09 & 241.37 & 424.35 & 28.69 & 8.78 \\
& PPOPID & \textcolor{green!60!black}{8.99} & \textcolor{green!60!black}{49.79} & 45.23 & 50.53 & \textcolor{green!60!black}{38.19} & \textcolor{green!60!black}{4.90} & 43.27 & 5.03 & \textcolor{green!60!black}{231.53} & \textcolor{green!60!black}{43.91} & 26.51 & 5.25 \\
& P3O & \textcolor{green!60!black}{8.60} & \textcolor{green!60!black}{40.72} & \textcolor{green!60!black}{43.55} & \textcolor{green!60!black}{46.10} & \textcolor{green!60!black}{37.90} & \textcolor{green!60!black}{4.78} & 46.52 & 5.97 & 242.53 & 176.41 & 29.62 & 6.59 \\
& PPO-SDLA (ours) & \textcolor{green!60!black}{\textbf{19.46}} & \textcolor{green!60!black}{12.80} & \textcolor{green!60!black}{\textbf{44.49}} & \textcolor{green!60!black}{29.49} & \textcolor{green!60!black}{\textbf{39.07}} & \textcolor{green!60!black}{2.71} & \textcolor{green!60!black}{\textbf{43.96}} & \textcolor{green!60!black}{3.01} & \textcolor{green!60!black}{\textbf{243.30}} & \textcolor{green!60!black}{43.09} & \textcolor{green!60!black}{\textbf{27.24}} & \textcolor{green!60!black}{2.97} \\
\midrule
\multirow{7}{*}{2}
& PPO & 4.24 & 99.59 & 38.31 & 186.44 & 61.94 & 62.22 & 53.58 & 61.13 & 324.69 & 608.07 & 40.09 & 19.63 \\
& PPOCost & \textcolor{green!60!black}{7.59} & \textcolor{green!60!black}{6.20} & \textcolor{green!60!black}{21.10} & \textcolor{green!60!black}{3.93} & \textcolor{green!60!black}{0.01} & \textcolor{green!60!black}{0.03} & 21.86 & 8.19 & 162.01 & 61.19 & 40.37 & 15.56 \\
& PPOLag & 4.50 & 53.50 & 26.11 & 53.10 & 28.75 & 5.72 & 18.87 & 5.61 & 105.01 & 52.39 & 19.57 & 5.34 \\
& PPOSaute & \textcolor{green!60!black}{1.55} & \textcolor{green!60!black}{30.47} & 23.68 & 75.39 & \textcolor{green!60!black}{3.72} & \textcolor{green!60!black}{4.80} & \textcolor{green!60!black}{6.37} & \textcolor{green!60!black}{3.21} & 119.42 & 159.30 & 20.79 & 9.97 \\
& PPOPID & 5.50 & 50.30 & 33.52 & 50.38 & 32.92 & 5.15 & 27.23 & 5.12 & 162.16 & 50.77 & \textcolor{green!60!black}{20.84} & \textcolor{green!60!black}{4.92} \\
& P3O & \textcolor{green!60!black}{5.33} & \textcolor{green!60!black}{39.82} & \textcolor{green!60!black}{32.38} & \textcolor{green!60!black}{45.89} & \textcolor{green!60!black}{31.91} & \textcolor{green!60!black}{4.85} & 27.41 & 5.12 & 247.05 & 178.56 & 25.48 & 6.97 \\
& PPO-SDLA (ours) & \textcolor{green!60!black}{\textbf{14.22}} & \textcolor{green!60!black}{19.01} & \textcolor{green!60!black}{\textbf{33.71}} & \textcolor{green!60!black}{28.01} & \textcolor{green!60!black}{\textbf{44.02}} & \textcolor{green!60!black}{4.67} & \textcolor{green!60!black}{\textbf{31.09}} & \textcolor{green!60!black}{4.44} & \textcolor{green!60!black}{\textbf{172.40}} & \textcolor{green!60!black}{40.98} & \textcolor{green!60!black}{\textbf{22.72}} & \textcolor{green!60!black}{4.19} \\
\midrule
\multirow{7}{*}{3}
& PPO & 1.99 & 118.22 & 42.20 & 347.77 & 53.16 & 68.27 & 34.86 & 84.44 & 487.17 & 894.22 & 49.36 & 23.72 \\
& PPOCost & \textcolor{green!60!black}{0.04} & \textcolor{green!60!black}{0.05} & \textcolor{green!60!black}{10.61} & \textcolor{green!60!black}{14.76} & \textcolor{green!60!black}{0.01} & \textcolor{green!60!black}{0.02} & \textcolor{green!60!black}{3.93} & \textcolor{green!60!black}{1.35} & \textcolor{green!60!black}{15.39} & \textcolor{green!60!black}{6.64} & 49.95 & 19.79 \\
& PPOLag & \textcolor{green!60!black}{2.03} & \textcolor{green!60!black}{31.89} & 22.77 & 52.54 & 8.02 & 9.74 & 12.22 & 5.29 & \textcolor{green!60!black}{23.87} & \textcolor{green!60!black}{34.98} & 19.27 & 5.26 \\
& PPOSaute & \textcolor{green!60!black}{2.32} & \textcolor{green!60!black}{37.68} & 18.89 & 246.80 & \textcolor{green!60!black}{1.17} & \textcolor{green!60!black}{3.50} & \textcolor{green!60!black}{2.74} & \textcolor{green!60!black}{2.76} & 101.64 & 176.14 & 22.20 & 10.36 \\
& PPOPID & \textcolor{green!60!black}{2.78} & \textcolor{green!60!black}{33.89} & \textcolor{green!60!black}{25.80} & \textcolor{green!60!black}{49.02} & 13.51 & 5.14 & \textcolor{green!60!black}{14.51} & \textcolor{green!60!black}{4.97} & \textcolor{green!60!black}{54.02} & \textcolor{green!60!black}{49.57} & \textcolor{green!60!black}{20.49} & \textcolor{green!60!black}{4.87} \\
& P3O & \textcolor{green!60!black}{2.61} & \textcolor{green!60!black}{29.94} & \textcolor{green!60!black}{24.93} & \textcolor{green!60!black}{49.84} & \textcolor{green!60!black}{14.29} & \textcolor{green!60!black}{4.19} & \textcolor{green!60!black}{13.57} & \textcolor{green!60!black}{4.12} & 269.14 & 428.72 & 26.73 & 7.49 \\
& PPO-SDLA (ours) & \textcolor{green!60!black}{\textbf{2.87}} & \textcolor{green!60!black}{2.85} & \textcolor{green!60!black}{\textbf{28.22}} & \textcolor{green!60!black}{44.25} & \textcolor{green!60!black}{\textbf{26.21}} & \textcolor{green!60!black}{4.48} & \textcolor{green!60!black}{\textbf{17.60}} & \textcolor{green!60!black}{3.52} & \textcolor{green!60!black}{\textbf{95.02}} & \textcolor{green!60!black}{36.69} & \textcolor{green!60!black}{\textbf{20.82}} & \textcolor{green!60!black}{3.87} \\
\bottomrule
\end{tabular}
\end{table*}

\begin{table*}[t]
\caption{Human player rewards and costs averaged over 10 episodes compared to PPO-SDLA terminal training summaries. \textbf{Bold}: PPO-SDLA achieves higher reported mean reward than human.}
\label{tab:human_comparison}
\centering
\footnotesize
\setlength{\tabcolsep}{4pt}
\begin{tabular}{@{}clcccccccccccc@{}}
\toprule
\multirow{3}{*}{Level} & \multirow{3}{*}{Method} & \multicolumn{2}{c}{Armament} & \multicolumn{2}{c}{Volcanic} & \multicolumn{2}{c}{Remedy} & \multicolumn{2}{c}{Collateral} & \multicolumn{2}{c}{Precipice} & \multicolumn{2}{c}{Detonator's} \\
& & \multicolumn{2}{c}{Burden} & \multicolumn{2}{c}{Venture} & \multicolumn{2}{c}{Rush} & \multicolumn{2}{c}{Damage} & \multicolumn{2}{c}{Plunge} & \multicolumn{2}{c}{Dilemma} \\
\cmidrule(lr){3-4} \cmidrule(lr){5-6} \cmidrule(lr){7-8} \cmidrule(lr){9-10} \cmidrule(lr){11-12} \cmidrule(lr){13-14}
& & $R \uparrow$ & $C \downarrow$ & $R \uparrow$ & $C \downarrow$ & $R \uparrow$ & $C \downarrow$ & $R \uparrow$ & $C \downarrow$ & $R \uparrow$ & $C \downarrow$ & $R \uparrow$ & $C \downarrow$ \\
\midrule
\multirow{2}{*}{1} 
& Human & 18.60 & 42.83 & 57.12 & 45.97 & 51.04 & 5.65 & 42.33 & 4.11 & 243.09 & 46.88 & 31.54 & 4.49 \\
& PPO-SDLA (ours) & \textbf{19.46} & 12.80 & 44.49 & 29.49 & 39.07 & 2.71 & \textbf{43.96} & 3.01 & \textbf{243.30} & 43.09 & 27.24 & 2.97 \\
\midrule
\multirow{2}{*}{2} 
& Human & 16.85 & 31.50 & 42.91 & 45.09 & 34.72 & 1.22 & 17.34 & 3.10 & 226.11 & 44.35 & 34.66 & 3.09 \\
& PPO-SDLA (ours) & 14.22 & 19.01 & 33.71 & 28.01 & \textbf{44.02} & 4.67 & \textbf{31.09} & 4.44 & 172.40 & 40.98 & 22.72 & 4.19 \\
\midrule
\multirow{2}{*}{3} 
& Human & 9.00 & 31.40 & 35.27 & 49.81 & 36.91 & 4.22 & 16.40 & 4.99 & 104.62 & 21.54 & 37.10 & 4.08 \\
& PPO-SDLA (ours) & 2.87 & 2.85 & 28.22 & 44.25 & 26.21 & 4.48 & \textbf{17.60} & 3.52 & 95.02 & 36.69 & 20.82 & 3.87 \\
\bottomrule
\end{tabular}
\end{table*}

\begin{table*}[t]
\centering

\caption{Performance on OmniSafe's MuJoCo Velocity Benchmark. Mean $\pm$ std over 5 random seeds. \safe{Green}: mean cost $\leq$ 25. \best{Bold Green}: highest mean reward within the same algorithm family (PPO-based or SAC-based) among methods meeting this budget.}
\label{tab:mujoco_velocity}
\scriptsize
\setlength{\tabcolsep}{2.8pt}
\resizebox{\textwidth}{!}{\begin{tabular}{@{}lcccccccccc@{}}
\toprule
\multirow{2}{*}{Method} & \multicolumn{2}{c}{Ant} & \multicolumn{2}{c}{HalfCheetah} & \multicolumn{2}{c}{Hopper} & \multicolumn{2}{c}{Swimmer} & \multicolumn{2}{c}{Walker2d} \\
\cmidrule(lr){2-3} \cmidrule(lr){4-5} \cmidrule(lr){6-7} \cmidrule(lr){8-9} \cmidrule(lr){10-11}
& $R\uparrow$ & $C\downarrow$ & $R\uparrow$ & $C\downarrow$ & $R\uparrow$ & $C\downarrow$ & $R\uparrow$ & $C\downarrow$ & $R\uparrow$ & $C\downarrow$ \\
\midrule
PPO & \shortstack{5977.73\\$\pm$885.65} & \shortstack{958.13\\$\pm$134.50} & \shortstack{6921.83\\$\pm$1721.79} & \shortstack{919.20\\$\pm$173.08} & \shortstack{2337.11\\$\pm$942.06} & \shortstack{550.02\\$\pm$237.70} & \shortstack{119.77\\$\pm$13.80} & \shortstack{165.27\\$\pm$20.15} & \shortstack{6239.52\\$\pm$879.99} & \shortstack{902.68\\$\pm$100.93} \\
PPOLag & \best{\shortstack{3205.96\\$\pm$85.16}} & \safe{\shortstack{22.46\\$\pm$4.82}} & \shortstack{2939.58\\$\pm$307.19} & \safe{\shortstack{18.72\\$\pm$2.87}} & \shortstack{961.92\\$\pm$752.87} & \safe{\shortstack{13.96\\$\pm$19.33}} & \shortstack{64.74\\$\pm$17.67} & \shortstack{28.02\\$\pm$4.09} & \shortstack{2982.27\\$\pm$681.55} & \safe{\shortstack{13.49\\$\pm$14.55}} \\
P3O & \shortstack{2636.62\\$\pm$181.09} & \safe{\shortstack{20.69\\$\pm$10.23}} & \shortstack{2117.84\\$\pm$313.55} & \shortstack{27.60\\$\pm$8.36} & \shortstack{1231.52\\$\pm$465.35} & \safe{\shortstack{16.33\\$\pm$11.38}} & \shortstack{38.02\\$\pm$34.18} & \safe{\shortstack{18.40\\$\pm$12.13}} & \shortstack{2713.57\\$\pm$313.20} & \safe{\shortstack{20.51\\$\pm$14.09}} \\
FOCOPS & \shortstack{3184.48\\$\pm$305.59} & \safe{\shortstack{14.75\\$\pm$6.36}} & \shortstack{2965.20\\$\pm$290.43} & \safe{\shortstack{2.37\\$\pm$3.50}} & \shortstack{1437.75\\$\pm$446.87} & \safe{\shortstack{10.13\\$\pm$8.87}} & \shortstack{53.87\\$\pm$17.90} & \shortstack{29.75\\$\pm$7.33} & \shortstack{3117.05\\$\pm$53.60} & \safe{\shortstack{8.78\\$\pm$12.38}} \\
PPOSaute & \shortstack{2978.74\\$\pm$93.65} & \safe{\shortstack{16.77\\$\pm$0.92}} & \shortstack{2901.40\\$\pm$25.49} & \safe{\shortstack{16.20\\$\pm$0.60}} & \shortstack{1650.91\\$\pm$152.65} & \safe{\shortstack{17.87\\$\pm$1.33}} & \shortstack{35.61\\$\pm$4.37} & \safe{\shortstack{3.44\\$\pm$1.35}} & \shortstack{2410.89\\$\pm$241.22} & \safe{\shortstack{18.88\\$\pm$2.38}} \\
PPOEarlyTermin & \shortstack{2801.53\\$\pm$19.66} & \safe{\shortstack{0.23\\$\pm$0.09}} & \shortstack{2447.25\\$\pm$346.84} & \safe{\shortstack{3.47\\$\pm$4.90}} & \shortstack{1643.39\\$\pm$2.58} & \safe{\shortstack{0.77\\$\pm$0.26}} & \shortstack{33.02\\$\pm$7.26} & \safe{\shortstack{24.23\\$\pm$0.54}} & \shortstack{2195.57\\$\pm$1046.29} & \safe{\shortstack{7.63\\$\pm$10.44}} \\
PPO-SDLA & \safe{\shortstack{3200.44\\$\pm$75.58}} & \safe{\shortstack{19.73\\$\pm$2.46}} & \best{\shortstack{3026.59\\$\pm$11.12}} & \safe{\shortstack{19.81\\$\pm$6.18}} & \best{\shortstack{1681.64\\$\pm$46.70}} & \safe{\shortstack{14.66\\$\pm$6.08}} & \best{\shortstack{115.77\\$\pm$57.56}} & \safe{\shortstack{20.39\\$\pm$3.56}} & \best{\shortstack{3146.70\\$\pm$79.06}} & \safe{\shortstack{14.24\\$\pm$5.61}} \\
\midrule
SAC & \shortstack{5456.31\\$\pm$156.04} & \shortstack{943.10\\$\pm$47.51} & \shortstack{11488.86\\$\pm$513.09} & \shortstack{981.93\\$\pm$0.33} & \shortstack{3537.70\\$\pm$32.23} & \shortstack{975.23\\$\pm$2.39} & \shortstack{46.44\\$\pm$1.23} & \shortstack{40.97\\$\pm$0.47} & \shortstack{4419.29\\$\pm$232.06} & \shortstack{877.70\\$\pm$8.95} \\
SACLag & \shortstack{1870.08\\$\pm$334.29} & \shortstack{42.45\\$\pm$7.49} & \shortstack{2596.66\\$\pm$12.07} & \shortstack{43.65\\$\pm$1.31} & \shortstack{963.49\\$\pm$291.64} & \safe{\shortstack{20.23\\$\pm$28.47}} & \shortstack{11.03\\$\pm$11.17} & \safe{\shortstack{22.70\\$\pm$32.10}} & \shortstack{2676.47\\$\pm$300.43} & \shortstack{30.67\\$\pm$32.30} \\
SAC-SDLA & \best{\shortstack{2316.61\\$\pm$95.88}} & \safe{\shortstack{14.64\\$\pm$13.00}} & \best{\shortstack{2863.05\\$\pm$9.53}} & \safe{\shortstack{15.06\\$\pm$11.27}} & \best{\shortstack{1136.15\\$\pm$300.73}} & \safe{\shortstack{15.96\\$\pm$12.09}} & \best{\shortstack{21.59\\$\pm$9.44}} & \safe{\shortstack{17.52\\$\pm$15.43}} & \best{\shortstack{2732.98\\$\pm$58.81}} & \safe{\shortstack{12.90\\$\pm$8.26}} \\
\bottomrule
\end{tabular}
}
\end{table*}

\section{Experiments}
\label{sec:exp}
We evaluate neural PPO-SDLA and SAC-SDLA on reward--cost performance, training-time constraint tracking, computational overhead, and component and parameter ablations.

\subsection{Experimental Setup}

\noindent\textbf{HASARD.}
HASARD~\citep{tomilin2025hasard} uses egocentric RGB observations. Following its protocol, the cost threshold is 5 for Remedy Rush, Collateral Damage, and Detonator's Dilemma, and 50 for all other environments.

\noindent\textbf{OmniSafe's MuJoCo Velocity Benchmark.}
We evaluate Ant, HalfCheetah, Hopper, Walker2d, and Swimmer with a cost limit of 25, following OmniSafe~\citep{ji2024omnisafe}. These velocity-constrained continuous-control tasks complement HASARD's visual observations.

\noindent\textbf{Baselines.}
We use baselines available in each benchmark's official implementation.
For HASARD, following \citep{tomilin2025hasard}, we compare against: PPOLag \citep{ray2019benchmarking}, 
PPOPID \citep{stooke2020responsivesafe}, PPOCost \citep{tomilin2025hasard}, P3O \citep{zhang2022penalized}, 
and PPOSaute \citep{sootla2022saute}.
For OmniSafe, we compare against: PPOLag, SACLag \citep{ray2019benchmarking}, 
P3O \citep{zhang2022penalized}, FOCOPS \citep{zhang2020first}, PPOSaute \citep{sootla2022saute}, and PPOEarlyTermin \citep{ji2024omnisafe}.

\noindent\textbf{Implementation Details.}
We implement SDLA on top of PPO and SAC. The default surrogate-dual parameters are $\tau=1$ and $\beta=0.1$ for both HASARD and OmniSafe Safety-Gymnasium. Ternary search 
runs for $K=15$ iterations per dual update.
Our experiments use 5 random seeds and $\lambda_{\max}=10$; HASARD baselines use archived runs (Appendix~\ref{sec:extra_exp}).

\noindent\textbf{Evaluation Metrics.}
We report episode reward and cost, comparing mean cost against the task budget.
HASARD tables report means and standard deviations across per-run summaries of terminal training performance.

\begin{figure*}[t]
\centering
\includegraphics[width=\textwidth]{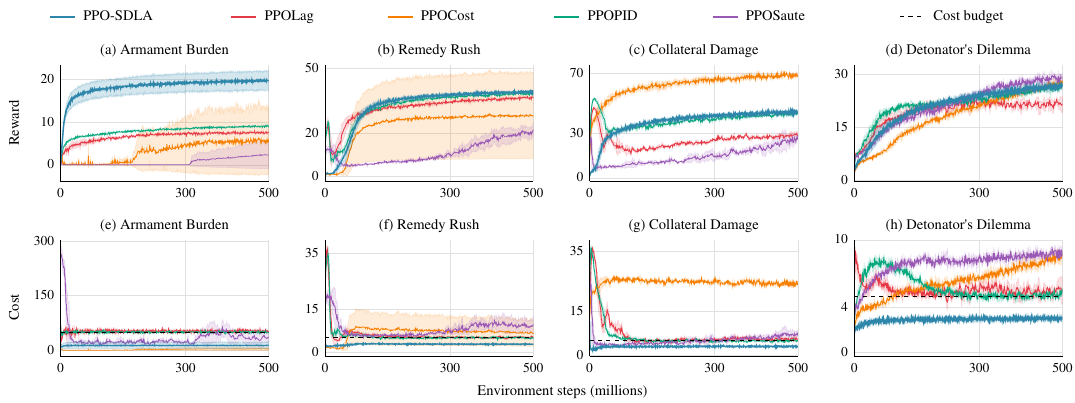}
\caption{HASARD Level 1: reward (top) and cost (bottom). Lines show means and shading one standard deviation over five SDLA runs and three runs per baseline. Dashed lines mark budgets of 50 for Armament Burden and 5 for the other tasks. PPOCost shows task reward with its logged penalty added back.}
\label{fig:hasard_main_l1}
\end{figure*}

\begin{figure*}[t]
\centering
\includegraphics[width=\textwidth]{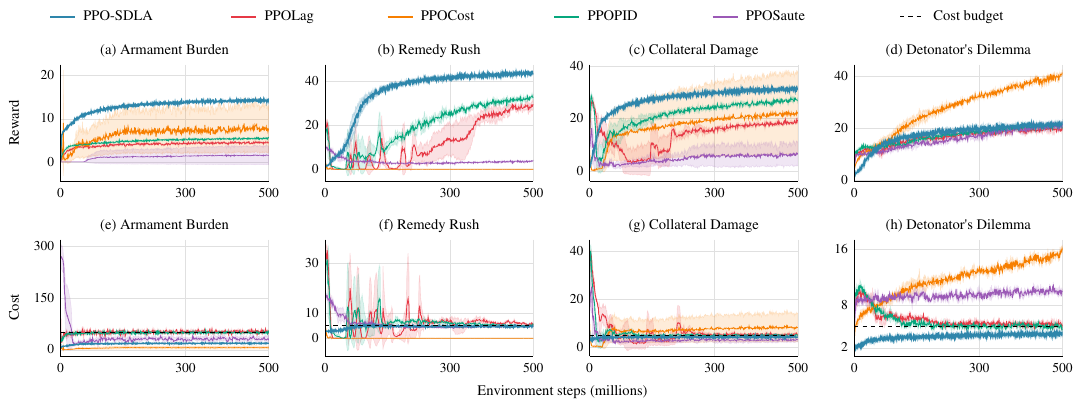}
\caption{HASARD Level 2: reward (top) and cost (bottom) for the same four tasks, with methods, budgets, and statistical conventions as in Figure~\ref{fig:hasard_main_l1}.}
\label{fig:hasard_main_l2}
\end{figure*}

\subsection{Results for HASARD}
\textbf{Baseline Comparison.} PPO-SDLA is the only compared method whose reported mean cost meets the budget in all 18 HASARD settings, and it has the highest mean reward among budget-meeting methods in each setting (Table~\ref{tab:baseline_comparison}). For example, Armament Burden L1 reward increases from PPOPID's $8.99$ to $19.46$, while mean cost is $12.80$ against a budget of $50$. Table~\ref{tab:baseline_comparison_std} reports standard deviations.

\textbf{Human Comparison.} On HASARD, PPO-SDLA achieves \textbf{higher reported mean reward than the human baseline in 6 of 18 task--difficulty settings, including four with lower mean cost} (Table~\ref{tab:human_comparison}). All six settings meet their respective mean-cost budgets. Human results from Tomilin et al.~\citep{tomilin2025hasard} are averaged over 10 episodes, while PPO-SDLA results use terminal training summaries.

Figures~\ref{fig:hasard_main_l1} and~\ref{fig:hasard_main_l2} compare reward and cost trajectories on four HASARD tasks at Levels 1 and 2. Learning curves for all 18 settings appear in Appendix~\ref{sec:extra_exp}.

\begin{table*}[!t]
\centering
\footnotesize
\caption{Training-time constraint tracking with $d=25$. All statistics use the complete training logs. Violation and CrossRate are percentages; overshoot is averaged over all logged updates, including those below the budget.}
\label{tab:training_tracking}
\setlength{\tabcolsep}{6pt}
\begin{tabular}{@{}llcccc@{}}
\toprule
Task & Method & Mean $|\Delta\lambda|$ & Violation (\%) & Overshoot & CrossRate (\%) \\
\midrule
Ant & PPOLag & $0.0075\pm0.0007$ & $38.7\pm3.5$ & $3.2546\pm0.2850$ & $5.2\pm0.8$ \\
Ant & PPO-SDLA & $0.0865\pm0.0162$ & $6.7\pm4.5$ & $0.1807\pm0.1653$ & $3.4\pm1.8$ \\
Walker2d & PPOLag & $0.0115\pm0.0031$ & $41.7\pm2.1$ & $4.8960\pm0.4163$ & $9.0\pm2.1$ \\
Walker2d & PPO-SDLA & $0.1154\pm0.0380$ & $13.0\pm6.9$ & $1.4061\pm1.2572$ & $4.0\pm1.2$ \\
\bottomrule
\end{tabular}
\end{table*}

\begin{table}[t]
\centering
\caption{Adaptive feedback versus fixed early penalization on Ant and Swimmer ($d=25$; five runs per method). Entries report mean $\pm$ standard deviation. PPO-QR-$d/2$ uses $\beta=0.2$ versus SDLA's $\beta=0.1$, matching their constant-cost activation threshold and response slope. Violation \% is the fraction of logged updates exceeding the cost limit.}
\label{tab:dual_effectiveness}
\footnotesize
\setlength{\tabcolsep}{3pt}
\resizebox{\columnwidth}{!}{\begin{tabular}{lccc}
\toprule
Algorithm & Reward ($\uparrow$) & Cost ($\downarrow$) & Violation (\%, $\downarrow$) \\
\midrule
\multicolumn{4}{c}{Ant} \\
\midrule
PPO-SDLA (ours) & $3200.4 \pm 75.6$ & $\mathbf{19.73 \pm 2.46}$ & $\mathbf{6.7 \pm 4.5}$ \\
PPOEntropy & $3136.4 \pm 86.6$ & $40.86 \pm 22.39$ & $27.8 \pm 3.5$ \\
PPO-QR & $3226.8 \pm 124.7$ & $51.35 \pm 50.00$ & $52.9 \pm 7.2$ \\
PPO-QR-\(\frac{d}{2}\) & $3169.3 \pm 112.6$ & $25.47 \pm 0.37$ & $24.8 \pm 4.9$ \\
\midrule
\multicolumn{4}{c}{Swimmer} \\
\midrule
PPO-SDLA (ours) & $\mathbf{115.8 \pm 57.6}$ & $\mathbf{20.39 \pm 3.56}$ & $\mathbf{27.1 \pm 11.9}$ \\
PPOEntropy & $96.2 \pm 58.9$ & $29.54 \pm 6.10$ & $87.7 \pm 5.4$ \\
PPO-QR & $69.6 \pm 46.4$ & $28.75 \pm 3.86$ & $77.9 \pm 7.7$ \\
PPO-QR-\(\frac{d}{2}\) & $92.6 \pm 58.6$ & $23.67 \pm 3.43$ & $74.1 \pm 18.8$ \\
\bottomrule
\end{tabular}}
\end{table}

\subsection{Performance on MuJoCo Velocity Benchmark}

\textbf{PPO-based methods.} Among the compared PPO-based methods meeting the mean-cost budget, PPO-SDLA achieves the highest mean reward on four of five tasks (Table~\ref{tab:mujoco_velocity}). On Ant, its reward is close to PPOLag's ($3200.44$ versus $3205.96$), with lower mean cost ($19.73$ versus $22.46$). Figures~\ref{fig:ppo_all} and~\ref{fig:sac_all} report the learning curves.

\textbf{SAC-based methods.} SAC-SDLA improves mean reward over SACLag on all five tasks with mean costs below budget; the Ant reward gain is $23.9\%$. On HalfCheetah, SACLag exceeds the cost budget ($43.65$), while SAC-SDLA stays below it ($15.06$).

PPO-SDLA adds approximately $1.5\%$ GPU time over PPOLag on Ant (Appendix~\ref{app:runtime}).

\subsection{Ablations}
\label{sec:ablations}

\textbf{Adaptive feedback versus fixed early penalization.} We compare SDLA with PPO-QR-$d/2$ to evaluate adaptive centering against the fixed center $d/2$.
For the fixed trajectory batch in~\eqref{eq:trajectory-objective}, the ablation objectives are
\begin{align}
\widehat g_{\mathrm{E}}(\lambda)&=\widehat F_M(\lambda)+\lambda d,
\label{eq:ablation-entropy}\\
\widehat g_{\mathrm{QR}}(\lambda)&=\widehat F_M(\lambda)+\lambda d+\frac{\lambda^2}{2\beta},
\label{eq:ablation-qr}\\
\widehat g_{\mathrm{QR}\text{-}d/2}(\lambda)&=\widehat F_M(\lambda)+\frac{\lambda d}{2}+\frac{\lambda^2}{2\beta}.
\label{eq:ablation-qr-half}
\end{align}
PPOEntropy, PPO-QR, and PPO-QR-$d/2$ minimize these respective objectives over $[0,\lambda_{\max}]$, using ternary search. All retain trajectory weighting; PPOEntropy omits quadratic regularization, while PPO-QR and PPO-QR-$d/2$ use fixed centers $d$ and $d/2$. SDLA uses $d-\widehat c_M$ in~\eqref{eq:surrogate-dual}. Here ``Entropy'' refers to trajectory log-sum-exp smoothing. The PPO backbone and budget $d=25$ are shared. PPO-QR-$d/2$ uses $\beta=0.2$ versus SDLA's $0.1$, matching their constant-cost activation threshold and slope (Appendix~\ref{app:regularization-ablation}).

Table~\ref{tab:dual_effectiveness} compares reward, cost, and violation rates across the surrogate variants. On Ant, relative to the fixed-margin PPO-QR-$d/2$ baseline, SDLA reduces the violation rate from $24.8\%$ to $6.7\%$ and mean cost from $25.47$ to $19.73$, with reported mean reward of $3200.4$ versus $3169.3$.
On Swimmer, the corresponding violation rate decreases from $74.1\%$ to $27.1\%$, with mean cost of $20.39$ versus $23.67$ and reported mean reward of $115.8$ versus $92.6$.
On Ant, mean $|\Delta\lambda|$ is similar for SDLA and PPO-QR-$d/2$ ($0.0865$ versus $0.0849$), while its 95th percentile falls from $0.4219$ to $0.3576$ (Appendix~\ref{app:regularization-ablation}).

Across $\beta\in\{0.1,0.5,1,2,5,\infty\}$, all final mean costs remain below $25$, with rewards between $3086.35$ and $3200.44$ (Table~\ref{tab:ablation_safety_ant}). Mean $|\Delta\lambda|$ increases monotonically from $0.0865$ to $0.8037$ as $\beta$ increases.

Appendix~\ref{app:temperature-ablation} compares $\tau\in\{1,1.5,2\}$ at the default $\beta=0.1$. All three settings have mean costs below $25$ and similar mean absolute multiplier changes.

\subsection{Constraint Tracking during Training}
Table~\ref{tab:training_tracking} compares full-training constraint tracking on Ant and Walker2d, with PPOLag rerun for multiplier logging. PPO-SDLA reduces violation rates from $38.7\%$ to $6.7\%$ on Ant and from $41.7\%$ to $13.0\%$ on Walker2d. Average overshoot decreases by $94.4\%$ and $71.3\%$, respectively, and boundary crossings are less frequent on both tasks. PPO-SDLA has larger mean multiplier updates alongside fewer and smaller violations. Appendix~\ref{sec:additional_diagnostics} gives metric definitions, plots, and multiplier-cap usage.

\section{Conclusion}
SDLA replaces fixed-policy bang-bang switching with structured smooth feedback of bounded sensitivity. Tabular analysis establishes convergence under mild assumptions; PPO/SAC experiments demonstrate competitive rewards and improved constraint tracking.

\bibliographystyle{IEEEtran}
\bibliography{references}

@inproceedings{ying2022dual,
  title={A dual approach to constrained markov decision processes with entropy regularization},
  author={Ying, Donghao and Ding, Yuhao and Lavaei, Javad},
  booktitle={International Conference on Artificial Intelligence and Statistics},
  pages={1887--1909},
  year={2022},
  organization={PMLR},
  url={https://proceedings.mlr.press/v151/ying22a.html}
}

@inproceedings{lin2020gradient,
  title={On gradient descent ascent for nonconvex-concave minimax problems},
  author={Lin, Tianyi and Jin, Chi and Jordan, Michael},
  booktitle={International conference on machine learning},
  pages={6083--6093},
  year={2020},
  organization={PMLR}
}

@inproceedings{stooke2020responsivesafe,
  title={Responsive safety in reinforcement learning by pid lagrangian methods},
  author={Stooke, Adam and Achiam, Joshua and Abbeel, Pieter},
  booktitle={International Conference on Machine Learning},
  pages={9133--9143},
  year={2020},
  organization={PMLR}
}

@inproceedings{dai2023augmented,
  title={Augmented proximal policy optimization for safe reinforcement learning},
  author={Dai, Juntao and Ji, Jiaming and Yang, Long and Zheng, Qian and Pan, Gang},
  booktitle={Proceedings of the AAAI Conference on Artificial Intelligence},
  volume={37},
  number={6},
  pages={7288--7295},
  year={2023}
}

@article{li2024faster,
  title={Faster algorithm and sharper analysis for constrained Markov decision process},
  author={Li, Tianjiao and Guan, Ziwei and Zou, Shaofeng and Xu, Tengyu and Liang, Yingbin and Lan, Guanghui},
  journal={Operations Research Letters},
  volume={54},
  pages={107107},
  doi={10.1016/j.orl.2024.107107},
  year={2024},
  publisher={Elsevier}
}

@techreport{ray2019benchmarking,
  title={Benchmarking safe exploration in deep reinforcement learning},
  author={Ray, Alex and Achiam, Joshua and Amodei, Dario},
  institution={OpenAI},
  year={2019},
  url={https://cdn.openai.com/safexp-short.pdf}
}

@book{altman2021constrained,
  title={Constrained Markov decision processes},
  author={Altman, Eitan},
  year={2021},
  publisher={Routledge}
}

@inproceedings{achiam2017constrained,
  title={Constrained policy optimization},
  author={Achiam, Joshua and Held, David and Tamar, Aviv and Abbeel, Pieter},
  booktitle={International conference on machine learning},
  pages={22--31},
  year={2017},
  organization={PMLR}
}

@inproceedings{tomilin2025hasard,
  title={HASARD: A Benchmark for Vision-Based Safe Reinforcement Learning in Embodied Agents},
  author={Tomilin, Tristan and Fang, Meng and Pechenizkiy, Mykola},
  booktitle={The Thirteenth International Conference on Learning Representations},
  year={2025}

}

@article{ji2024omnisafe,
  title={Omnisafe: An infrastructure for accelerating safe reinforcement learning research},
  author={Ji, Jiaming and Zhou, Jiayi and Zhang, Borong and Dai, Juntao and Pan, Xuehai and Sun, Ruiyang and Huang, Weidong and Geng, Yiran and Liu, Mickel and Yang, Yaodong},
  journal={Journal of Machine Learning Research},
  volume={25},
  number={285},
  pages={1--6},
  year={2024}
}

@inproceedings{sootla2022saute,
  title={Saut{\'e} rl: Almost surely safe reinforcement learning using state augmentation},
  author={Sootla, Aivar and Cowen-Rivers, Alexander I and Jafferjee, Taher and Wang, Ziyan and Mguni, David H and Wang, Jun and Ammar, Haitham},
  booktitle={International Conference on Machine Learning},
  pages={20423--20443},
  year={2022},
  organization={PMLR}
}

@article{zhang2020first,
  title={First order constrained optimization in policy space},
  author={Zhang, Yiming and Vuong, Quan and Ross, Keith},
  journal={Advances in Neural Information Processing Systems},
  volume={33},
  pages={15338--15349},
  year={2020}
}

@article{zhang2022penalized,
  title={Penalized proximal policy optimization for safe reinforcement learning},
  author={Zhang, Linrui and Shen, Li and Yang, Long and Chen, Shixiang and Yuan, Bo and Wang, Xueqian and Tao, Dacheng},
  journal={arXiv preprint arXiv:2205.11814},
  year={2022}
}

@article{sootla2022enhancing,
  title={Enhancing safe exploration using safety state augmentation},
  author={Sootla, Aivar and Cowen-Rivers, Alexander and Wang, Jun and Bou Ammar, Haitham},
  journal={Advances in Neural Information Processing Systems},
  volume={35},
  pages={34464--34477},
  year={2022}
}

@article{yang2020projection,
  title={Projection-based constrained policy optimization},
  author={Yang, Tsung-Yen and Rosca, Justinian and Narasimhan, Karthik and Ramadge, Peter J},
  journal={arXiv preprint arXiv:2010.03152},
  year={2020}
}

@article{yang2022constrained,
  title={Constrained update projection approach to safe policy optimization},
  author={Yang, Long and Ji, Jiaming and Dai, Juntao and Zhang, Linrui and Zhou, Binbin and Li, Pengfei and Yang, Yaodong and Pan, Gang},
  journal={Advances in Neural Information Processing Systems},
  volume={35},
  pages={9111--9124},
  year={2022}
}

@article{cen2022fast,
  title={Fast Global Convergence of Natural Policy Gradient Methods with Entropy Regularization},
  author={Cen, Shicong and Cheng, Chen and Chen, Yuxin and Wei, Yuting and Chi, Yuejie},
  journal={Operations Research},
  volume={70},
  number={4},
  pages={2563--2578},
  year={2022},
  doi={10.1287/opre.2021.2151},
  url={https://yuxinchen2020.github.io/publications/NPG_reg.pdf}
}

@inproceedings{liu2019neural,
  title={Neural Proximal/Trust Region Policy Optimization Attains Globally Optimal Policy},
  author={Liu, Boyi and Cai, Qi and Yang, Zhuoran and Wang, Zhaoran},
  booktitle={Advances in Neural Information Processing Systems},
  volume={32},
  year={2019},
  url={https://arxiv.org/abs/1906.10306}
}

@inproceedings{haarnoja2018soft,
  title={Soft actor-critic: Off-policy maximum entropy deep reinforcement learning with a stochastic actor},
  author={Haarnoja, Tuomas and Zhou, Aurick and Abbeel, Pieter and Levine, Sergey},
  booktitle={International conference on machine learning},
  pages={1861--1870},
  year={2018},
  organization={Pmlr}
}

@article{garcia2015comprehensive,
  title={A comprehensive survey on safe reinforcement learning},
  author={Garc{\'i}a, Javier and Fern{\'a}ndez, Fernando},
  journal={Journal of Machine Learning Research},
  volume={16},
  number={42},
  pages={1437--1480},
  year={2015}
}

@article{gu2024review,
  title={A review of safe reinforcement learning: Methods, theories and applications},
  author={Gu, Shangding and Yang, Long and Du, Yali and Chen, Guang and Walter, Florian and Wang, Jun and Knoll, Alois},
  journal={IEEE Transactions on Pattern Analysis and Machine Intelligence},
  volume={46},
  number={12},
  pages={11216--11235},
  doi={10.1109/TPAMI.2024.3457538},
  year={2024},
  publisher={IEEE}
}

@article{schulman2017proximal,
  title={Proximal policy optimization algorithms},
  author={Schulman, John and Wolski, Filip and Dhariwal, Prafulla and Radford, Alec and Klimov, Oleg},
  journal={arXiv preprint arXiv:1707.06347},
  year={2017}
}

@article{spoor2025empirical,
  title={An Empirical Study of Lagrangian Methods in Safe Reinforcement Learning},
  author={Spoor, Lindsay and Serra-G{\'o}mez, {\'A}lvaro and Plaat, Aske and Moerland, Thomas},
  journal={arXiv preprint arXiv:2510.17564v1},
  year={2025}
}

@article{chen2024adaptive,
  title={Adaptive primal-dual method for safe reinforcement learning},
  author={Chen, Weiqin and Onyejizu, James and Vu, Long and Hoang, Lan and Subramanian, Dharmashankar and Kar, Koushik and Mishra, Sandipan and Paternain, Santiago},
  journal={arXiv preprint arXiv:2402.00355},
  year={2024}
}

@inproceedings{liu2022constrained,
  title={Constrained Variational Policy Optimization for Safe Reinforcement Learning},
  author={Liu, Zuxin and Cen, Zhepeng and Isenbaev, Vladislav and Liu, Wei and Wu, Steven and Li, Bo and Zhao, Ding},
  booktitle={Proceedings of the 39th International Conference on Machine Learning},
  volume={162},
  pages={13644--13668},
  year={2022},
  publisher={PMLR},
  url={https://proceedings.mlr.press/v162/liu22b.html}
}

@inproceedings{xu2021crpo,
  title={{CRPO}: A New Approach for Safe Reinforcement Learning with Convergence Guarantee},
  author={Xu, Tengyu and Liang, Yingbin and Lan, Guanghui},
  booktitle={Proceedings of the 38th International Conference on Machine Learning},
  volume={139},
  pages={11480--11491},
  year={2021},
  publisher={PMLR},
  url={https://proceedings.mlr.press/v139/xu21a.html}
}

@article{paternain2022safe,
  title={Safe policies for reinforcement learning via primal-dual methods},
  author={Paternain, Santiago and Calvo-Fullana, Miguel and Chamon, Luiz FO and Ribeiro, Alejandro},
  journal={IEEE Transactions on Automatic Control},
  volume={68},
  number={3},
  pages={1321--1336},
  year={2023},
  doi={10.1109/TAC.2022.3152724},
  publisher={IEEE}
}

@article{wu2024off,
  title={Off-policy primal-dual safe reinforcement learning},
  author={Wu, Zifan and Tang, Bo and Lin, Qian and Yu, Chao and Mao, Shangqin and Xie, Qianlong and Wang, Xingxing and Wang, Dong},
  journal={arXiv preprint arXiv:2401.14758},
  year={2024}
}

@article{ding2023last,
  title={Last-iterate convergent policy gradient primal-dual methods for constrained mdps},
  author={Ding, Dongsheng and Wei, Chen-Yu and Zhang, Kaiqing and Ribeiro, Alejandro},
  journal={Advances in Neural Information Processing Systems},
  volume={36},
  pages={66138--66200},
  year={2023},
  url={https://proceedings.neurips.cc/paper_files/paper/2023/hash/d0949cbcec31c09431610553a284f94a-Abstract-Conference.html}
}

@article{montenegro2024last,
  title={Last-iterate global convergence of policy gradients for constrained reinforcement learning},
  author={Montenegro, Alessandro and Mussi, Marco and Papini, Matteo and Metelli, Alberto Maria},
  journal={Advances in Neural Information Processing Systems},
  volume={37},
  pages={126363--126416},
  year={2024},
  url={https://proceedings.neurips.cc/paper_files/paper/2024/hash/e47470631adb188a30f14c9738fe157b-Abstract-Conference.html}
}

@book{boyd2004convex,
  title={Convex Optimization},
  author={Boyd, Stephen and Vandenberghe, Lieven},
  publisher={Cambridge University Press},
  year={2004},
  url={https://web.stanford.edu/~boyd/cvxbook/}
}

@article{antczak2013lower,
  title={A Lower Bound for the Penalty Parameter in the Exact Minimax Penalty Function Method for Solving Nondifferentiable Extremum Problems},
  author={Antczak, T.},
  journal={Journal of Optimization Theory and Applications},
  volume={159},
  pages={437--453},
  year={2013},
  doi={10.1007/s10957-013-0335-3},
  url={https://link.springer.com/article/10.1007/s10957-013-0335-3}
}

@inproceedings{guo2019learning,
  title={Learning Mean-Field Games},
  author={Guo, Xin and Hu, Anran and Xu, Renyuan and Zhang, Junzi},
  booktitle={Advances in Neural Information Processing Systems},
  volume={32},
  year={2019},
  url={https://papers.neurips.cc/paper/8742-learning-mean-field-games.pdf}
}

@inproceedings{xie2021learning,
  title={Learning While Playing in Mean-Field Games: Convergence and Optimality},
  author={Xie, Qiaomin and Yang, Zhuoran and Wang, Zhaoran and Minca, Andreea},
  booktitle={International Conference on Machine Learning},
  volume={139},
  pages={11436--11447},
  year={2021},
  publisher={PMLR},
  url={https://proceedings.mlr.press/v139/xie21g.html}
}

@inproceedings{mandal2023performative,
  title={Performative Reinforcement Learning},
  author={Mandal, Debmalya and Triantafyllou, Stelios and Radanovic, Goran},
  booktitle={International Conference on Machine Learning},
  volume={202},
  pages={23642--23680},
  year={2023},
  publisher={PMLR},
  url={https://proceedings.mlr.press/v202/mandal23a.html}
}

@article{olshevsky2023small,
  title={A Small Gain Analysis of Single Timescale Actor Critic},
  author={Olshevsky, Alex and Gharesifard, Bahman},
  journal={SIAM Journal on Control and Optimization},
  volume={61},
  number={2},
  pages={980--1007},
  year={2023},
  doi={10.1137/22M1483335}
}

@article{gu2025safe,
  title={Safe and Balanced: A Framework for Constrained Multi-Objective Reinforcement Learning},
  author={Gu, Shangding and Sel, Bilgehan and Ding, Yuhao and Wang, Lu and Lin, Qingwei and Knoll, Alois and Jin, Ming},
  journal={IEEE Transactions on Pattern Analysis and Machine Intelligence},
  volume={47},
  number={5},
  pages={3322--3331},
  year={2025},
  doi={10.1109/TPAMI.2025.3528944}
}
\clearpage
\onecolumn
\markboth{Appendices}{Structural Dual Adaptation}
\setcounter{equation}{0}
\setcounter{table}{0}
\setcounter{figure}{0}
\setcounter{algorithm}{0}
\providecommand{\theHequation}{}
\renewcommand{\theHequation}{supp.\arabic{equation}}
\providecommand{\theHtable}{}
\renewcommand{\theHtable}{supp.\arabic{table}}
\providecommand{\theHfigure}{}
\renewcommand{\theHfigure}{supp.\arabic{figure}}
\providecommand{\theHalgorithm}{}
\renewcommand{\theHalgorithm}{supp.\arabic{algorithm}}

\renewcommand{\theequation}{S\arabic{equation}}
\renewcommand{\thetable}{S\arabic{table}}
\renewcommand{\thefigure}{S\arabic{figure}}
\renewcommand{\thealgorithm}{S\arabic{algorithm}}
\appendices
\section{Notations Table}
\begin{table}[h]
\caption{Notation Table}
\label{tab:notation}
\centering
\begin{tabular}{p{0.18\linewidth}p{0.76\linewidth}}
\toprule
\textbf{Symbol} & \textbf{Description} \\
\midrule
\multicolumn{2}{l}{\textit{CMDP}} \\
$\mathcal S,\mathcal A,P$ & State space, action space, and transition kernel \\
$r,c,\gamma,d$ & Reward, cost, discount factor, and cost budget \\
$J^r(\pi),J^c(\pi)$ & Expected discounted reward and cost under policy $\pi$ \\
$R,C$ & Complete discounted trajectory returns \\
$\bar C$ & Cost-return bound $C_{\max}/(1-\gamma)$ \\
\midrule
\multicolumn{2}{l}{\textit{Policy and Trajectory Objectives}} \\
$\alpha,H(\pi),J_\alpha(\pi)$ & Policy-entropy coefficient, discounted entropy, and regularized reward \\
$F_\pi(\lambda),\widehat F_M(\lambda)$ & Population and empirical trajectory log-sum-exp objectives \\
$c_\pi,\widehat c_M$ & Population mean cost and uniform sample mean of trajectory costs \\
$m_\pi(\lambda),\widehat m(\lambda)$ & Population and empirical trajectory-weighted costs \\
\midrule
\multicolumn{2}{l}{\textit{SDLA}} \\
$\tau,\beta$ & Trajectory temperature and quadratic regularization parameter \\
$\lambda_{\max}$ & Upper bound for scalar dual optimization \\
$g_\pi,\widehat g_M$ & Population and empirical centered surrogate duals \\
$\widehat\lambda^*$ & Exact empirical dual minimizer \\
$\varepsilon_\lambda$ & Absolute multiplier error of the scalar solve \\
$\Lambda,\ell_t,\varepsilon_t^{\mathrm{lag}}$ & Population response map, its value at $\pi_t$, and multiplier-lag bound \\
$\mathcal D_t,\delta_t$ & Composite-cost signal change and recomputed multiplier change \\
$\overline V_{t+1}$ & Average trajectory-weighted cost variance divided by $\tau$ in~\eqref{eq:release-law} \\
$u_t,v_t,p_t,\kappa_t$ & Reward/cost gradients, actor direction, and local balance coefficient \\
$\kappa,q,\varepsilon_t,\zeta_t$ & Quadratic-growth constant, two-step contraction factor, actor objective gap, and dual error bound \\
$K_t,\omega,\chi,\xi_t$ & Actor inner steps, mixing weight, inner contraction factor, and critic error bound \\
$A_0,A_1,\mathcal E_{K,\xi,\zeta}$ & Uniform actor-error constants and coupled asymptotic error radius \\
$\pi_\alpha^\star,\lambda_\alpha^\star,\beta^\star(\tau)$ & Entropy-regularized CMDP saddle point and matching parameter \\
$\pi_s,\sigma,\Delta,V_{\beta,L}$ & Strictly feasible reference policy, feasibility margin, objective range bound, and equilibrium violation bound \\
\midrule
\multicolumn{2}{l}{\textit{Actor-Critic}} \\
$Q^r_{\phi_r},Q^c_{\phi_c},\pi_\theta$ & Reward critic, cost critic, and parameterized policy \\
\bottomrule
\end{tabular}
\end{table}

\section{Proofs}
\label{sec:proofs}
\subsection{Properties of the Surrogate Dual}

\subsubsection{Proof of Lemma~\ref{lem:smooth_convex}}
\begin{proof}
Bounded returns and $\tau>0$ justify differentiation under the expectation in~\eqref{eq:population-weighted-cost}. Using $F_\pi'=-m_\pi$ and $m_\pi'=-\operatorname{Var}_{q_{\pi,\lambda}}(C)/\tau$ gives~\eqref{eq:derivs}--\eqref{eq:curvature}. For $C\in[0,\bar C]$, the variance lies in $[0,\bar C^2/4]$, giving the strong-convexity and smoothness constants.
Strong convexity and compactness give a unique response. The constrained first-order conditions give
$\Lambda(\pi)=0$ exactly when $g_\pi'(0)\ge0$, and
$\Lambda(\pi)=\lambda_{\max}$ exactly when $g_\pi'(\lambda_{\max})\le0$.
Otherwise, $g_\pi'(\Lambda(\pi))=0$ gives the interior cost-signal identity.
These conditions establish~\eqref{eq:activate} and the three response regimes.
Strong convexity implies
$g_\pi'(\lambda)\ge d-c_\pi-m_\pi(0)+\lambda/\beta$.
If $c_\pi+m_\pi(0)\le d$, the response is zero. Otherwise, the interior equation
or the upper endpoint inequality gives
$\Lambda(\pi)\le\beta(c_\pi+m_\pi(0)-d)$.

\end{proof}

\subsubsection{Trajectory Weighting and the Cost Signal}
At $\lambda=0$, the population trajectory-weighted cost and mean cost satisfy
\begin{equation}
m_\pi(0)-c_\pi
=\frac{\operatorname{Cov}_\pi(C,e^{R/\tau})}
{\mathbb E_\pi[e^{R/\tau}]}.
\label{eq:cov}
\end{equation}
Expanding the covariance gives the identity. Its sign determines whether trajectory weighting raises or lowers the cost signal relative to the policy mean.

\subsection{Multiplier Feedback and Feasible Policy Updates}
\label{app:penalty-release}
We use the population definitions of Section~\ref{sec:surrogate_duality}, the structural properties of Section~\ref{sec:structural-properties}, and the notation of Section~\ref{sec:penalty-release}. The local actor assumptions and delayed updates are specified in Appendix~\ref{app:actor-regimes}. The convergence and optimality proofs use the entropy-regularized actor best responses of Section~\ref{sec:concrete-actor} and delayed dual update~\eqref{eq:inexact-dual-update}. For a policy $\pi_\theta$, write $g_\theta$ and $m_\theta$ for $g_{\pi_\theta}$ and $m_{\pi_\theta}$ in~\eqref{eq:release-population} and~\eqref{eq:population-weighted-cost}. Let $q_{\theta,\lambda}$ be the reweighted trajectory law with density proportional to $e^{(R-\lambda C)/\tau}$ relative to the original trajectory law. Bounded returns justify differentiation and give
\begin{equation}
g_\theta'(\lambda)=d-c(\theta)-m_\theta(\lambda)+\lambda/\beta,\qquad
m_\theta'(\lambda)=-\frac{\operatorname{Var}_{q_{\theta,\lambda}}(C)}{\tau}.
\label{eq:release-derivatives}
\end{equation}
Hence $g_\theta''(\lambda)=\beta^{-1}+\operatorname{Var}_{q_{\theta,\lambda}}(C)/\tau\ge\beta^{-1}$ on the common interval $[0,\lambda_{\max}]$. We use $g_t=g_{\theta_t}$ and $m_t=m_{\theta_t}$.

\subsubsection{Proof of Theorem~\ref{thm:penalty-release}}
\begin{proof}
The constrained optimality conditions give
\[
g_t'(\ell_t)(\ell_{t+1}-\ell_t)\ge0,\qquad
g_{t+1}'(\ell_{t+1})(\ell_t-\ell_{t+1})\ge0.
\]
Since $g_{t+1}'(\ell_t)-g_t'(\ell_t)=-\mathcal D_t$, combining these inequalities yields
\[
\delta_t\bigl[g_{t+1}'(\ell_{t+1})-g_{t+1}'(\ell_t)\bigr]
\le\delta_t\mathcal D_t.
\]
Strong monotonicity of $g_{t+1}'$ bounds the left-hand side below by $\delta_t^2/\beta$. This proves~\eqref{eq:release-feedback-bound}, including the absolute-value bound when $\delta_t\ne0$; it also holds when $\delta_t=0$. A nonzero response change must have the same sign as $\mathcal D_t$. If $\ell_t$ is interior and $\mathcal D_t\ne0$, then $g_t'(\ell_t)=0$ and $g_{t+1}'(\ell_t)=-\mathcal D_t\ne0$. The unique new minimizer therefore moves strictly in the indicated direction.

If both responses are interior, $g_t'(\ell_t)=g_{t+1}'(\ell_{t+1})=0$. With
\begin{equation}
\overline V_{t+1}:=\frac1\tau\int_0^1
\operatorname{Var}_{q_{\theta_{t+1},\ell_t+z\delta_t}}(C)\,dz\ge0,
\label{eq:release-variance}
\end{equation}
the fundamental theorem of calculus gives
\[
\mathcal D_t=g_{t+1}'(\ell_{t+1})-g_{t+1}'(\ell_t)
=(\beta^{-1}+\overline V_{t+1})\delta_t.
\]
This proves~\eqref{eq:release-law}.
\end{proof}

\subsubsection{Feasible actor regimes under delayed dual feedback}
\label{app:actor-regimes}
\textbf{Actor regularity and updates.} For the actor analysis, assume $r,c$ have locally Lipschitz gradients near the iterate under consideration and $v_t:=\nabla c(\theta_t)\ne0$. Write $u_t=\nabla r(\theta_t)$ and $\kappa_t=u_t^\top v_t/\|v_t\|^2$. Local smoothness controls the second-order error of a finite actor step, while $\kappa_t$ is the penalty at which its first-order cost change vanishes. Following the ordering in Algorithm~\ref{alg:general-convex}, initialize $\lambda_0\in[0,\lambda_{\max}]$ and consider
\begin{equation}
\theta_{t+1}=\theta_t+\eta_t p_t,\qquad
p_t=u_t-\lambda_t v_t,\qquad
\lambda_{t+1}=\ell_t.
\label{eq:release-delayed-update}
\end{equation}
The actor uses the previous dual response, and the new multiplier is computed from the policy that collected the batch. The feedback law relates population dual responses; actor regularity connects these responses to finite policy steps.

For $t\ge1$, the delayed update gives $\lambda_t=\ell_{t-1}$. Applying Theorem~\ref{thm:penalty-release} to consecutive policies yields
\begin{equation}
|\lambda_t-\ell_t|\le\varepsilon_t^{\mathrm{lag}},
\qquad
\varepsilon_t^{\mathrm{lag}}:=\beta\left|
c_t-c_{t-1}+m_t(\lambda_t)-m_{t-1}(\lambda_t)
\right|.
\label{eq:release-lag}
\end{equation}
The bound measures the difference between the multiplier used by the actor and the current dual response. The following result compares that response with $\kappa_t$, allowing for this difference, to determine the direction of the actor's cost change. A positive safety margin also leaves room for cost to increase during a reward-improving step.

\begin{theorem}[Feasible actor regimes under delayed dual feedback]
\label{thm:actor-regimes}
Under the assumptions and actor updates above, let $\varepsilon_t^{\mathrm{lag}}\ge0$ satisfy $|\lambda_t-\ell_t|\le\varepsilon_t^{\mathrm{lag}}$.

\textbf{(i) Feasible reward recovery.} If $c_t<d$ and $\ell_t+\varepsilon_t^{\mathrm{lag}}<\kappa_t$, then $\lambda_t<\kappa_t$. For all sufficiently small positive $\eta_t$, the actor step in~\eqref{eq:release-delayed-update} satisfies
\begin{equation}
r(\theta_{t+1})>r(\theta_t),\qquad
c_t<c(\theta_{t+1})<d.
\label{eq:release-recovery}
\end{equation}

\textbf{(ii) Feasible-region constraint suppression.} If $c_t\le d$ and $\ell_t-\varepsilon_t^{\mathrm{lag}}>\kappa_t$, then $\lambda_t>\kappa_t$. For all sufficiently small positive $\eta_t$, the same actor step satisfies
\begin{equation}
c(\theta_{t+1})<c_t\le d.
\label{eq:suppression-cost}
\end{equation}
\end{theorem}
In part~(i), the actor increases reward and cost within the available safety margin. In part~(ii), the cost-decreasing update expands the margin. The multiplier comparisons apply throughout the clipped interval. Recomputing the multiplier before the actor step gives $\varepsilon_t^{\mathrm{lag}}=0$.

\textbf{Safety-margin criteria.} The surrogate derivative converts these multiplier comparisons into conditions on the available margin. When the comparison point $\kappa_t^-:=\kappa_t-\varepsilon_t^{\mathrm{lag}}$ lies in $(0,\lambda_{\max})$, the condition
\begin{equation}
d-c_t>m_t(\kappa_t^-)-\kappa_t^-/\beta
\label{eq:release-margin}
\end{equation}
gives $g_t'(\kappa_t^-)>0$, hence $\ell_t<\kappa_t^-$ and the recovery comparison in part~(i). When $\kappa_t^+:=\kappa_t+\varepsilon_t^{\mathrm{lag}}$ lies in $(0,\lambda_{\max})$, the condition
\begin{equation}
d-c_t<m_t(\kappa_t^+)-\kappa_t^+/\beta
\label{eq:suppression-margin}
\end{equation}
gives $g_t'(\kappa_t^+)<0$, hence $\ell_t>\kappa_t^+$ and the suppression comparison in part~(ii). These thresholds relate the available margin to trajectory feedback and actor gradients. From a strictly feasible policy, condition~\eqref{eq:release-margin} supports reward recovery within the budget. From a feasible policy, condition~\eqref{eq:suppression-margin} gives a cost-decreasing direction that expands the margin. Both conclusions use the actor assumptions and sufficiently small steps of Theorem~\ref{thm:actor-regimes}.

\subsubsection{Proof of Theorem~\ref{thm:actor-regimes}}
\begin{proof}
For the updates in~\eqref{eq:release-delayed-update}, apply~\eqref{eq:release-feedback-bound} to $\theta_{t-1},\theta_t$. Since $\ell_{t-1}=\lambda_t$, it gives
\begin{equation}
|\lambda_t-\ell_t|
\le\beta\left|c_t-c_{t-1}+m_t(\lambda_t)-m_{t-1}(\lambda_t)\right|
=\varepsilon_t^{\mathrm{lag}}.
\label{eq:release-lag-bound}
\end{equation}

\textbf{(i) Recovery.} From $\ell_t+\varepsilon_t^{\mathrm{lag}}<\kappa_t$ and the multiplier-error bound,
\begin{equation}
0\le\lambda_t\le\ell_t+\varepsilon_t^{\mathrm{lag}}<\kappa_t.
\label{eq:release-balance}
\end{equation}
Since $(u_t-\kappa_t v_t)^\top v_t=0$, the direction $p_t=u_t-\lambda_t v_t$ satisfies
\begin{align}
v_t^\top p_t&=(\kappa_t-\lambda_t)\|v_t\|^2>0,
\label{eq:release-cost-direction}\\
u_t^\top p_t&=\|u_t-\kappa_t v_t\|^2+\kappa_t(\kappa_t-\lambda_t)\|v_t\|^2>0.
\label{eq:release-reward-direction}
\end{align}
Here $\kappa_t>0$ follows from $0\le\lambda_t<\kappa_t$.

Let $L_{\nabla r},L_{\nabla c}$ be local gradient Lipschitz constants and take $\eta_t$ small enough to keep the step within their neighborhood. Smoothness gives
\begin{align*}
r(\theta_t+\eta_t p_t)-r(\theta_t)
&\ge\eta_t u_t^\top p_t-\frac{L_{\nabla r}\eta_t^2}{2}\|p_t\|^2,\\
\left|c(\theta_t+\eta_t p_t)-c_t-\eta_t v_t^\top p_t\right|
&\le\frac{L_{\nabla c}\eta_t^2}{2}\|p_t\|^2.
\end{align*}
Both directional derivatives are positive, so all sufficiently small positive steps satisfy
\[
\eta_t v_t^\top p_t+\tfrac12 L_{\nabla c}\eta_t^2\|p_t\|^2<d-c_t,\qquad
\tfrac12 L_{\nabla c}\eta_t\|p_t\|^2<v_t^\top p_t,\qquad
\tfrac12 L_{\nabla r}\eta_t\|p_t\|^2<u_t^\top p_t.
\]
The first inequality keeps the cost increase within the available margin; the other two give strict cost and reward increases. This proves~\eqref{eq:release-recovery}.

\textbf{(ii) Feasible-region constraint suppression.} The other regime condition gives
\[
\lambda_t\ge\ell_t-\varepsilon_t^{\mathrm{lag}}>\kappa_t.
\]
Thus the actor direction locally reduces cost:
\[
v_t^\top p_t=(\kappa_t-\lambda_t)\|v_t\|^2<0.
\]
For a local gradient Lipschitz constant $L_{\nabla c}$, smoothness along the step gives
\[
c(\theta_t+\eta_t p_t)-c_t
\le\eta_t v_t^\top p_t+\frac{L_{\nabla c}\eta_t^2}{2}\|p_t\|^2.
\]
For all sufficiently small positive $\eta_t$, this bound gives $c(\theta_{t+1})<c_t\le d$, proving~\eqref{eq:suppression-cost}.

Finally, the safety-margin criteria follow from the same surrogate derivative. If $\kappa_t^-\in(0,\lambda_{\max})$, condition~\eqref{eq:release-margin} gives
\[
g_t'(\kappa_t^-)=d-c_t-m_t(\kappa_t^-)+\kappa_t^-/\beta>0.
\]
Strict convexity places the constrained minimizer below this comparison point, so $\ell_t<\kappa_t^-$ and $\ell_t+\varepsilon_t^{\mathrm{lag}}<\kappa_t$. Similarly, if $\kappa_t^+\in(0,\lambda_{\max})$, condition~\eqref{eq:suppression-margin} gives
\[
g_t'(\kappa_t^+)=d-c_t-m_t(\kappa_t^+)+\kappa_t^+/\beta<0,
\]
which implies $\ell_t>\kappa_t^+$ and $\ell_t-\varepsilon_t^{\mathrm{lag}}>\kappa_t$. Each criterion therefore implies the corresponding regime comparison.
\end{proof}

\subsubsection{Convergence with Approximate Actor Best Responses}
\label{app:parameter-convergence}
We use $\Pi$, $\mathcal L_\alpha$, $B$, $\kappa$, and $L_\Lambda$ from Section~\ref{sec:concrete-actor}. We establish convergence of delayed tabular updates with approximate actor best responses under mild assumptions specified below. Theorem~\ref{thm:concrete-actor-convergence} specializes the analysis to concrete actor updates.

The soft-policy optimality identity underlying Proposition~3.5 of \citep{ying2022dual} gives, with natural logarithms, the quadratic-growth bound
\begin{equation}
\begin{aligned}
\mathcal L_\alpha(B(\lambda),\lambda)-\mathcal L_\alpha(\pi,\lambda)
&\ge\frac{\kappa}{2}\|\pi-B(\lambda)\|_F^2,\\
\kappa=\alpha\rho_{\min},\qquad \rho_{\min}&=\min_s\rho(s)>0.
\end{aligned}
\label{eq:actor-quadratic-growth}
\end{equation}

\begin{theorem}[Convergence of SDLA]
\label{thm:parameter-convergence}
Under the finite-CMDP and entropy assumptions of Section~\ref{sec:concrete-actor}, fix $\beta,\tau,\lambda_{\max}>0$ such that
\begin{equation}
q:=\frac{K_cL_\Lambda(\beta,\tau)}{\kappa}<1.
\label{eq:sdla-contraction}
\end{equation}
Starting from any $\pi_0\in\Pi$ and $\lambda_0\in[0,\lambda_{\max}]$, let $\pi_t\in\Pi$ and $\lambda_t\in[0,\lambda_{\max}]$ satisfy, for $t\ge0$,
\begin{equation}
\mathcal L_\alpha(B(\lambda_t),\lambda_t)-\mathcal L_\alpha(\pi_{t+1},\lambda_t)\le\varepsilon_t,
\label{eq:inexact-actor-update}
\end{equation}
and the delayed dual bound~\eqref{eq:inexact-dual-update}, where $\varepsilon_t,\zeta_t\ge0$. Then there is a unique equilibrium $(\bar\pi,\bar\lambda)$ with $\bar\pi=B(\bar\lambda)$ and $\bar\lambda=\Lambda_{\beta,\tau}(\bar\pi)$. Writing $e_t=\|\pi_t-\bar\pi\|_F$, for $t\ge1$,
\begin{equation}
e_{t+1}\le q e_{t-1}+\sqrt{2\varepsilon_t/\kappa}+(K_c/\kappa)\zeta_{t-1}.
\label{eq:delayed-error-recursion}
\end{equation}
For exact updates, with $E_0=\max\{e_0,e_1\}$,
\begin{equation}
e_t\le q^{\lfloor t/2\rfloor}E_0,\qquad
|\lambda_{t+1}-\bar\lambda|\le L_\Lambda(\beta,\tau)q^{\lfloor t/2\rfloor}E_0.
\label{eq:exact-delayed-rate}
\end{equation}
If $\varepsilon_t\le\varepsilon$ and $\zeta_t\le\zeta$, then
\begin{equation}
\limsup_{t\to\infty}e_t
\le\frac{\sqrt{2\varepsilon/\kappa}+(K_c/\kappa)\zeta}{1-q}.
\label{eq:inexact-error-floor}
\end{equation}
If both errors vanish, $(\pi_t,\lambda_t)\to(\bar\pi,\bar\lambda)$.
\end{theorem}

\begin{proof}
Fix $\alpha,\beta,\tau,\lambda_{\max}>0$ and abbreviate $L_\Lambda=L_\Lambda(\beta,\tau)$. We first derive the regularity constants, then establish contraction of the response maps and convergence of the delayed updates.

\textbf{Trajectory-weighted cost regularity.}
For $\pi,\pi'\in\Pi$, couple their trajectories using the same initial state and the same transition randomness while their state-action pairs agree. At a common state, use a maximal coupling of the action distributions. Let
\[
\Delta_\pi=\max_s\operatorname{TV}(\pi(\cdot\mid s),\pi'(\cdot\mid s))
\le\frac{\sqrt{|\mathcal A|}}2\|\pi-\pi'\|_F.
\]
The probability of a first action disagreement by time $j$ is at most $(j+1)\Delta_\pi$. With $R,C$ and $R',C'$ denoting the coupled discounted returns, bounded one-step rewards and costs imply
\begin{align}
\mathbb E|C-C'|&\le C_{\max}\sum_{j\ge0}\gamma^j(j+1)\Delta_\pi
\le K_c\|\pi-\pi'\|_F,\nonumber\\
\mathbb E|R-R'|&\le K_r\|\pi-\pi'\|_F,\qquad
K_c=\frac{C_{\max}\sqrt{|\mathcal A|}}{2(1-\gamma)^2},\quad
K_r=\frac{R_{\max}\sqrt{|\mathcal A|}}{2(1-\gamma)^2}.
\label{eq:coupled-return-lipschitz}
\end{align}
In particular, $|J^c(\pi)-J^c(\pi')|\le K_c\|\pi-\pi'\|_F$. Write $\bar R=R_{\max}/(1-\gamma)$ and
\[
Z_\pi(\lambda)=\mathbb E_\pi e^{(R-\lambda C)/\tau},\qquad
N_\pi(\lambda)=\mathbb E_\pi[Ce^{(R-\lambda C)/\tau}].
\]
For all $\lambda\in[0,\lambda_{\max}]$, the exponential weights lie between $e^{-\lambda_{\max}\bar C/\tau}$ and $e^{\bar R/\tau}$. The mean-value inequality and~\eqref{eq:coupled-return-lipschitz} give
\begin{align*}
|Z_\pi(\lambda)-Z_{\pi'}(\lambda)|
&\le\frac{e^{\bar R/\tau}}{\tau}(K_r+\lambda_{\max}K_c)\|\pi-\pi'\|_F,\\
|N_\pi(\lambda)-N_{\pi'}(\lambda)|
&\le e^{\bar R/\tau}
\left[K_c+\frac{\bar C}{\tau}(K_r+\lambda_{\max}K_c)\right]\|\pi-\pi'\|_F.
\end{align*}
Since $m_{\pi,\tau}=N_\pi/Z_\pi$ and $0\le m_{\pi,\tau}\le\bar C$,
\[
|m_{\pi,\tau}(\lambda)-m_{\pi',\tau}(\lambda)|
\le\frac{|N_\pi-N_{\pi'}|+\bar C|Z_\pi-Z_{\pi'}|}{Z_\pi}.
\]
Consequently,
\begin{equation}
\begin{aligned}
|m_{\pi,\tau}(\lambda)-m_{\pi',\tau}(\lambda)|
&\le K_m(\tau)\|\pi-\pi'\|_F,\\
K_m(\tau)&=e^{(\bar R+\lambda_{\max}\bar C)/\tau}
\left[K_c+\frac{2\bar C}{\tau}(K_r+\lambda_{\max}K_c)\right].
\end{aligned}
\label{eq:weighted-cost-policy-lipschitz}
\end{equation}
This bound is uniform over the clipped multiplier interval.

\textbf{Dual-response sensitivity.}
Apply Theorem~\ref{thm:penalty-release} to $\pi$ and $\pi'$, evaluating the weighted-cost change at $\ell=\Lambda_{\beta,\tau}(\pi)$. The combined signal is bounded by
\[
|J^c(\pi')-J^c(\pi)|+|m_{\pi',\tau}(\ell)-m_{\pi,\tau}(\ell)|
\le[K_c+K_m(\tau)]\|\pi-\pi'\|_F.
\]
Thus, including responses at the endpoints,
\begin{equation}
|\Lambda_{\beta,\tau}(\pi)-\Lambda_{\beta,\tau}(\pi')|
\le L_\Lambda\|\pi-\pi'\|_F.
\label{eq:parameter-response-lipschitz}
\end{equation}

\textbf{Actor best-response sensitivity.}
The policy space is compact and $\mathcal L_\alpha$ is continuous, so an actor maximizer exists. The normalized discounted visitation distribution satisfies
\[
d_\rho^\pi(s)=(1-\gamma)\sum_{j\ge0}\gamma^j\Pr_\pi(s_j=s)
\ge(1-\gamma)\rho(s)\ge(1-\gamma)\rho_{\min}.
\]
The soft Bellman optimum obeys $B(\lambda)(a\mid s)\propto\exp(Q_\lambda^\star(s,a)/\alpha)$. Subtracting the Bellman expectation equation for $V_\lambda^\pi$ from the soft optimality equation and iterating the residual gives the identity underlying Proposition~3.5 of \citep{ying2022dual}:
\begin{equation}
\mathcal L_\alpha(B(\lambda),\lambda)-\mathcal L_\alpha(\pi,\lambda)
=\frac{\alpha}{1-\gamma}\sum_s d_\rho^\pi(s)
\operatorname{KL}\!\left(\pi(\cdot\mid s)\,\Vert\,B(\lambda)(\cdot\mid s)\right).
\label{eq:soft-suboptimality-identity}
\end{equation}
Pinsker's inequality for natural logarithms gives $\operatorname{KL}(p\Vert p')\ge\frac12\|p-p'\|_1^2\ge\frac12\|p-p'\|_2^2$. Together with $d_\rho^\pi(s)\ge(1-\gamma)\rho_{\min}$, this proves~\eqref{eq:actor-quadratic-growth} with $\kappa=\alpha\rho_{\min}$ and makes the maximizer unique. The actor reward is $r-\lambda c$, up to the constant $\lambda d$.
For $\pi_i=B(\lambda_i)$, apply the quadratic-growth bound at each multiplier and add the two inequalities. The reward and entropy terms cancel, giving
\[
\kappa\|\pi_1-\pi_2\|_F^2
\le(\lambda_2-\lambda_1)[J^c(\pi_1)-J^c(\pi_2)]
\le K_c|\lambda_1-\lambda_2|\,\|\pi_1-\pi_2\|_F.
\]
Thus
\begin{equation}
\|B(\lambda)-B(\lambda')\|_F\le(K_c/\kappa)|\lambda-\lambda'|.
\label{eq:actor-best-response-lipschitz}
\end{equation}

\textbf{Existence and uniqueness.}
Define $F_{\beta,\tau}(\lambda)=\Lambda_{\beta,\tau}(B(\lambda))$. Equations~\eqref{eq:parameter-response-lipschitz} and~\eqref{eq:actor-best-response-lipschitz} imply
\begin{equation}
|F_{\beta,\tau}(\lambda)-F_{\beta,\tau}(\lambda')|
\le q|\lambda-\lambda'|.
\label{eq:parameter-loop-contraction}
\end{equation}
Since $q<1$, this self-map of the closed interval $[0,\lambda_{\max}]$ has a unique fixed point $\bar\lambda$. Setting $\bar\pi=B(\bar\lambda)$ gives the unique equilibrium.

\textbf{Delayed and inexact updates.}
The actor objective gap and~\eqref{eq:actor-quadratic-growth} give
\[
\|\pi_{t+1}-B(\lambda_t)\|_F\le\sqrt{2\varepsilon_t/\kappa}.
\]
For $t\ge1$, the delayed dual update implies
\[
|\lambda_t-\bar\lambda|
\le\zeta_{t-1}+L_\Lambda e_{t-1}.
\]
The triangle inequality and~\eqref{eq:actor-best-response-lipschitz} now yield
\[
e_{t+1}\le\sqrt{2\varepsilon_t/\kappa}
+(K_c/\kappa)|\lambda_t-\bar\lambda|
\le q e_{t-1}+\sqrt{2\varepsilon_t/\kappa}+(K_c/\kappa)\zeta_{t-1}.
\]
This proves~\eqref{eq:delayed-error-recursion}. Exact updates contract each of the even and odd policy subsequences by $q$, giving~\eqref{eq:exact-delayed-rate}. For bounded errors, let $a=\sqrt{2\varepsilon/\kappa}+(K_c/\kappa)\zeta$. Each subsequence satisfies a recurrence of the form $x_{j+1}\le qx_j+a$; summing the geometric series gives~\eqref{eq:inexact-error-floor}. If both errors vanish, applying the same argument to every tail gives $\limsup e_t=0$. The bound $|\lambda_{t+1}-\bar\lambda|\le\zeta_t+L_\Lambda e_t$ then proves multiplier convergence.
\end{proof}

\textbf{Interpretation of the contraction condition.}
The bound $q=bL_\Lambda<1$ controls the product of the actor and multiplier response gains. Similar conditions are used to establish stability of coupled learning systems. Guo et al.~\citep[Theorem~1]{guo2019learning} and Xie et al.~\citep[Lemma~1 and Theorem~1]{xie2021learning} use $d_1d_2+d_3<1$ for a composite policy--population response. In SDLA, $F_{\beta,\tau}(\lambda)=\Lambda_{\beta,\tau}(B(\lambda))$ depends on $\lambda$ through $B$, yielding the product bound $q=bL_\Lambda$. Mandal et al.~\citep[Theorem~1]{mandal2023performative} obtain a contraction factor $\mu<1$ by choosing regularization sufficiently strong relative to environment sensitivity. Olshevsky and Gharesifard~\citep{olshevsky2023small} use a small-gain argument to control coupled actor and critic errors.

For fixed problem parameters, $\alpha$, $\tau$, and $\lambda_{\max}$, the constants $K_c$, $K_m(\tau)$, and $\kappa>0$ are independent of $\beta$. For $K_c>0$, the explicit choice
\[
0<\beta<\frac{\kappa}{K_c[K_c+K_m(\tau)]}
\]
therefore ensures $q<1$, giving a nonempty sufficient range for stability of the response loop. If $K_c=0$, then $q=0$. Corollary~\ref{cor:concrete-actor-feasibility} combines the convergence condition with parameter matching to identify the equilibrium with the CMDP optimum.

\textbf{Sampling and scalar-solver errors.}
Let $\widehat\lambda_t$ minimize the empirical surrogate on the batch collected under $\pi_t$, and suppose the numerical solve satisfies $|\lambda_{t+1}-\widehat\lambda_t|\le\upsilon_t$. The derivative discrepancy between empirical and population surrogates is bounded by
\[
|\widehat c_t-J^c(\pi_t)|
+\sup_{\lambda\in[0,\lambda_{\max}]}|\widehat m_t(\lambda)-m_{\pi_t,\tau}(\lambda)|.
\]
Applying the constrained strong-monotonicity argument above and the triangle inequality gives the pathwise bound
\begin{equation}
|\lambda_{t+1}-\Lambda_{\beta,\tau}(\pi_t)|
\le\beta\left(
|\widehat c_t-J^c(\pi_t)|
+\sup_{\lambda\in[0,\lambda_{\max}]}|\widehat m_t(\lambda)-m_{\pi_t,\tau}(\lambda)|
\right)+\upsilon_t.
\label{eq:sample-dual-error}
\end{equation}
The right-hand side can therefore be used as $\zeta_t$ in Theorem~\ref{thm:parameter-convergence}. It separates trajectory-estimation error from numerical optimization error.

\subsection{Convergence with Concrete Actor Updates}
\label{app:concrete-actor}
We prove the convergence result under mild assumptions stated in Theorem~\ref{thm:concrete-actor-convergence}, using the same delayed updates and actor and dual error bounds as in Section~\ref{sec:concrete-actor}.

\subsubsection{Uniform constants for actor optimization}
For the updates in Section~\ref{sec:concrete-actor}, let $n=|\mathcal S||\mathcal A|$. The constants in~\eqref{eq:concrete-actor-error} can be chosen as
\begin{align}
D_Q&=\frac{R_{\max}+\lambda_{\max}C_{\max}+\alpha\log|\mathcal A|}{1-\gamma},
\label{eq:concrete-critic-range}\\
p_*&=\min\left\{\min_{s,a}\pi_0(a\mid s),
\frac{e^{-(D_Q+2\bar\xi)/\alpha}}{|\mathcal A|}\right\},
\label{eq:concrete-policy-support}\\
A_0&=\frac{2\sqrt n}{\alpha}
\left[D_Q+2\alpha(1-\omega)\log(1/p_*)\right],
\label{eq:concrete-actor-constant}\\
A_1&=\frac{4\sqrt n}{\alpha(1-\gamma)}
\left(1+\frac{\gamma}{(1-\gamma)\omega}\right).
\label{eq:concrete-critic-constant}
\end{align}
These constants depend on the fixed problem parameters, initialization, and uniform critic-error bound $\bar\xi$, and are independent of the outer iteration. We use $K_c,K_m(\tau),L_\Lambda$ from Appendix~\ref{app:parameter-convergence}, $\kappa=\alpha\rho_{\min}$, $b=K_c/\kappa$, and $q=bL_\Lambda$.

\subsubsection{Tabular Actor Updates and Approximation Bounds}
We now derive the actor approximation error from explicit policy updates. Work in the finite tabular setting of Section~\ref{sec:concrete-actor}, with $0<\gamma<1$, fixed $\alpha>0$, and an initial policy satisfying $\min_{s,a}\pi_0(a\mid s)>0$. At outer iteration $t$, hold $\lambda_t$ fixed and initialize $\pi_t^0=\pi_t$. The soft critic for the penalized reward satisfies
\begin{align}
Q_\lambda^\pi(s,a)&=r(s,a)-\lambda c(s,a)+\gamma\mathbb E_{s'\mid s,a}V_\lambda^\pi(s'),\nonumber\\
V_\lambda^\pi(s)&=\sum_a\pi(a\mid s)Q_\lambda^\pi(s,a)+\alpha\mathcal H(\pi(\cdot\mid s)),
\label{eq:concrete-soft-critic}
\end{align}
where $\mathcal H$ is single-state Shannon entropy. For $k=0,\ldots,K_t-1$, suppose $\|\widehat Q_t^k-Q_{\lambda_t}^{\pi_t^k}\|_\infty\le\xi_t$, with $0\le\xi_t\le\bar\xi<\infty$, and update
\begin{equation}
\pi_t^{k+1}(a\mid s)=
\frac{[\pi_t^k(a\mid s)]^{1-\omega}\exp(\omega\widehat Q_t^k(s,a)/\alpha)}
{\sum_b[\pi_t^k(b\mid s)]^{1-\omega}\exp(\omega\widehat Q_t^k(s,b)/\alpha)},
\qquad 0<\omega\le1.
\label{eq:concrete-actor-update}
\end{equation}
Set $\pi_{t+1}=\pi_t^{K_t}$, where $K_t\ge1$, and retain the delayed dual update~\eqref{eq:inexact-dual-update}, with error $\zeta_t$.

For $\omega=1$, this is soft policy improvement over the tabular simplex, which underlies SAC~\citep{haarnoja2018soft}. For $0<\omega<1$, it solves the statewise entropy-regularized KL-penalty problem
\begin{equation}
\max_{p\in\Delta(\mathcal A)}
\left\{\langle p,\widehat Q_t^k(s,\cdot)\rangle+\alpha\mathcal H(p)
-\frac{\alpha(1-\omega)}{\omega}
\operatorname{KL}\!\left(p\,\Vert\,\pi_t^k(\cdot\mid s)\right)\right\}.
\label{eq:concrete-kl-ppo}
\end{equation}
This KL-PPO variant uses the new-to-old KL direction studied by Liu et al.~\citep{liu2019neural}. Its exact tabular update coincides with entropy-regularized natural policy gradient at learning rate $\eta=(1-\gamma)\omega/\alpha$~\citep{cen2022fast}. This equivalence supplies the actor convergence rate below. The analyzed updates use a fixed entropy coefficient and the full tabular simplex; the PPO-Clip and neural SAC implementations in the experiments use function approximation.

Write $b=K_c/\kappa$, $L_\Lambda=L_\Lambda(\beta,\tau)$, and $\chi=1-(1-\gamma)\omega\in(0,1)$. The constants $A_0,A_1$ in~\eqref{eq:concrete-actor-constant}--\eqref{eq:concrete-critic-constant} are uniform over outer iterations and satisfy
\begin{equation}
\|\pi_{t+1}-B(\lambda_t)\|_F
\le A_0\chi^{K_t-1}+A_1\xi_t.
\label{eq:concrete-actor-error}
\end{equation}
The two terms quantify finite actor optimization and critic estimation error.

\subsubsection{Proof of Theorem~\ref{thm:concrete-actor-convergence}}
\begin{proof}
\textbf{Identification of the actor update.}
For $0<\omega<1$, the statewise objective in~\eqref{eq:concrete-kl-ppo} is strictly concave. Its first-order conditions, including the probability-normalization multiplier, give
\[
\log p(a)=(1-\omega)\log\pi_t^k(a\mid s)
+\frac{\omega}{\alpha}\widehat Q_t^k(s,a)+\text{constant}.
\]
Normalizing gives~\eqref{eq:concrete-actor-update}. At $\omega=1$, the same formula gives the Gibbs soft-policy improvement. The entropy-regularized natural policy gradient update of Cen et al.~\citep{cen2022fast} has old-policy exponent $1-\eta\alpha/(1-\gamma)$ and critic exponent $\eta/(1-\gamma)$. Setting $\eta=(1-\gamma)\omega/\alpha$ therefore gives exactly the specified actor update.

\textbf{Uniform support across warm starts.}
For every policy and $\lambda\in[0,\lambda_{\max}]$, the soft critic lies in the common interval
\[
\left[-\frac{\lambda_{\max}C_{\max}}{1-\gamma},
\frac{R_{\max}+\alpha\log|\mathcal A|}{1-\gamma}\right].
\]
Its width is $D_Q$. Since $\|\widehat Q_t^k-Q_{\lambda_t}^{\pi_t^k}\|_\infty\le\bar\xi$, the Gibbs distribution $u(a)\propto\exp(\widehat Q_t^k(s,a)/\alpha)$ has every coordinate at least $e^{-(D_Q+2\bar\xi)/\alpha}/|\mathcal A|$. The actor update is the normalized geometric mixture $p(a)^{1-\omega}u(a)^\omega$, where $p=\pi_t^k(\cdot\mid s)$. H\"older's inequality gives
\[
\sum_a p(a)^{1-\omega}u(a)^\omega
\le\left(\sum_a p(a)\right)^{1-\omega}
\left(\sum_a u(a)\right)^\omega=1.
\]
Thus $p(a),u(a)\ge p_*$ imply $\pi_t^{k+1}(a\mid s)\ge p_*$. Induction over inner and outer iterations, with $\pi_t^0=\pi_t$, proves a uniform lower bound $p_*$ for every iterate. The Gibbs optimal policy $B(\lambda)$ satisfies the same bound.

\textbf{Actor error from finite inner iterations.}
Theorems~1--2 of Cen et al.~\citep{cen2022fast}, specifically the log-policy error bound with approximate policy evaluation, give
\begin{equation}
\|\log\pi_{t+1}-\log B(\lambda_t)\|_\infty
\le\frac2\alpha\left[
C_{1,t}\chi^{K_t-1}
+\frac{2\xi_t}{1-\gamma}
\left(1+\frac{\gamma}{(1-\gamma)\omega}\right)\right],
\label{eq:concrete-cen-bound}
\end{equation}
where
\[
C_{1,t}=\|Q_{\lambda_t}^\star-Q_{\lambda_t}^{\pi_t}\|_\infty
+2\alpha(1-\omega)\|\log B(\lambda_t)-\log\pi_t\|_\infty.
\]
To apply the reward normalization in that result, choose $S=\max\{1,R_{\max}+\lambda_{\max}C_{\max}\}$ and transform the reward to $(r-\lambda c+\lambda_{\max}C_{\max})/S$, the entropy coefficient to $\alpha/S$, and critic errors to $\xi_t/S$. The transformed reward lies in $[0,1]$. The shift adds the same constant to all action values, and scaling the critic and entropy coefficient by $1/S$ preserves the policy update. Rescaling the bound gives~\eqref{eq:concrete-cen-bound}.

The common critic interval and uniform support bound imply
\[
C_{1,t}\le D_Q+2\alpha(1-\omega)\log(1/p_*).
\]
For $x,y\in(0,1]$, $|x-y|\le|\log x-\log y|$. Hence the Frobenius policy error is at most $\sqrt n$ times the log-policy error in the supremum norm. Substituting the preceding bound into~\eqref{eq:concrete-cen-bound} proves~\eqref{eq:concrete-actor-error} with the constants in~\eqref{eq:concrete-actor-constant}--\eqref{eq:concrete-critic-constant}.

\textbf{Coupled convergence.}
Appendix~\ref{app:parameter-convergence} establishes
\[
\|B(\lambda)-B(\lambda')\|_F\le b|\lambda-\lambda'|,
\qquad
|\Lambda_{\beta,\tau}(\pi)-\Lambda_{\beta,\tau}(\pi')|
\le L_\Lambda\|\pi-\pi'\|_F.
\]
Thus $\Lambda_{\beta,\tau}\circ B$ is a contraction on $[0,\lambda_{\max}]$ with factor $q<1$ and has a unique fixed point $\bar\lambda$. Set $\bar\pi=B(\bar\lambda)$. Combining the actor error bound with the delayed dual update gives, for $t\ge1$,
\begin{equation}
\begin{aligned}
e_{t+1}
&\le A_0\chi^{K_t-1}+A_1\xi_t+b|\lambda_t-\bar\lambda|\\
&\le q e_{t-1}+A_0\chi^{K_t-1}+A_1\xi_t+b\zeta_{t-1}.
\end{aligned}
\label{eq:concrete-delayed-recursion}
\end{equation}
Write $h_t=A_0\chi^{K_t-1}+A_1\xi_t+b\zeta_{t-1}$. Applying the recursion separately to even and odd subsequences yields
\[
e_{2j}\le q^j e_0+\sum_{i=1}^j q^{j-i}h_{2i-1},
\qquad
e_{2j+1}\le q^j e_1+\sum_{i=1}^j q^{j-i}h_{2i}.
\]
If $K_t\to\infty$ and both estimation errors vanish, then $h_t\to0$. Splitting each sum into a finite prefix and a uniformly small tail shows that $e_t\to0$. If $h_t\le A_0\chi^{K-1}+A_1\xi+b\zeta$, summing the geometric series gives the policy bound in~\eqref{eq:concrete-error-floor}.

Under the geometric-error assumption, the shifted error satisfies $\zeta_{t-1}=O(r^t)$ because $r>0$ is fixed, so $h_t=O(r^t)$. The two parity sums are bounded by constant multiples of
\[
\sum_{i=1}^j q^{j-i}r^{2i}
=\begin{cases}
\displaystyle\frac{r^2(q^j-r^{2j})}{q-r^2},&q\ne r^2,\\[4pt]
jq^j,&q=r^2.
\end{cases}
\]
Together with the initial-error terms, this gives $e_t=O((\max\{\sqrt{q},r\})^t)$ when $r\ne\sqrt{q}$, and $e_t=O((t+1)(\sqrt{q})^t)$ when $r=\sqrt{q}$. Both bounds imply $e_t=O(\nu^t)$ for every $\max\{\sqrt{q},r\}<\nu<1$. Finally,
\[
|\lambda_t-\bar\lambda|\le\zeta_{t-1}+L_\Lambda e_{t-1}
\]
gives multiplier convergence, its asymptotic error bound, and~\eqref{eq:concrete-linear-rate}.
\end{proof}

\subsubsection{Fixed-Parameter Feasibility and Reward Bounds}
\label{app:fixed-parameter-bounds}
We prove Proposition~\ref{prop:fixed-parameter-bounds} for any fixed $\beta,\tau,L>0$ and any equilibrium satisfying exact actor and population-surrogate optimality. The proof uses bounded nonnegative costs, a strictly feasible reference policy, and actor best responses, and holds independently of the contraction condition. In the finite discounted setting, a valid objective bound is
\[
U=\frac{R_{\max}+\alpha\log|\mathcal A|}{1-\gamma}.
\]
A feasible reference policy and the problem bounds determine the value of this bound.

\begin{proof}
Write $c=J^c(\bar\pi)$, $v=c-d$, $m=m_{\bar\pi,\tau}(\bar\lambda)$, and $J=J_\alpha(\bar\pi)$. Nonnegative bounded costs give $0\le m\le\bar C$. Actor optimality against the strictly feasible policy yields
\[
J-\bar\lambda(c-d)
\ge J_\alpha(\pi_s)-\bar\lambda[J^c(\pi_s)-d]
\ge J_\alpha(\pi_s)+\bar\lambda\sigma.
\]
Since $J\le U$, it follows that
\begin{equation}
\bar\lambda(v+\sigma)\le J-J_\alpha(\pi_s)\le\Delta.
\label{eq:fixed-slater-comparison}
\end{equation}
The population surrogate derivative at the equilibrium is $g'(\bar\lambda)=-v-m+\bar\lambda/\beta$. Its constrained optimality conditions are
\begin{equation}
\begin{cases}
-v-m\ge0,&\bar\lambda=0,\\
-v-m+\bar\lambda/\beta=0,&0<\bar\lambda<L,\\
-v-m+L/\beta\le0,&\bar\lambda=L.
\end{cases}
\label{eq:fixed-endpoint-conditions}
\end{equation}

\textbf{Feasibility.}
If $\bar\lambda=0$, then $v\le-m\le0$. If $0<\bar\lambda<L$ and $v>0$, the interior condition gives
\[
v=\bar\lambda/\beta-m\le\bar\lambda/\beta.
\]
Combining this inequality with~\eqref{eq:fixed-slater-comparison} yields
\[
\beta v(v+\sigma)\le\bar\lambda(v+\sigma)\le\Delta,
\qquad
v\le\frac{\sqrt{\sigma^2+4\Delta/\beta}-\sigma}{2}.
\]
If $\bar\lambda=L$, equation~\eqref{eq:fixed-slater-comparison} gives $v\le\Delta/L-\sigma$. Taking positive parts and combining the three cases proves~\eqref{eq:fixed-feasibility-bound}. When $L\ge\Delta/\sigma$, the saturated-case violation bound is zero. The nonnegative root $x$ of $x(x+\sigma)=\Delta/\beta$ satisfies $x\sigma\le\Delta/\beta$, giving $V_{\beta,L}\le\Delta/(\beta\sigma)$.

\textbf{Regularized reward.}
Let $\pi^\star$ maximize $J_\alpha$ over feasible policies; this maximum exists by compactness and continuity. Actor optimality and $J^c(\pi^\star)\le d$ imply
\[
J-\bar\lambda(c-d)
\ge J_\alpha^\star(d)-\bar\lambda[J^c(\pi^\star)-d]
\ge J_\alpha^\star(d).
\]
Hence $J_\alpha^\star(d)-J\le\bar\lambda(d-c)$. When $\bar\lambda=0$, this gap is nonpositive. For an interior multiplier, $d-c=m-\bar\lambda/\beta$; at the upper endpoint, $g'(L)\le0$ gives $d-c\le m-L/\beta$. In both cases,
\begin{align*}
J_\alpha^\star(d)-J
&\le\bar\lambda m-\bar\lambda^2/\beta\\
&\le\bar\lambda\bar C-\bar\lambda^2/\beta\\
&=\frac{\beta\bar C^2}{4}
-\frac1\beta\left(\bar\lambda-\frac{\beta\bar C}{2}\right)^2
\le\frac{\beta\bar C^2}{4}.
\end{align*}
The right-hand side is nonnegative, so taking the positive part proves~\eqref{eq:fixed-reward-bound}. The violation bound uses strict feasibility; the reward bound follows from feasibility of the comparison optimum and the two best-response conditions.
\end{proof}

\textbf{Iteration errors.}
Under Theorem~\ref{thm:concrete-actor-convergence} and the strict-feasibility assumption of Proposition~\ref{prop:fixed-parameter-bounds}, the unique equilibrium obeys these bounds. Since the positive-part map is 1-Lipschitz,
\begin{equation}
[J^c(\pi_t)-d]_+
\le[J^c(\bar\pi)-d]_++K_c\|\pi_t-\bar\pi\|_F
\le V_{\beta,L}+K_c e_t.
\label{eq:fixed-iterate-feasibility}
\end{equation}
Consequently, the uniform error conditions of that theorem imply
\begin{equation}
\limsup_{t\to\infty}[J^c(\pi_t)-d]_+
\le V_{\beta,L}+K_c\mathcal E_{K,\xi,\zeta}.
\label{eq:fixed-feasibility-floor}
\end{equation}
The bound combines the equilibrium term $V_{\beta,L}$ with actor, critic, and dual approximation errors. If these errors vanish as specified in Theorem~\ref{thm:concrete-actor-convergence}, the asymptotic feasibility bound reduces to $V_{\beta,L}$, and continuity of $J_\alpha$ gives
\[
\limsup_{t\to\infty}[J_\alpha^\star(d)-J_\alpha(\pi_t)]_+
\le\beta\bar C^2/4.
\]

\textbf{Parameter interpretation.}
An interior equilibrium satisfies $J^c(\bar\pi)-d=\bar\lambda/\beta-m$. Increasing $\beta$ tightens the interior violation bound and increases the reward bound and response gain $q=b\beta(K_c+K_m)$. Proposition~\ref{prop:fixed-parameter-bounds} applies to any exact equilibrium with positive fixed parameters; the iterative bounds additionally use $q<1$ and the approximation assumptions of Theorem~\ref{thm:concrete-actor-convergence}.

Numerical evaluation uses the strict-feasibility margin $\sigma$, objective bound $U$, and tabular response constants. The neural experiments evaluate $\beta=0.1$, $\tau=1$, and $L=10$ empirically. Corollary~\ref{cor:concrete-actor-feasibility} identifies the parameter matching that recovers an entropy-regularized CMDP saddle point.

\subsubsection{Proof of Corollary~\ref{cor:concrete-actor-feasibility}}
\begin{proof}
Since costs are nonnegative and $J^c(\pi_\alpha^\star)=d>0$, the weighted cost $m_{\pi_\alpha^\star,\tau}(\lambda_\alpha^\star)$ is positive. Parameter matching~\eqref{eq:optimality-parameter-curve} gives
\[
g_{\pi_\alpha^\star,\beta,\tau}'(\lambda_\alpha^\star)
=-m_{\pi_\alpha^\star,\tau}(\lambda_\alpha^\star)
+\lambda_\alpha^\star/\beta=0.
\]
Strong convexity and the interior multiplier yield $\Lambda_{\beta,\tau}(\pi_\alpha^\star)=\lambda_\alpha^\star$. The saddle-point condition gives $B(\lambda_\alpha^\star)=\pi_\alpha^\star$. Uniqueness in Theorem~\ref{thm:concrete-actor-convergence} identifies the equilibrium with this saddle point. The cost Lipschitz bound then gives
\[
[J^c(\pi_t)-d]_+
\le|J^c(\pi_t)-J^c(\pi_\alpha^\star)|
\le K_c e_t.
\]
The convergence and error-floor conclusions follow from Theorem~\ref{thm:concrete-actor-convergence}. Continuity of $J_\alpha$ gives convergence of the entropy-regularized objective when $e_t\to0$.
\end{proof}

\clearpage
\section{PPO and SAC Implementations}
\label{app:theory-practice}
Related safe RL methods also pair analyses under explicit structural assumptions with neural implementations. Chen et al.~\citep{chen2024adaptive} assume Lipschitz gradients and strong convexity of the Lagrangian in actor parameters, and evaluate PPO-Lagrangian and DDPG-Lagrangian variants on Bullet-Safety-Gym. CVPO~\citep{liu2022constrained} establishes optimality of its nonparametric variational E-step and fits a neural policy in the M-step. CRPO~\citep{xu2021crpo} analyzes both tabular policies and, under additional assumptions, two-layer neural-network approximation.

Algorithm~\ref{alg:general-convex} gives the general framework; Algorithms~\ref{alg:sdla-ppo} and~\ref{alg:sdla-sac} give its PPO-SDLA and SAC-SDLA implementations.

\begin{algorithm}[htbp]
\caption{SDLA (General Framework)}
\label{alg:general-convex}
\begin{algorithmic}[1]
\STATE \textbf{Input:} Base RL algorithm, hyperparameters $\tau,\beta,d,\lambda_{\max}$
\STATE \textbf{Initialize:} $\pi_\theta$, critics $Q^r_{\phi_r},Q^c_{\phi_c}$, $\lambda=0$
\FOR{each iteration}
\STATE Run $\pi_\theta$, collect $\mathcal T$, compute $\{R_i,C_i\}_{i=1}^M$ and $\widehat c_M$
\STATE $Q^r_{\phi_r},Q^c_{\phi_c}\leftarrow\mathrm{UpdateCritics}(\mathcal T)$
\STATE $\pi_\theta\leftarrow\mathrm{UpdateActor}(Q^r_{\phi_r}-\lambda Q^c_{\phi_c})$
\STATE \textit{// Dual update with the fixed trajectory batch}
\STATE Define $\widehat F_M$ by~\eqref{eq:trajectory-objective}
\STATE Define $\widehat g_M$ by~\eqref{eq:surrogate-dual}
\STATE $\lambda\leftarrow\mathrm{TernarySearch}_{15}(\widehat g_M,[0,\lambda_{\max}])$
\ENDFOR
\end{algorithmic}
\end{algorithm}

\begin{algorithm}[htbp]
\caption{PPO-SDLA}
\label{alg:sdla-ppo}
\begin{algorithmic}[1]
\STATE Initialize policy \(\pi_\theta\), reward value \(V^r_{\phi_r}\), cost value \(V^c_{\phi_c}\), and dual variable \(\lambda=0\).
\STATE Hyperparameters: cost limit \(d\), dual temperature \(\tau\), regularization \(\beta\), dual cap \(\lambda_{\max}\).
\FOR{each iteration}
    \STATE Run the current policy \(\pi_\theta\) for \(M\) trajectories and collect
    \[
    \mathcal T=\{\xi_i\}_{i=1}^{M}.
    \]
    \STATE Compute trajectory-wise discounted reward and cost returns:
    \[
    R_i=\sum_{t\ge0}\gamma^t r(s_t^i,a_t^i),
    \qquad
    C_i=\sum_{t\ge0}\gamma^t c(s_t^i,a_t^i).
    \]
    \STATE Estimate current policy cost:
    \[
    \widehat c_M=\frac{1}{M}\sum_{i=1}^{M}C_i.
    \]

    \STATE Compute reward and cost advantages \(\hat A^r_t,\hat A^c_t\) using GAE.
    \STATE Update \(V^r_{\phi_r},V^c_{\phi_c}\) by value regression.
    \STATE Update actor by PPO using the Lagrangian advantage
    \[
    \hat A_t^\lambda=\hat A^r_t-\lambda \hat A^c_t.
    \]

    \STATE \textbf{Dual update.} For any candidate \(\lambda'\in[0,\lambda_{\max}]\), compute
    \[
    \widehat F_M(\lambda')
    =
    \tau\log
    \left(
    \frac{1}{M}
    \sum_{i=1}^{M}
    \exp\left(
    \frac{R_i-\lambda' C_i}{\tau}
    \right)
    \right).
    \]
    \STATE Define the trajectory-wise empirical surrogate dual
    \[
    \widehat g_M(\lambda')
    =
    \widehat F_M(\lambda')
    +
    \lambda'
    \bigl(d-\widehat c_M\bigr)
    +
    \frac{(\lambda')^2}{2\beta}.
    \]
    \STATE Update dual variable by $K=15$ ternary-search iterations:
    \[
    \lambda
    \leftarrow
    \mathrm{TernarySearch}_{15}(\widehat g_M,[0,\lambda_{\max}]).
    \]
\ENDFOR
\STATE Output \(\pi_\theta\).
\end{algorithmic}
\end{algorithm}

\begin{algorithm}[p]
\caption{SAC-SDLA}
\label{alg:sdla-sac}
\begin{algorithmic}[1]
\STATE \textbf{Initialize:} Policy \(\pi_\theta\), Q-functions \(Q^r_{\phi_{r,1}},Q^r_{\phi_{r,2}},Q^c_{\phi_{c,1}},Q^c_{\phi_{c,2}}\), target networks \(\bar Q^r_{\bar\phi_{r,1}},\bar Q^r_{\bar\phi_{r,2}},\bar Q^c_{\bar\phi_{c,1}},\bar Q^c_{\bar\phi_{c,2}}\), replay buffer \(\mathcal B\), \(\lambda=0\).
\STATE \textbf{Hyperparameters:} dual temperature \(\tau\), regularization \(\beta\), cost limit \(d\), dual cap \(\lambda_{\max}\), SAC temperature \(\alpha\), target update rate \(\tau_{\rm polyak}\).
\FOR{each iteration}
    \STATE Run \(\pi_\theta\) in the environment, add transitions \((s,a,r,c,s')\) to \(\mathcal B\).
    \STATE From recent rollout/evaluation episodes, compute trajectory-wise discounted returns
    \[
    R_i=\sum_{t\ge0}\gamma^t r_t^i,
    \qquad
    C_i=\sum_{t\ge0}\gamma^t c_t^i,
    \qquad i=1,\ldots,M.
    \]
    \STATE Estimate current policy cost:
    \[
    \widehat c_M
    =
    \frac{1}{M}\sum_{i=1}^{M}C_i.
    \]

    \STATE \textit{// SAC critic and actor updates}
    \FOR{gradient step \(g=1,\ldots,N_{\mathrm{grad}}\)}
        \STATE Sample mini-batch \(\{(s_i,a_i,r_i,c_i,s'_i)\}_{i=1}^{B}\) from \(\mathcal B\).
        \STATE Sample next actions \(a'_i\sim\pi_\theta(\cdot|s'_i)\).
        \STATE
        \[
        y^r_i
        \leftarrow
        r_i+\gamma
        \left[
        \min_{j=1,2}\bar Q^r_{\bar\phi_{r,j}}(s'_i,a'_i)
        -
        \alpha\log\pi_\theta(a'_i|s'_i)
        \right].
        \]
        \STATE
        \[
        y^c_i
        \leftarrow
        c_i+\gamma
        \min_{j=1,2}\bar Q^c_{\bar\phi_{c,j}}(s'_i,a'_i).
        \]
        \STATE Update \(Q^r_{\phi_{r,1}},Q^r_{\phi_{r,2}},Q^c_{\phi_{c,1}},Q^c_{\phi_{c,2}}\) by minimizing TD losses.
        \STATE Sample actions \(\tilde a_i\sim\pi_\theta(\cdot|s_i)\).
        \STATE Define
        \[
        \hat Q^r(s_i,\tilde a_i)=\min_{j=1,2}Q^r_{\phi_{r,j}}(s_i,\tilde a_i),
        \qquad
        \hat Q^c(s_i,\tilde a_i)=\min_{j=1,2}Q^c_{\phi_{c,j}}(s_i,\tilde a_i).
        \]
        \STATE Update actor by minimizing
        \[
        L_\pi
        =
        \frac{1}{B}
        \sum_{i=1}^{B}
        \left[
        \alpha\log\pi_\theta(\tilde a_i|s_i)
        -
        \hat Q^r(s_i,\tilde a_i)
        +
        \lambda\hat Q^c(s_i,\tilde a_i)
        \right].
        \]
        \STATE Update target networks by Polyak averaging.
    \ENDFOR

    \STATE \textit{// Dual update via trajectory-wise empirical surrogate}
    \STATE For any candidate \(\lambda'\in[0,\lambda_{\max}]\), compute
    \[
    \widehat F_M(\lambda')
    =
    \tau\log
    \left(
    \frac{1}{M}
    \sum_{i=1}^{M}
    \exp\left(
    \frac{R_i-\lambda' C_i}{\tau}
    \right)
    \right).
    \]
    \STATE Define the trajectory-wise empirical surrogate dual
    \[
    \widehat g_M(\lambda')
    =
    \widehat F_M(\lambda')
    +
    \lambda'
    \bigl(d-\widehat c_M\bigr)
    +
    \frac{(\lambda')^2}{2\beta}.
    \]
    \STATE Update dual variable by $K=15$ ternary-search iterations:
    \[
    \lambda
    \leftarrow
    \mathrm{TernarySearch}_{15}(\widehat g_M,[0,\lambda_{\max}]).
    \]
\ENDFOR
\STATE \textbf{Output:} Policy \(\pi_\theta\).
\end{algorithmic}
\end{algorithm}

\subsection{Scalar Solver Complexity}
\label{app:scalar-solver}
The empirical surrogate $\widehat g_M$ is strictly convex on $[0,\lambda_{\max}]$, admitting a unique minimum. For single-constraint CMDPs, ternary search computes the multiplier without a dual learning rate.
Each ternary-search iteration retains an interval containing the minimizer and reduces its length by a factor $2/3$. Starting from $[0,\lambda_{\max}]$, after $K$ iterations its width is $\lambda_{\max}(2/3)^K$. Thus $K=\lceil\log(\lambda_{\max}/\varepsilon_\lambda)/\log(3/2)\rceil$ iterations suffice for multiplier error at most $\varepsilon_\lambda$, for $0<\varepsilon_\lambda<\lambda_{\max}$, using two objective evaluations per iteration. We use $K=15$ in practice, giving a final interval width of $\lambda_{\max}(2/3)^{15}$.
Each empirical objective evaluation costs $O(M)$ for a fixed batch of $M$ trajectories, giving $O(M\log(\lambda_{\max}/\varepsilon_\lambda))$ arithmetic operations. In practice, log-sum-exp is evaluated after subtracting the largest trajectory score to avoid numerical overflow.

\clearpage
\section{Extra Experiments}
\label{sec:extra_exp}
HASARD results use five SDLA runs per setting and three archived runs per baseline, except PPOCost on Volcanic Venture L1, for which the archive contains one run. The HASARD learning curves below show one standard deviation across the available runs for each method.

\subsection{Regularization Ablation}
\label{app:regularization-ablation}
\textbf{Component objectives.} Equations~\eqref{eq:ablation-entropy}--\eqref{eq:ablation-qr-half} define the three PPO ablations using the same empirical trajectory objective $\widehat F_M$ as SDLA. Their derivatives are
\[
\widehat g_{\mathrm E}'=d-\widehat m(\lambda),\qquad
\widehat g_{\mathrm{QR}}'=d-\widehat m(\lambda)+\lambda/\beta,\qquad
\widehat g_{\mathrm{QR}\text{-}d/2}'=d/2-\widehat m(\lambda)+\lambda/\beta.
\]
SDLA has derivative $d-\widehat c_M-\widehat m(\lambda)+\lambda/\beta$. Thus PPO-QR here retains trajectory log-sum-exp weighting; the quadratic-regularized linear multiplier objective discussed in Section~\ref{sec:preliminary} uses the mean cost alone.

The fixed-center and adaptive-center configurations also differ in multiplier updates. Define $|\Delta\lambda_t|=|\lambda_{t+1}-\lambda_t|$ between consecutive logged values. Relative to PPO-QR-$d/2$, PPO-SDLA changes the mean update from $0.0849$ to $0.0865$, an increase of $1.9\%$, while reducing its 95th percentile from $0.4219$ to $0.3576$, a reduction of $15.2\%$.

\begin{figure}[htbp]
\centering
\includegraphics[width=\textwidth]{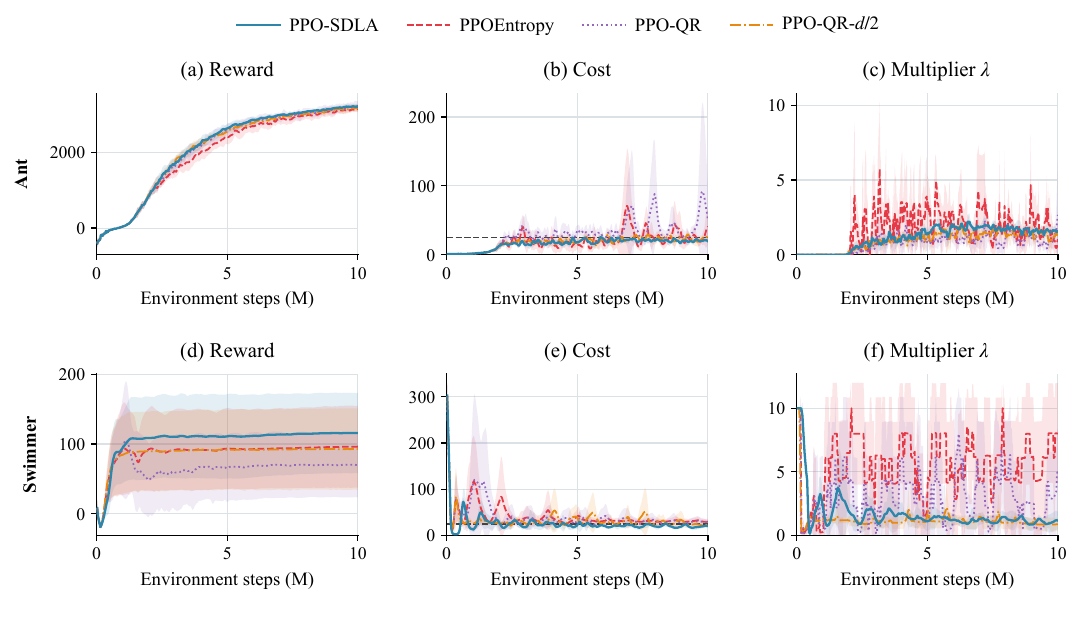}
\caption{Reward, cost, and logged multiplier trajectories on Ant (top) and Swimmer (bottom). Lines show means over five distinct seeds per method and environment; shading shows one standard deviation, clipped at zero for cost and $\lambda$. Curves use all 500 logged points without temporal smoothing. Dashed cost lines mark $d=25$. Swimmer uses $\tau=1$, with $\beta=10$ for PPO-QR, $\beta=0.1$ for SDLA, and $\beta=0.2$ for PPO-QR-$d/2$.}
\label{fig:dual_ablation_summary}
\end{figure}

Figure~\ref{fig:dual_ablation_summary} shows reward, cost, and multiplier trajectories on Ant and Swimmer. The Ant runs reproduce main Table~\ref{tab:dual_effectiveness}; the Swimmer SDLA runs reproduce its reward and cost in main Table~\ref{tab:mujoco_velocity}.

\textbf{Regularization sweep.}
Table~\ref{tab:ablation_safety_ant} reports the full regularization sweep. The default $\beta=0.1$ gives the smallest mean multiplier change, while all tested settings have final mean costs below the budget.

\begin{table}[htbp]
\centering
\small
\caption{Regularization ablation on SafetyAntVelocity-v1 with $d=25$. The $\beta$ sweep uses a 10M-step horizon and $\tau=1$; $\beta=\infty$ removes quadratic regularization. Mean $|\Delta\lambda|$ measures the average change between consecutive logged multipliers.}
\label{tab:ablation_safety_ant}
\setlength{\tabcolsep}{5pt}
\begin{tabular}{ccccc}
\toprule
$\beta$ & $\tau$ & Reward ($\uparrow$) & Cost ($\downarrow$) & Mean $|\Delta\lambda|$ \\
\midrule
0.1 & 1 & $3200.44\pm75.58$ & $19.73\pm2.46$ & $0.0865\pm0.0162$ \\
0.5 & 1 & $3132.15\pm115.28$ & $12.44\pm3.12$ & $0.3587\pm0.0250$ \\
1.0 & 1 & $3088.90\pm97.37$ & $15.31\pm3.31$ & $0.4969\pm0.0598$ \\
2.0 & 1 & $3095.12\pm129.66$ & $10.49\pm4.61$ & $0.5589\pm0.0671$ \\
5.0 & 1 & $3107.92\pm50.84$ & $11.65\pm5.40$ & $0.7639\pm0.1144$ \\
$\infty$ & 1 & $3086.35\pm93.44$ & $12.33\pm5.18$ & $0.8037\pm0.1017$ \\
\bottomrule
\end{tabular}
\end{table}

\subsection{Temperature Ablation}
\label{app:temperature-ablation}
Table~\ref{tab:temperature_ablation} compares the default $\tau=1$ with $\tau=1.5$ and $\tau=2$ on SafetyAntVelocity-v1. All three settings have mean costs below the budget of $25$. The default gives the highest reported mean reward, with $\tau=1.5$ close in mean reward. Mean $|\Delta\lambda|$ is $0.0865$, $0.08649$, and $0.08890$, respectively.

\begin{table}[htbp]
\centering
\small
\caption{Temperature ablation on SafetyAntVelocity-v1 with $d=25$ and the default $\beta=0.1$. Entries report mean $\pm$ standard deviation. Mean $|\Delta\lambda|$ reports the average absolute change between consecutive logged multipliers. Bold marks the highest mean reward.}
\label{tab:temperature_ablation}
\setlength{\tabcolsep}{5pt}
\begin{tabular}{cccc}
\toprule
$\tau$ & Reward ($\uparrow$) & Cost ($\downarrow$) & Mean $|\Delta\lambda|$ \\
\midrule
1 & \textbf{3200.44 $\pm$ 75.58} & $19.73\pm2.46$ & $0.0865\pm0.0162$ \\
1.5 & $3192.72\pm62.06$ & $20.24\pm1.36$ & $0.08649\pm0.01625$ \\
2 & $3098.98\pm61.66$ & $20.34\pm1.07$ & $0.08890\pm0.00939$ \\
\bottomrule
\end{tabular}
\end{table}

\subsection{Learning Curves on Safety-Gym Velocity Tasks}
\begin{figure*}[t]
\centering
\captionsetup[subfigure]{justification=centering}

\subfloat[Ant]{\includegraphics[width=0.19\textwidth]{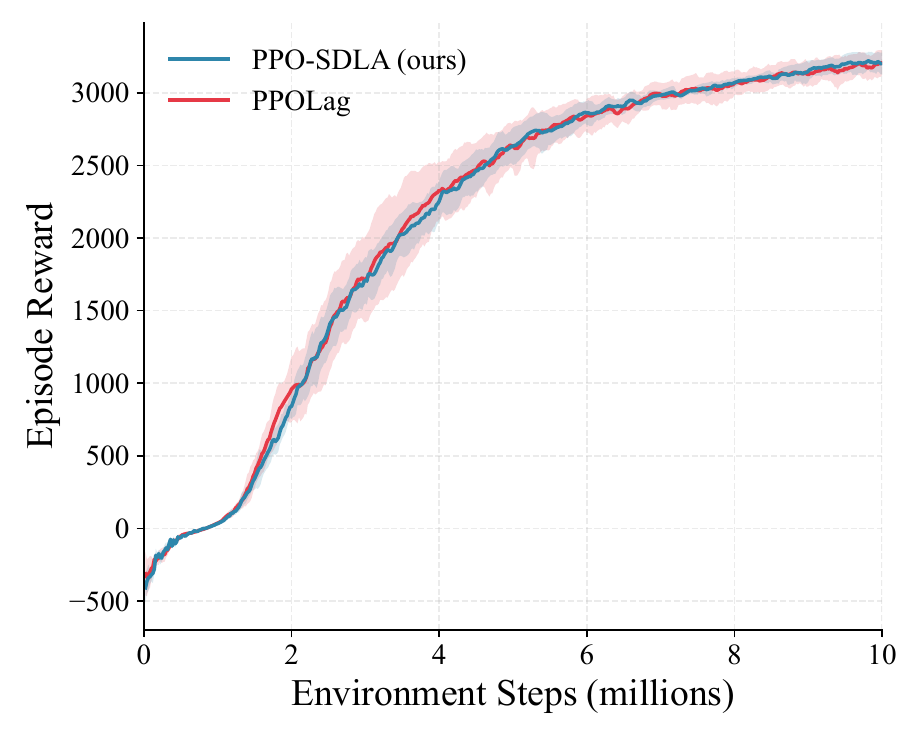}}
\hfill
\subfloat[HalfCheetah]{\includegraphics[width=0.19\textwidth]{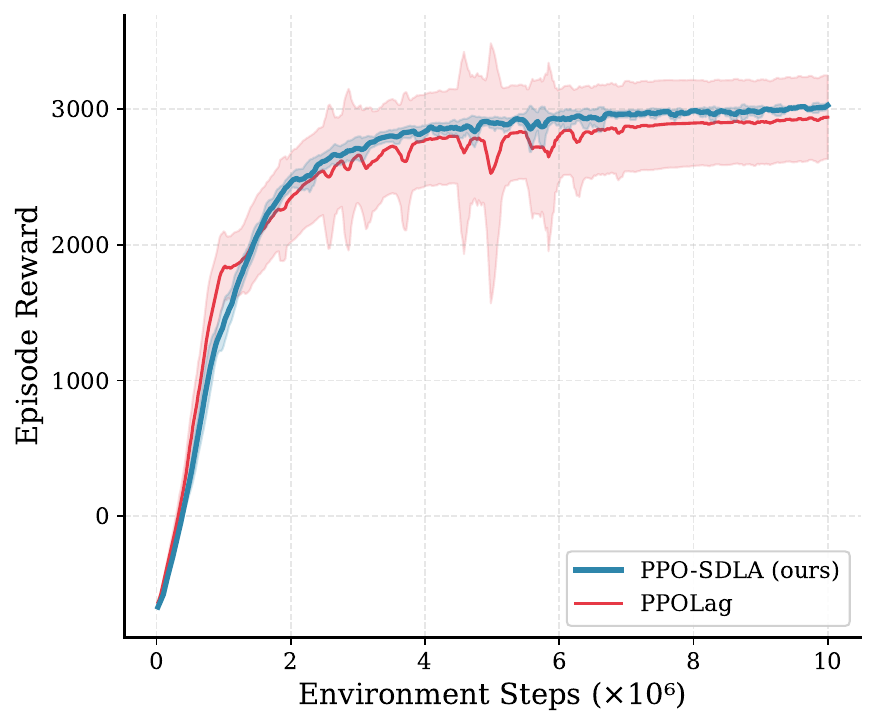}}
\hfill
\subfloat[Hopper]{\includegraphics[width=0.19\textwidth]{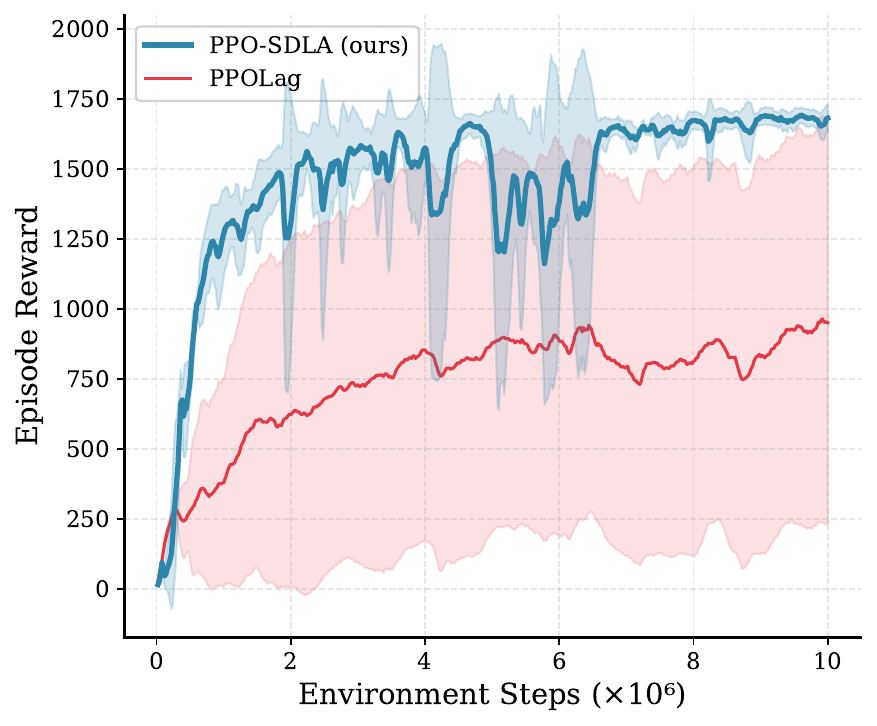}}
\hfill
\subfloat[Swimmer]{\includegraphics[width=0.19\textwidth]{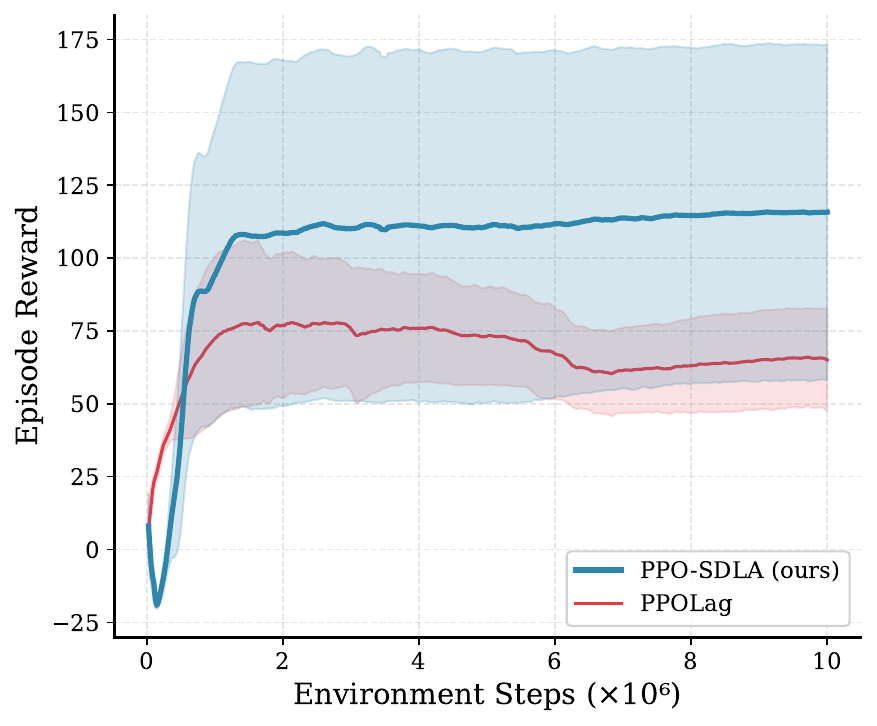}}
\hfill
\subfloat[Walker2d]{\includegraphics[width=0.19\textwidth]{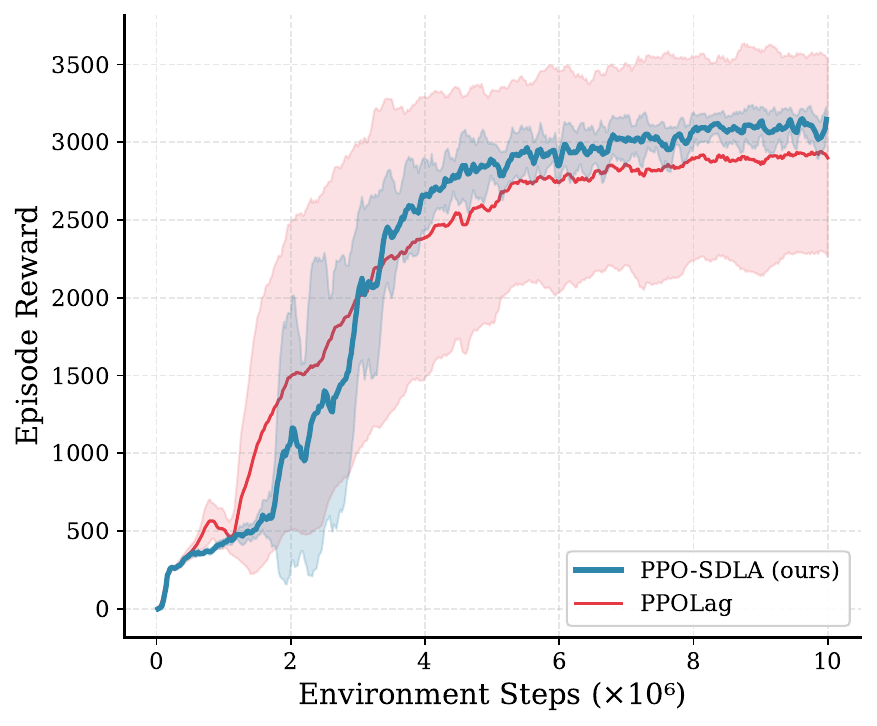}}

\vspace{0.2em}

\subfloat[Ant]{\includegraphics[width=0.19\textwidth]{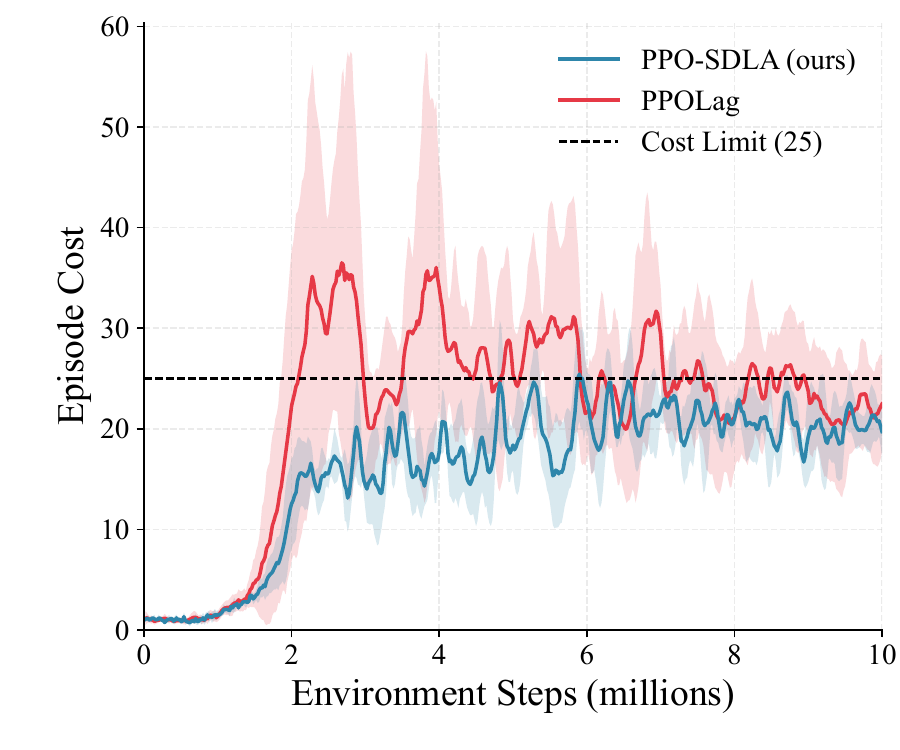}}
\hfill
\subfloat[HalfCheetah]{\includegraphics[width=0.19\textwidth]{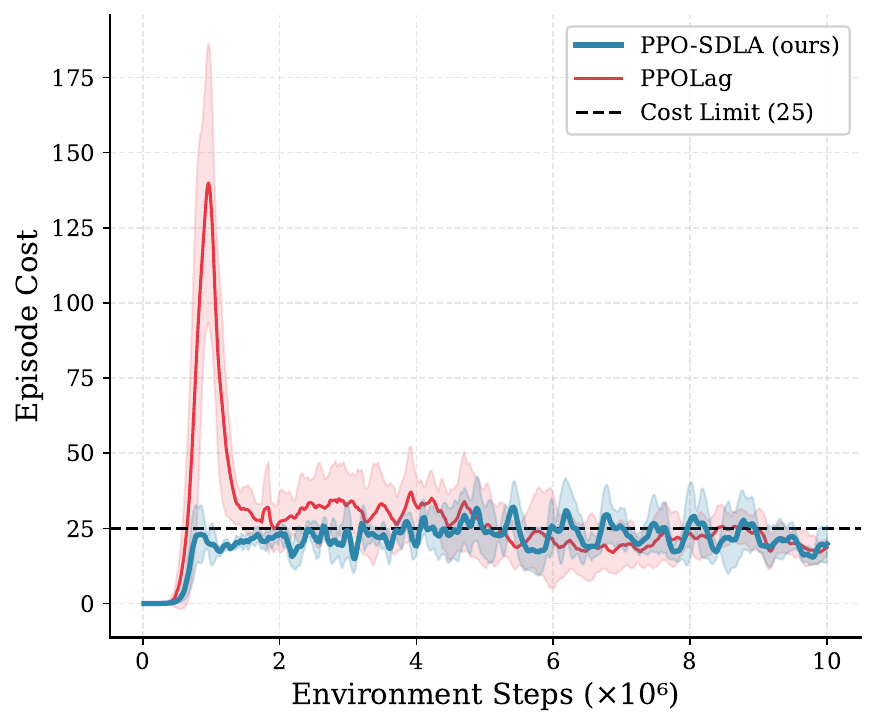}}
\hfill
\subfloat[Hopper]{\includegraphics[width=0.19\textwidth]{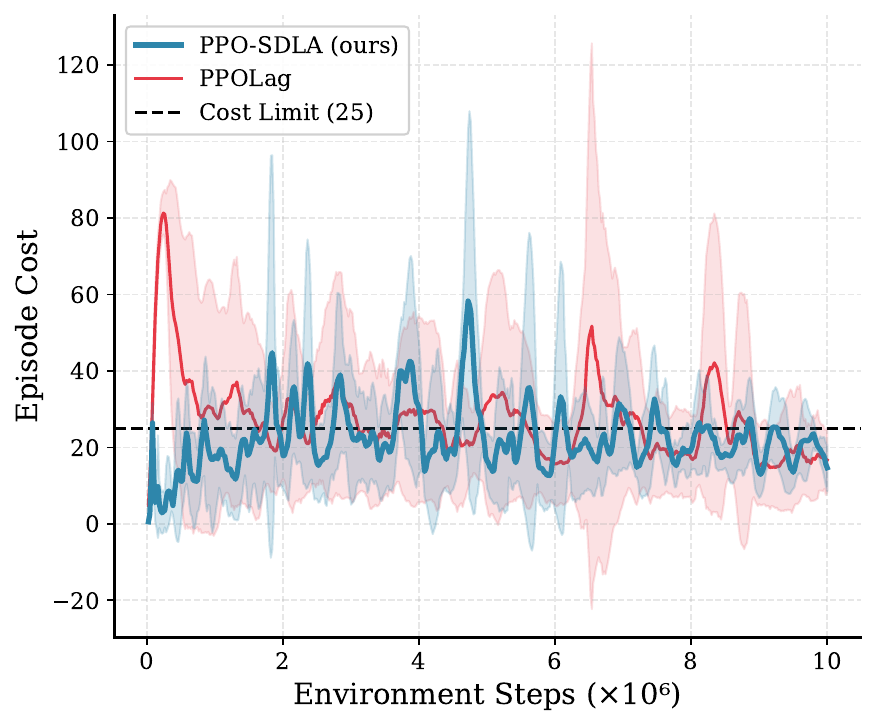}}
\hfill
\subfloat[Swimmer]{\includegraphics[width=0.19\textwidth]{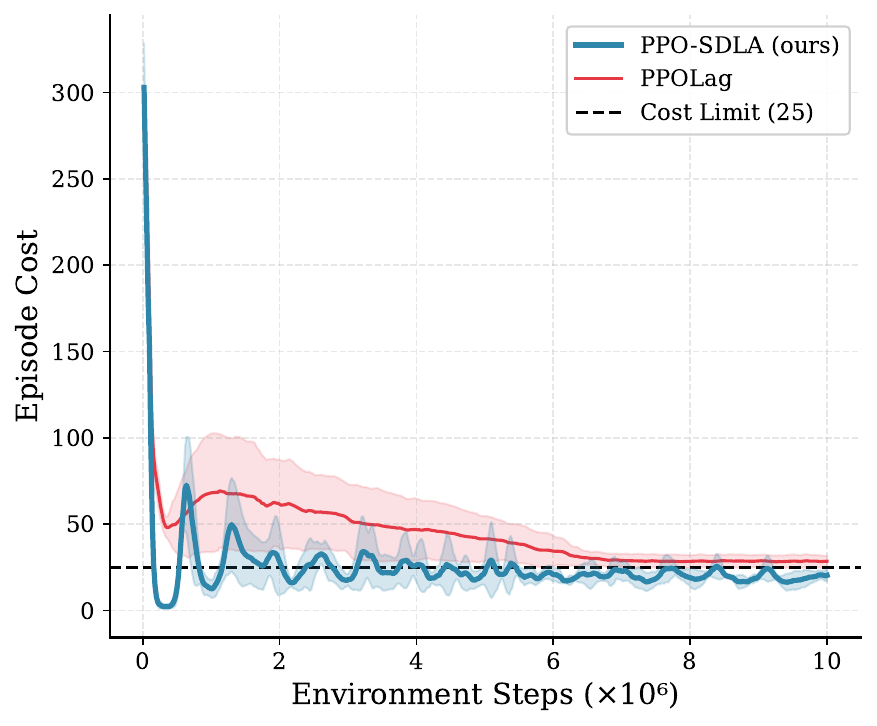}}
\hfill
\subfloat[Walker2d]{\includegraphics[width=0.19\textwidth]{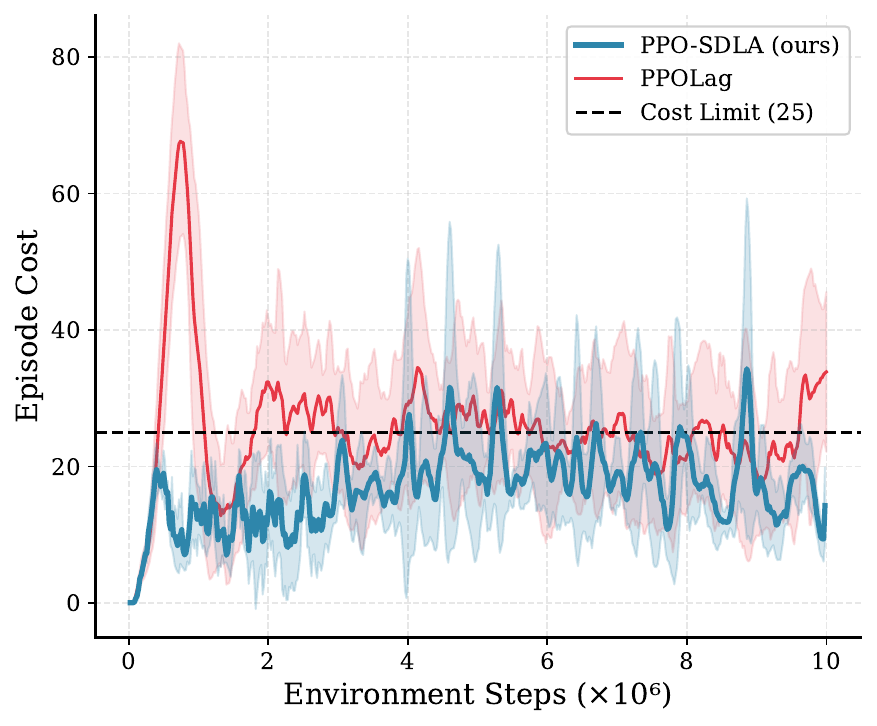}}

\vspace{0.2em}

\subfloat[Ant]{\includegraphics[width=0.19\textwidth]{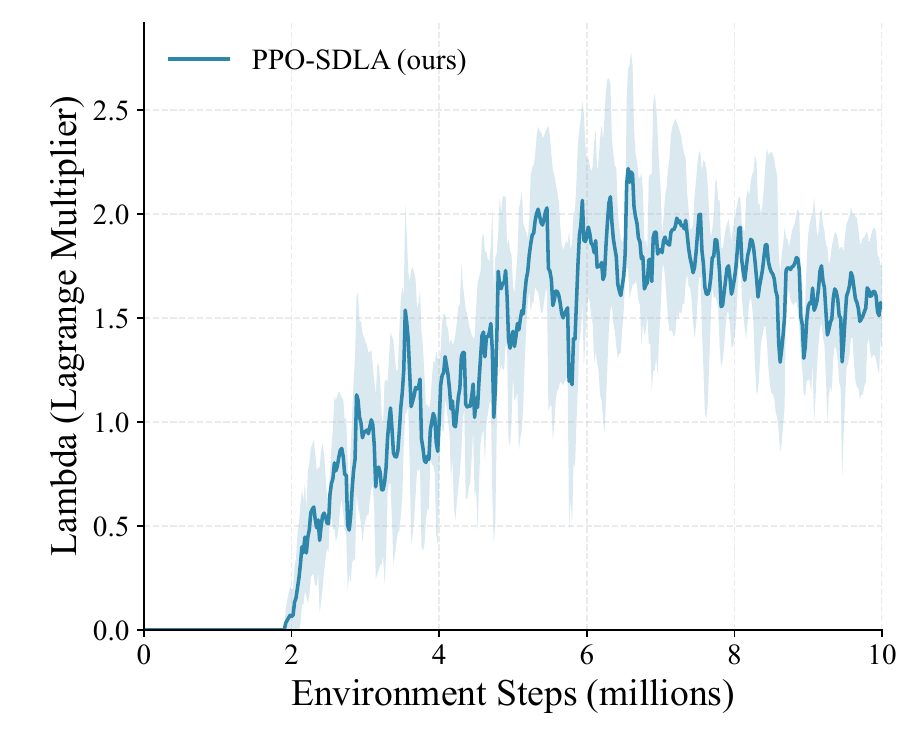}}
\hfill
\subfloat[HalfCheetah]{\includegraphics[width=0.19\textwidth]{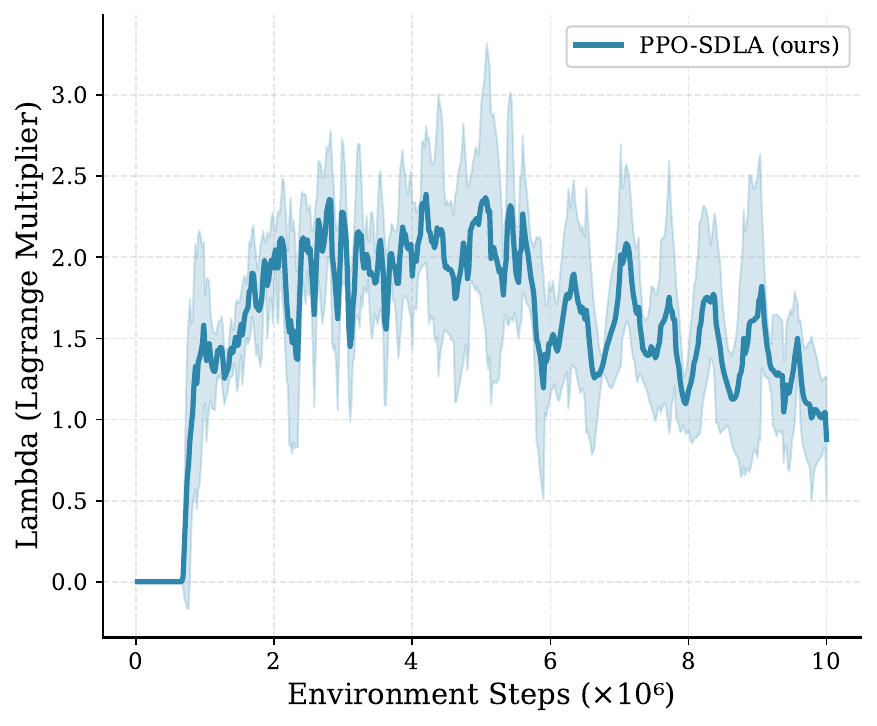}}
\hfill
\subfloat[Hopper]{\includegraphics[width=0.19\textwidth]{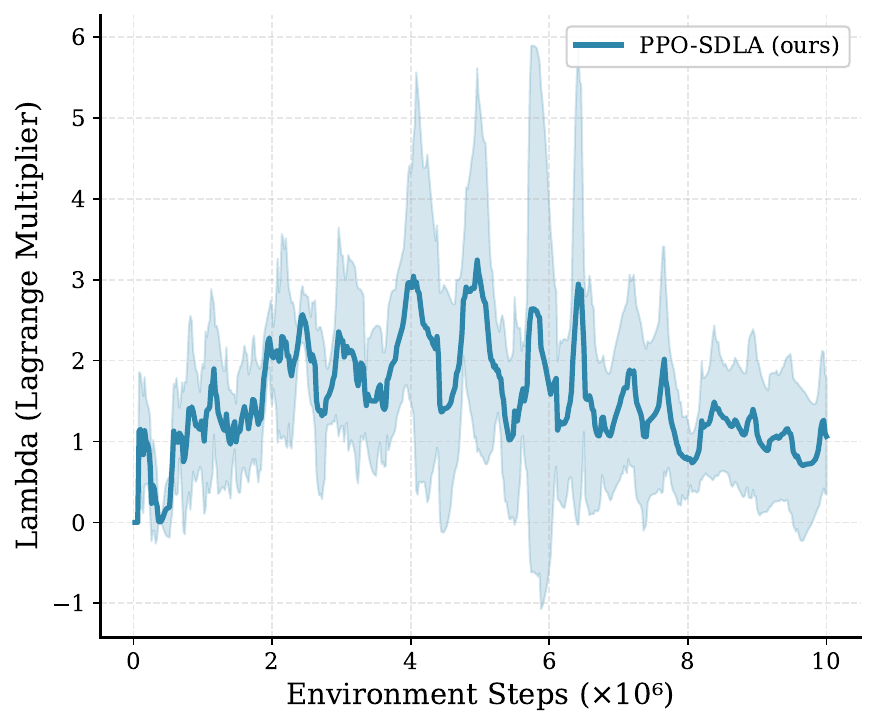}}
\hfill
\subfloat[Swimmer]{\includegraphics[width=0.19\textwidth]{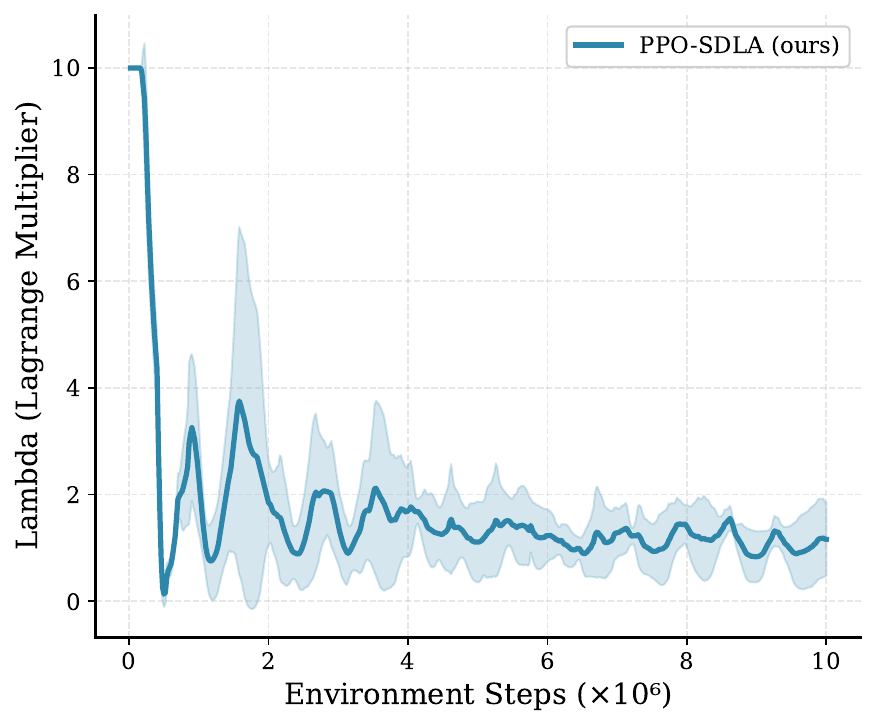}}
\hfill
\subfloat[Walker2d]{\includegraphics[width=0.19\textwidth]{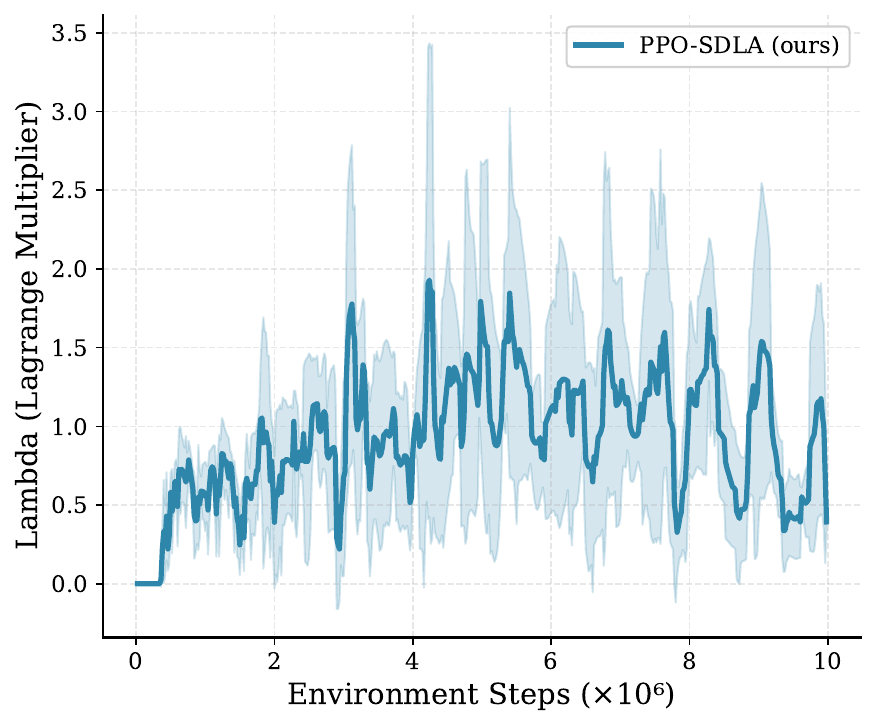}}

\caption{PPO-SDLA on Safety-Gym velocity tasks. Rows show episode reward, episode cost, and the dual variable \(\lambda\), respectively.}
\label{fig:ppo_all}
\end{figure*}

\begin{figure*}[t]
\centering
\captionsetup[subfigure]{justification=centering}

\subfloat[Ant]{\includegraphics[width=0.19\textwidth]{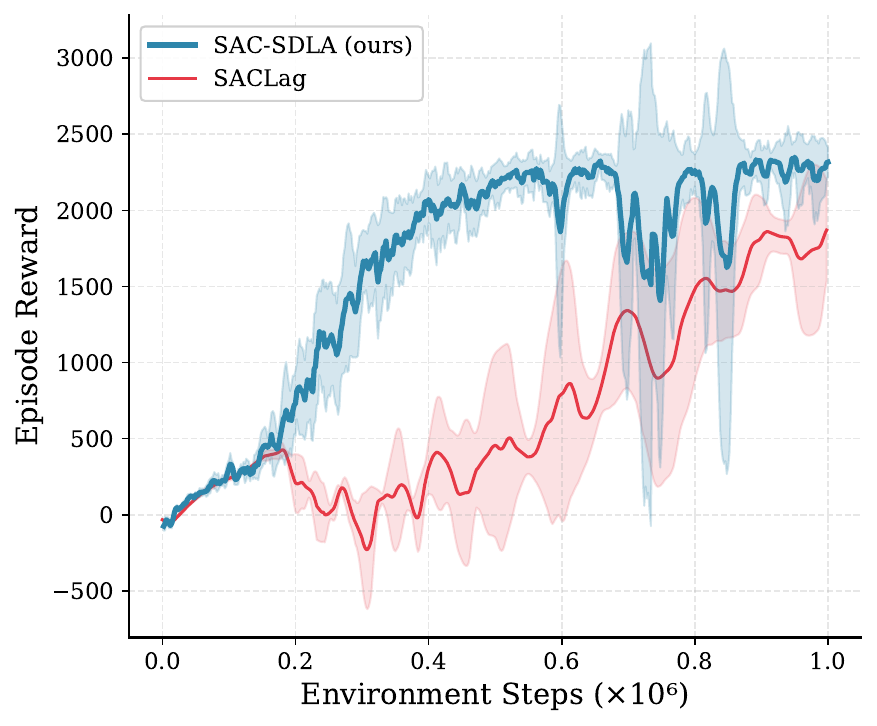}}
\hfill
\subfloat[HalfCheetah]{\includegraphics[width=0.19\textwidth]{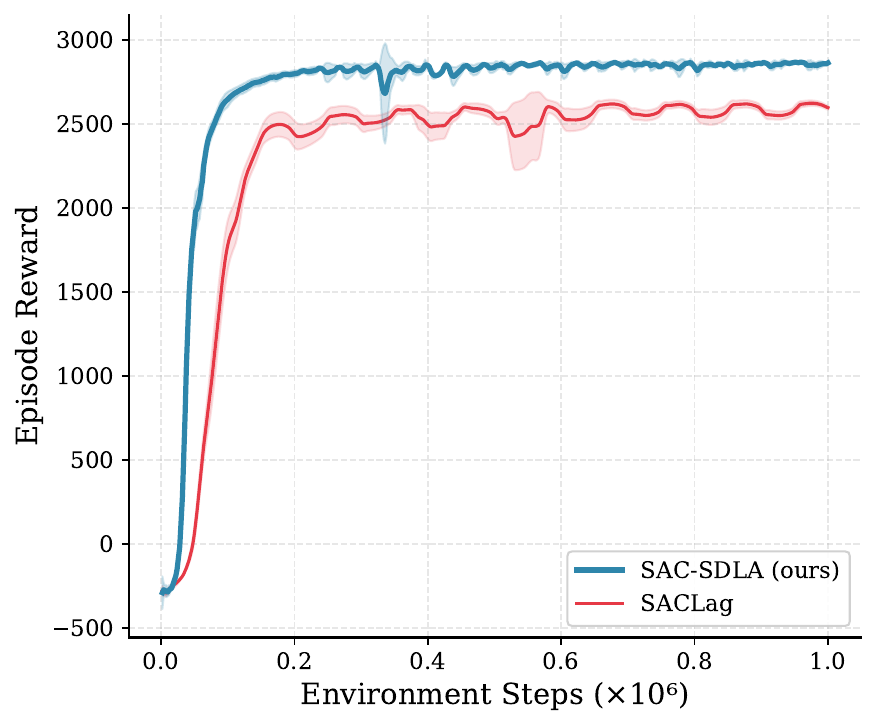}}
\hfill
\subfloat[Hopper]{\includegraphics[width=0.19\textwidth]{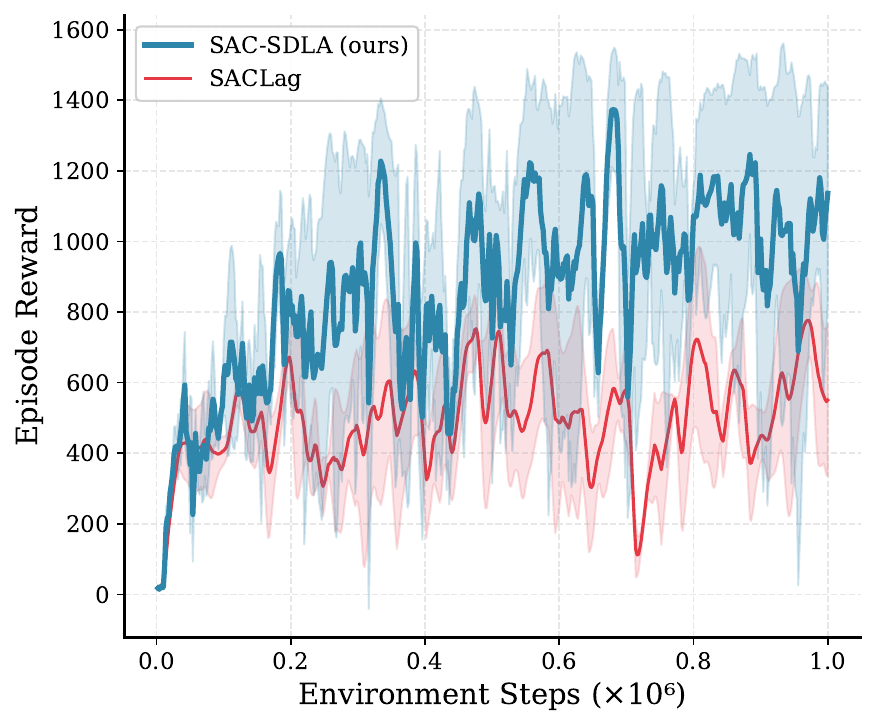}}
\hfill
\subfloat[Swimmer]{\includegraphics[width=0.19\textwidth]{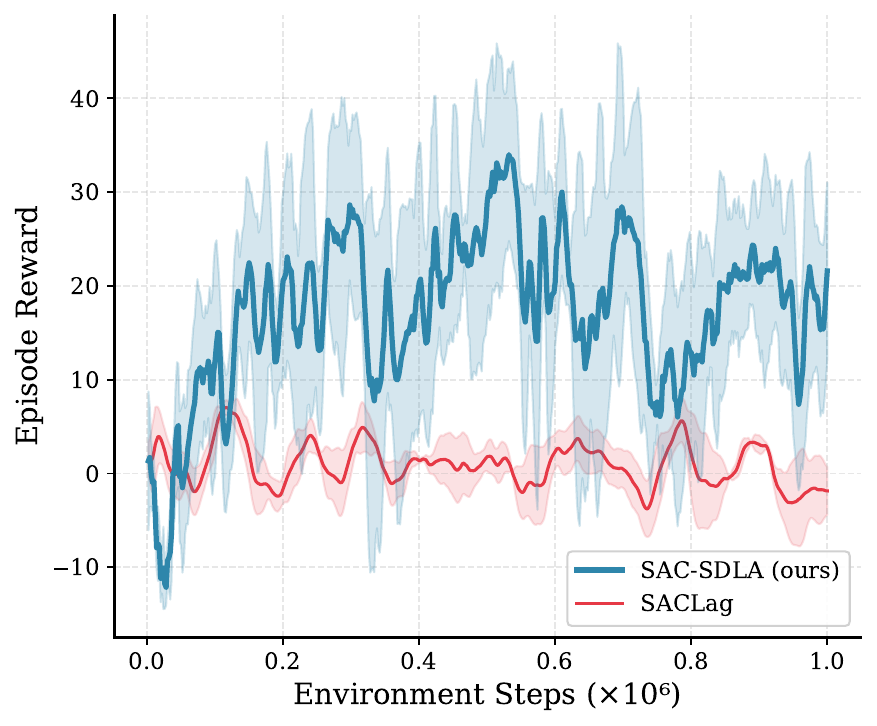}}
\hfill
\subfloat[Walker2d]{\includegraphics[width=0.19\textwidth]{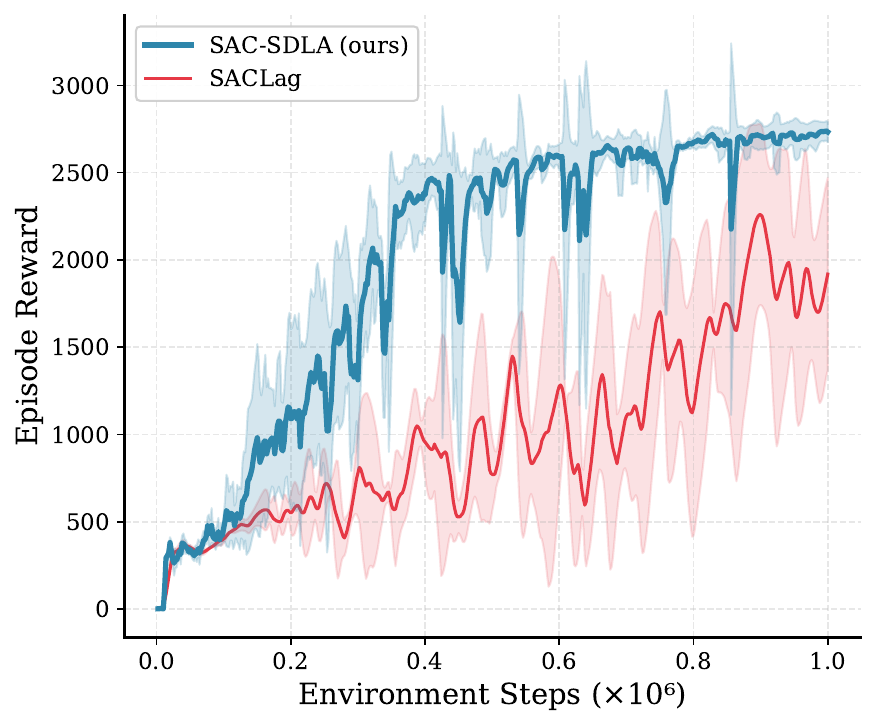}}

\vspace{0.2em}

\subfloat[Ant]{\includegraphics[width=0.19\textwidth]{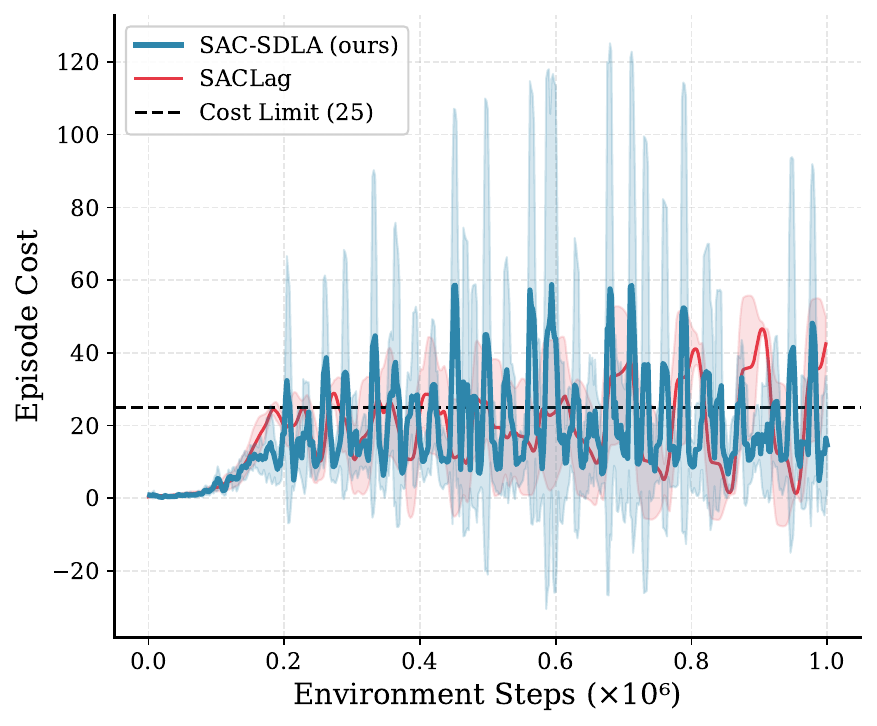}}
\hfill
\subfloat[HalfCheetah]{\includegraphics[width=0.19\textwidth]{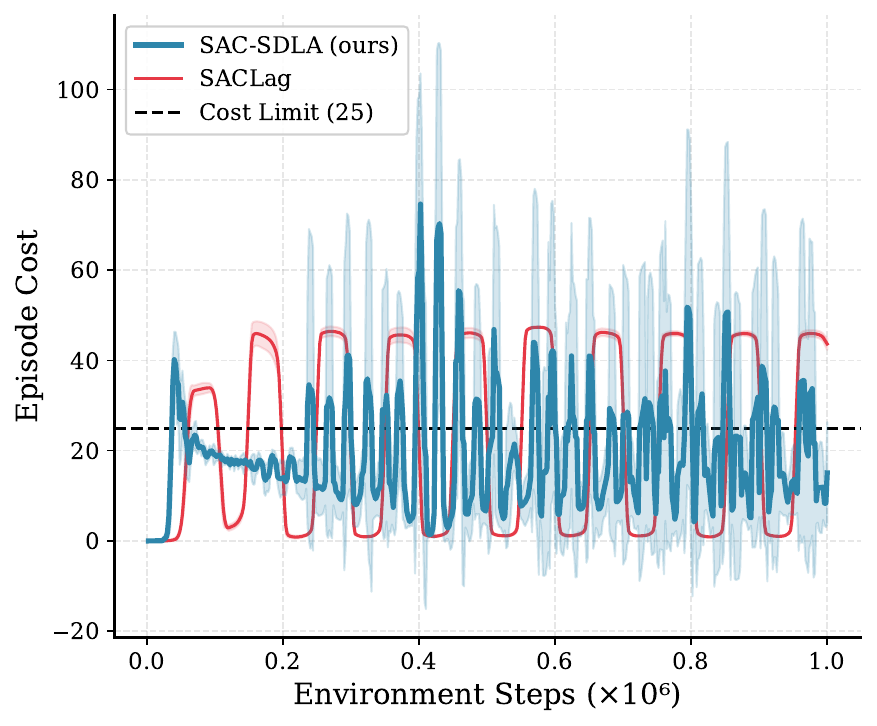}}
\hfill
\subfloat[Hopper]{\includegraphics[width=0.19\textwidth]{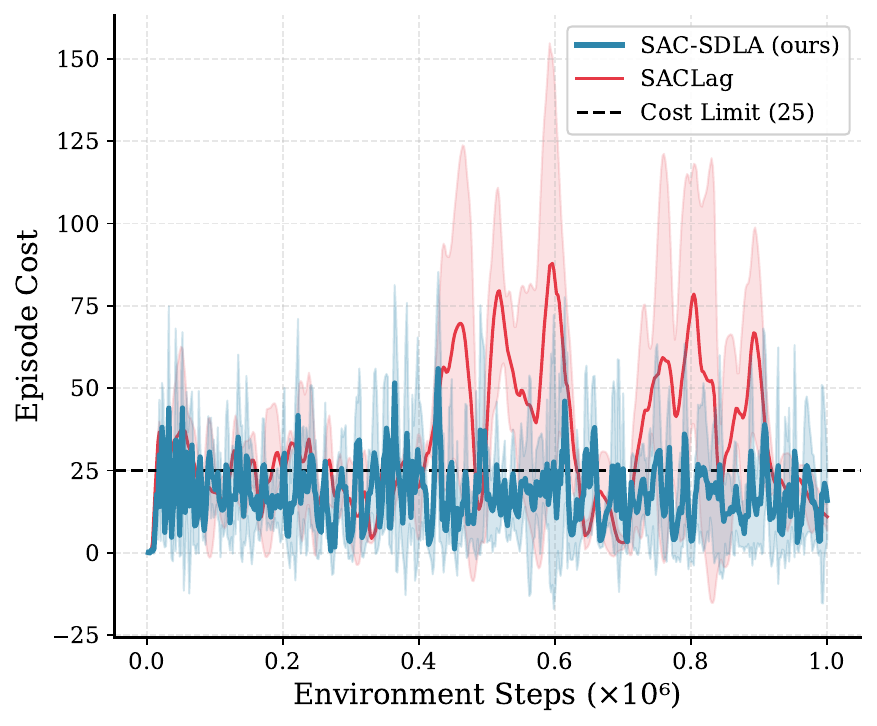}}
\hfill
\subfloat[Swimmer]{\includegraphics[width=0.19\textwidth]{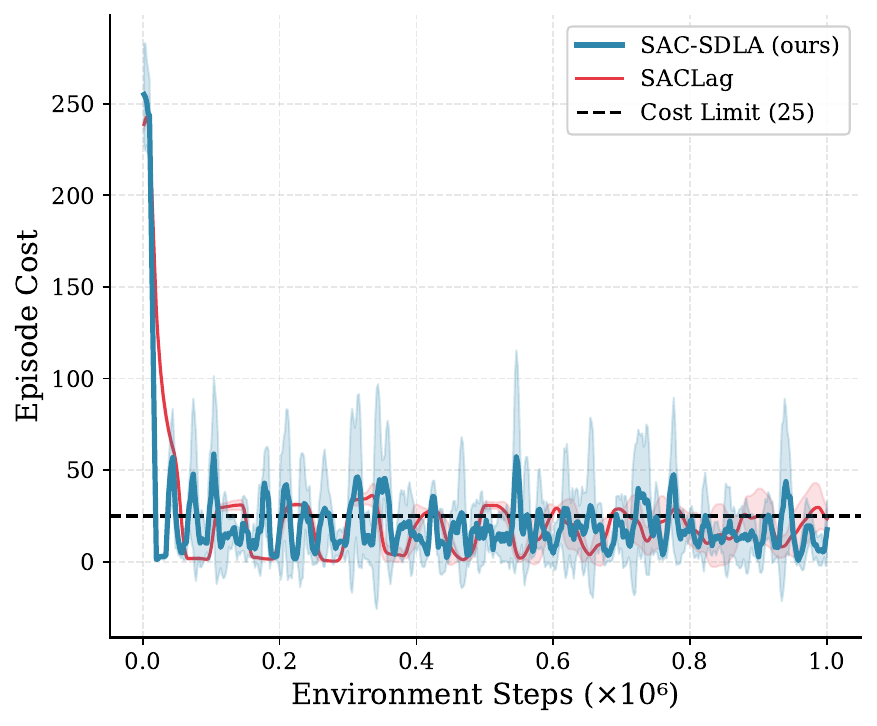}}
\hfill
\subfloat[Walker2d]{\includegraphics[width=0.19\textwidth]{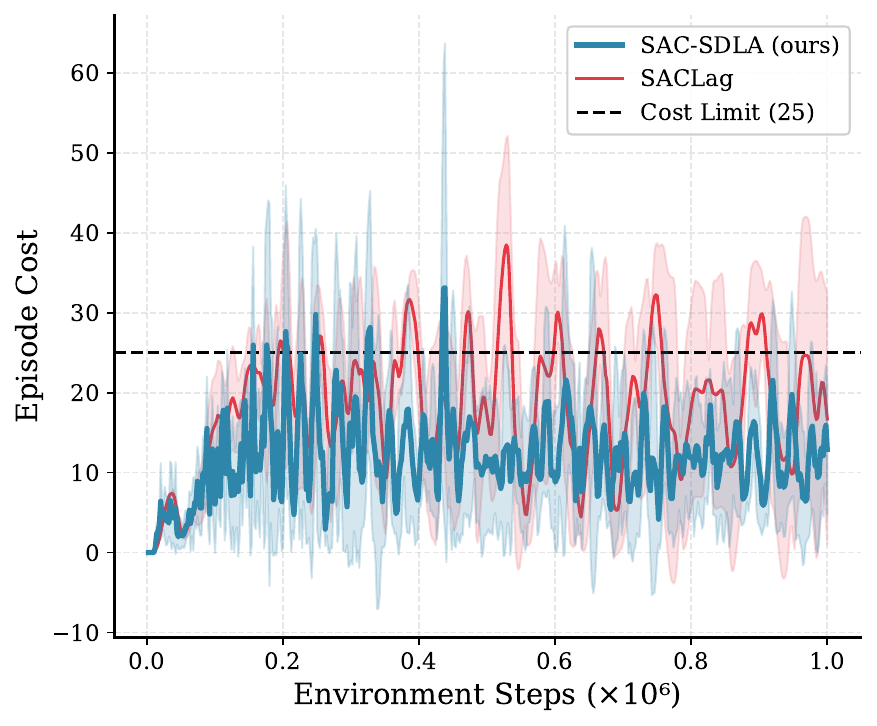}}

\vspace{0.2em}

\subfloat[Ant]{\includegraphics[width=0.19\textwidth]{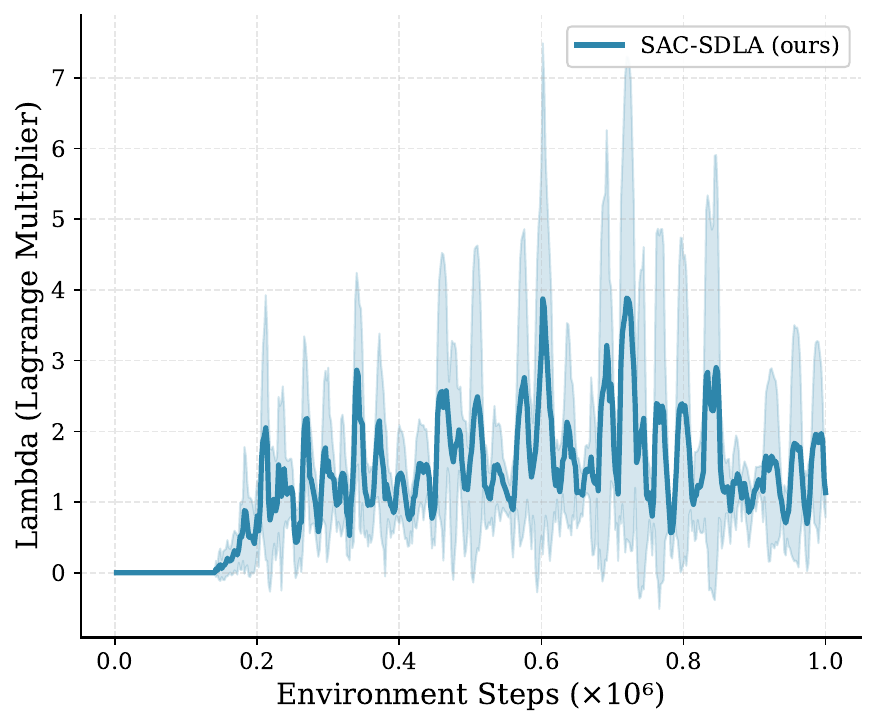}}
\hfill
\subfloat[HalfCheetah]{\includegraphics[width=0.19\textwidth]{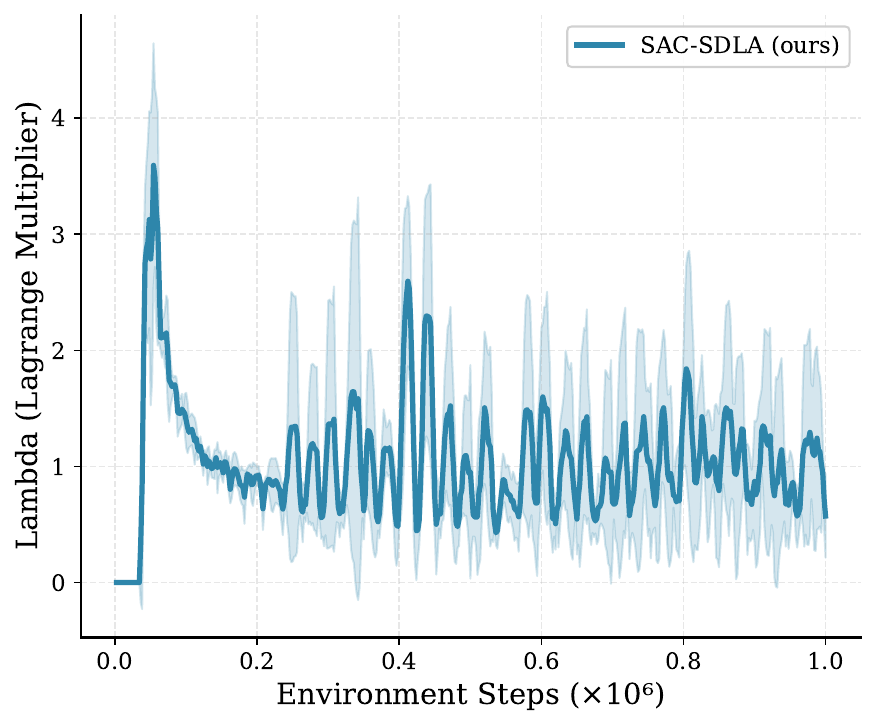}}
\hfill
\subfloat[Hopper]{\includegraphics[width=0.19\textwidth]{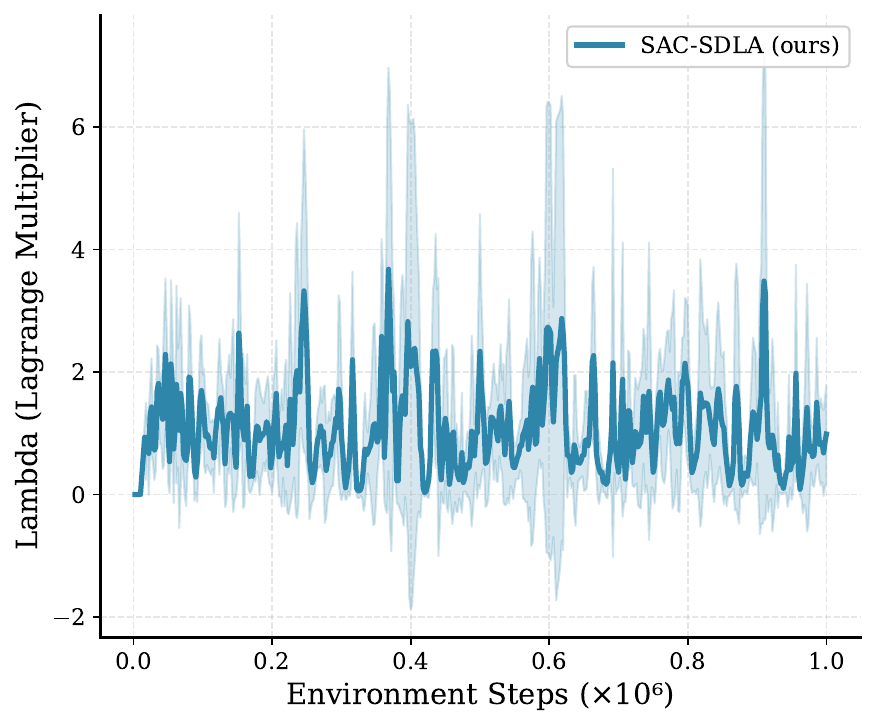}}
\hfill
\subfloat[Swimmer]{\includegraphics[width=0.19\textwidth]{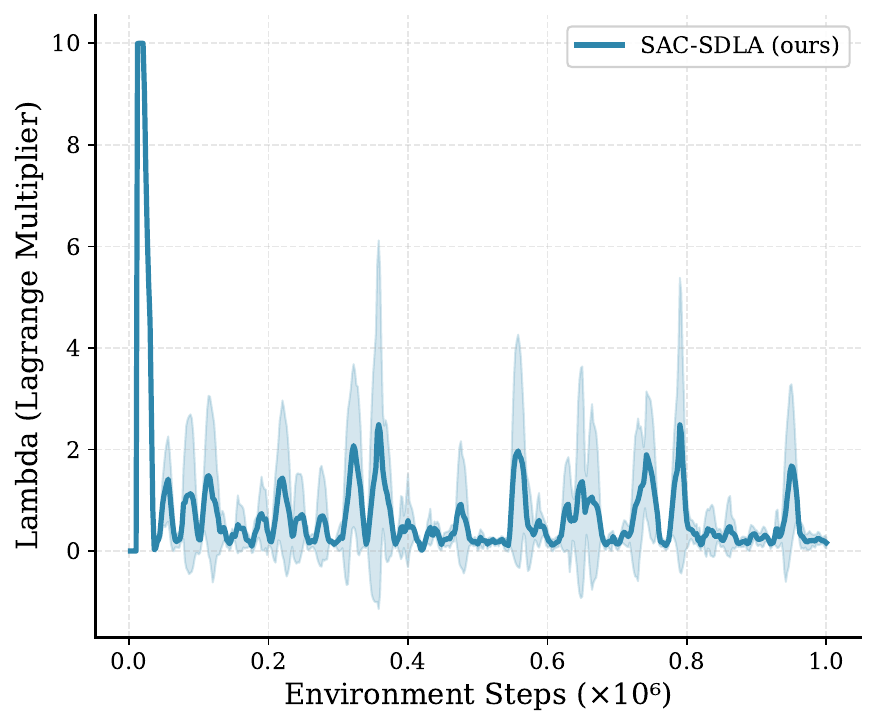}}
\hfill
\subfloat[Walker2d]{\includegraphics[width=0.19\textwidth]{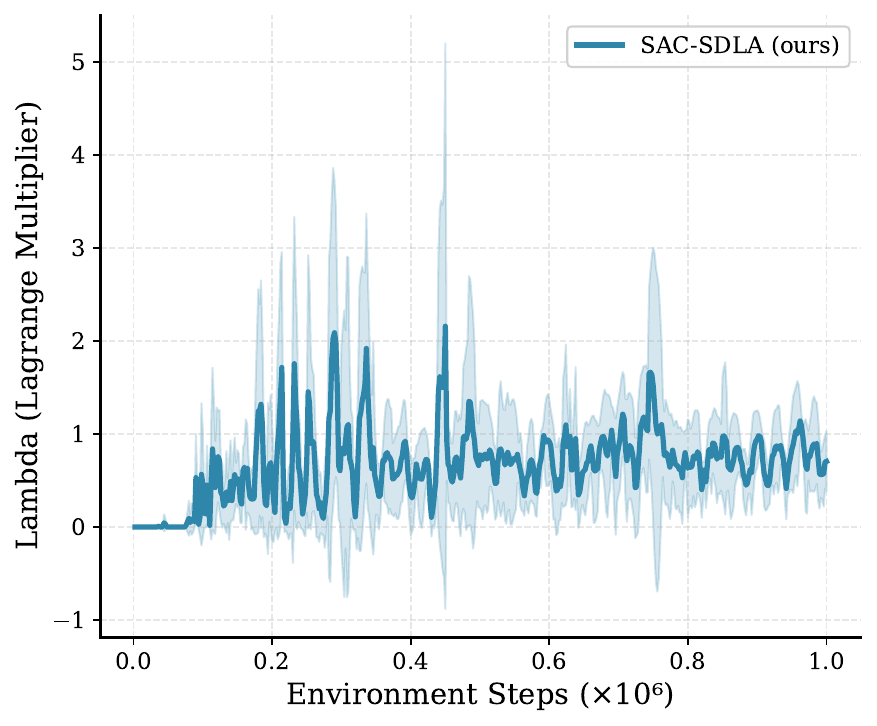}}

\caption{SAC-SDLA on Safety-Gym velocity tasks. Rows show episode reward, episode cost, and the dual variable \(\lambda\), respectively.}
\label{fig:sac_all}
\end{figure*}

Figure~\ref{fig:ppo_all} reports PPO-SDLA learning curves; Figure~\ref{fig:sac_all} reports the SAC-SDLA counterpart.

We integrate SDLA with SAC and evaluate it on Safety-Gym velocity tasks. Figure~\ref{fig:sac_all} presents the reward, cost, and Lagrange multiplier trajectories, respectively. 

\subsection{Additional Comparisons and Training Diagnostics}
\label{sec:additional_diagnostics}
\subsubsection{Constraint Tracking during Training}
\begin{figure}[t]
\centering
\includegraphics[width=\textwidth]{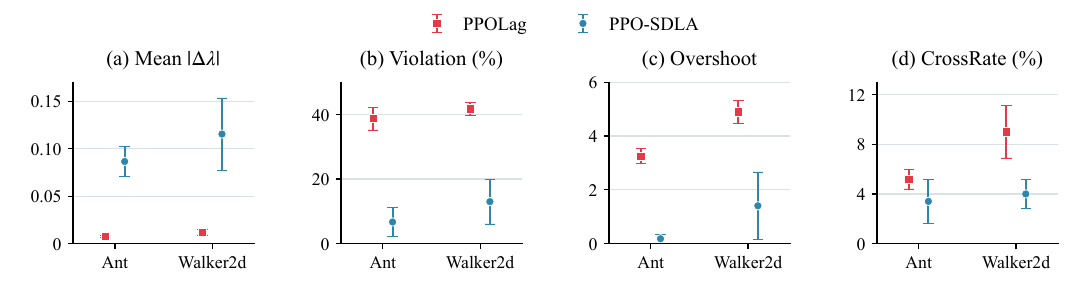}
\caption{Full-training constraint tracking on Ant and Walker2d, using the values in main Table~\ref{tab:training_tracking}. Points show means and error bars show one standard deviation. Panels report mean absolute multiplier change, violation rate, average overshoot, and constraint-boundary crossing rate. The cost budget is $d=25$; overshoot includes all logged updates.}
\label{fig:constraint_tracking_summary}
\end{figure}

We compare PPO-SDLA with PPOLag on the Ant and Walker2d velocity tasks using complete training logs. PPOLag was rerun with multiplier logging for this comparison. Let $c_t^{\mathrm{log}}$ denote the logged episode cost and let $T$ be the number of logged updates. We report the violation rate, average overshoot, and constraint-boundary crossing rate:
\begin{align*}
\mathrm{Violation}(\%)&=\frac{100}{T}\sum_{t=1}^T\mathbf 1\{c_t^{\mathrm{log}}>d\},\qquad
\mathrm{Overshoot}=\frac1T\sum_{t=1}^T[c_t^{\mathrm{log}}-d]_+,\\
\mathrm{CrossRate}(\%)&=\frac{100}{T-1}\sum_{t=1}^{T-1}
\mathbf 1\{(c_t^{\mathrm{log}}-d)(c_{t+1}^{\mathrm{log}}-d)<0\}.
\end{align*}
Table~\ref{tab:training_tracking} in the main paper reports these summaries over the full logged training horizon; Figure~\ref{fig:constraint_tracking_summary} presents the same values graphically.

\subsubsection{Wall-Clock Time}
\label{app:runtime}
On SafetyAntVelocity with a single NVIDIA RTX 4090 (24 GB), stable-training epochs take $23.58$ seconds for PPO-SDLA versus $23.22$ for PPOLag. Runs take approximately $3.28$ versus $3.23$ GPU-hours, a $1.5\%$ increase.

\subsubsection{Multiplier Cap}
The same $\lambda_{\max}=10$ is used across tasks. We measure near-cap frequency by
\[
\mathrm{NearCapRate}(\%)=\frac{100}{T}\sum_{t=1}^T
\mathbf 1\{\lambda_t\ge0.95\lambda_{\max}\}.
\]
Table~\ref{tab:dual_cap_usage} shows that the logged multiplier stays below this region on four tasks and enters it for $2.4\%$ of updates on Swimmer.

\begin{table}[t]
\centering
\small
\caption{Fraction of logged PPO-SDLA updates with $\lambda_t\ge0.95\lambda_{\max}$, using $\lambda_{\max}=10$.}
\label{tab:dual_cap_usage}
\begin{tabular}{lc}
\toprule
Environment & Near-cap frequency (\%) \\
\midrule
SafetyAntVelocity-v1 & 0.000 \\
SafetyHalfCheetahVelocity-v1 & 0.000 \\
SafetyHopperVelocity-v1 & 0.000 \\
SafetySwimmerVelocity-v1 & 2.400 \\
SafetyWalker2dVelocity-v1 & 0.000 \\
\bottomrule
\end{tabular}
\end{table}

\subsection{Evaluation on HASARD Benchmark}

\begin{figure*}[htbp]
\centering
\subfloat[Armament R]{\includegraphics[width=0.16\textwidth]{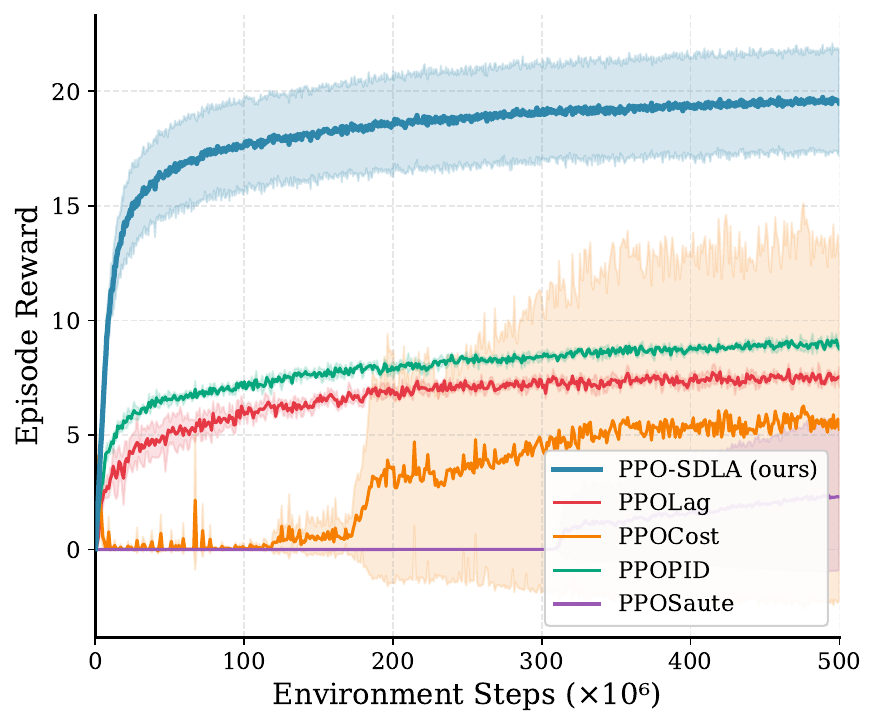}}
\subfloat[Armament C]{\includegraphics[width=0.16\textwidth]{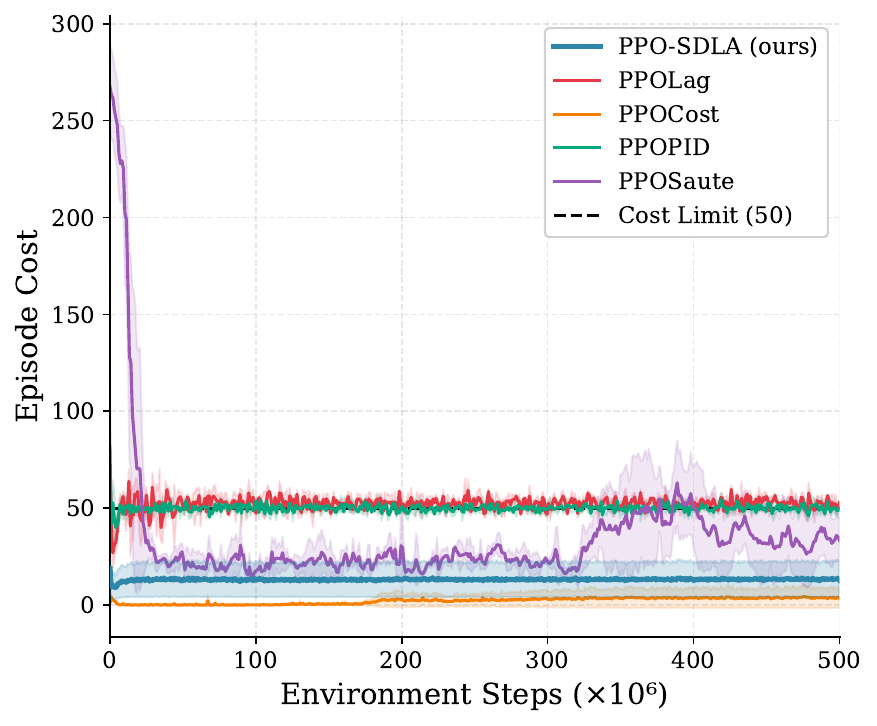}}
\subfloat[Volcanic R]{\includegraphics[width=0.16\textwidth]{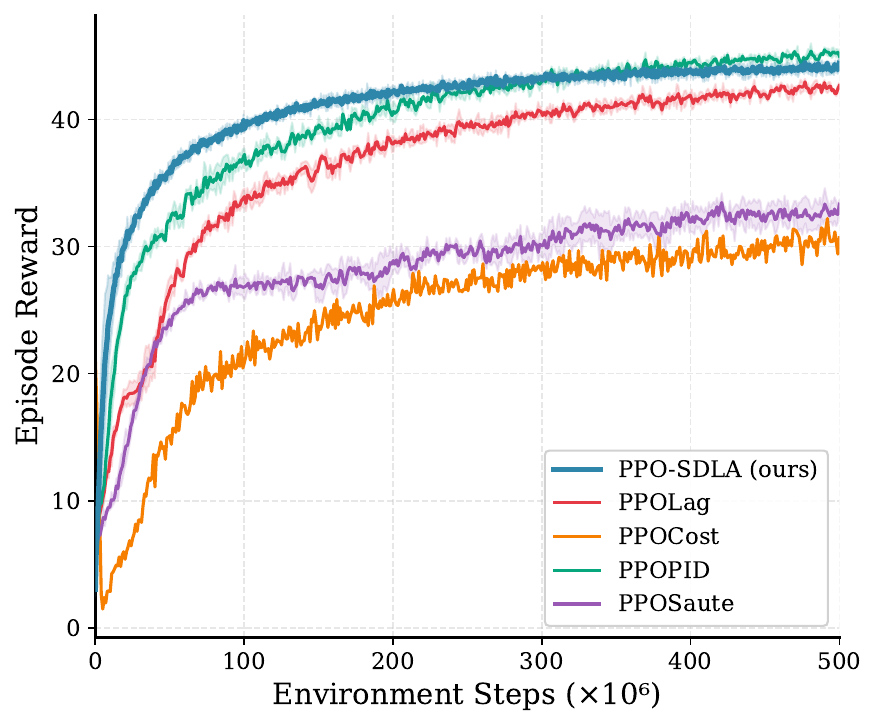}}
\subfloat[Volcanic C]{\includegraphics[width=0.16\textwidth]{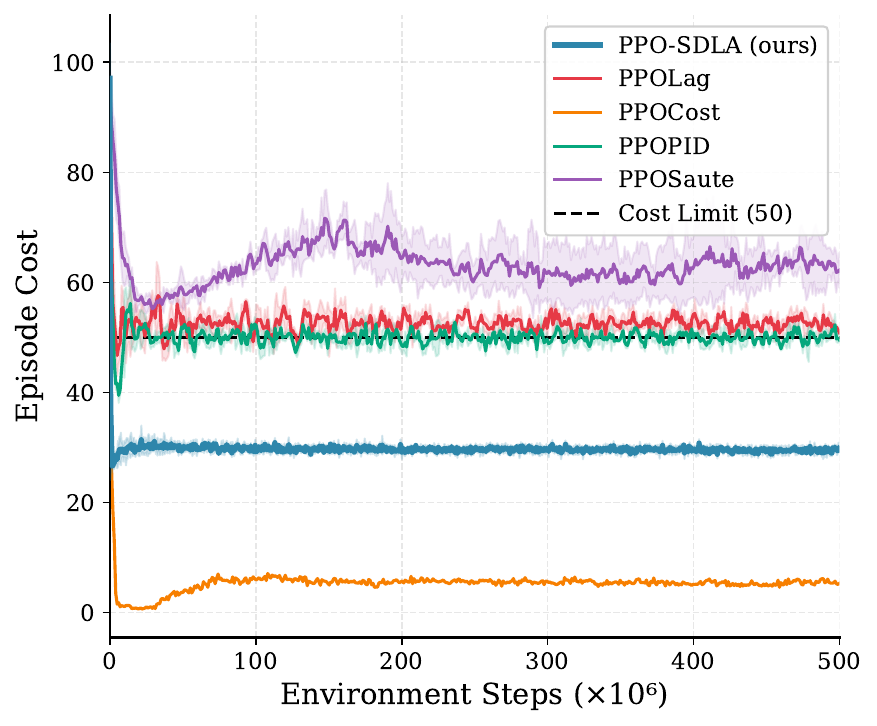}}
\subfloat[Remedy R]{\includegraphics[width=0.16\textwidth]{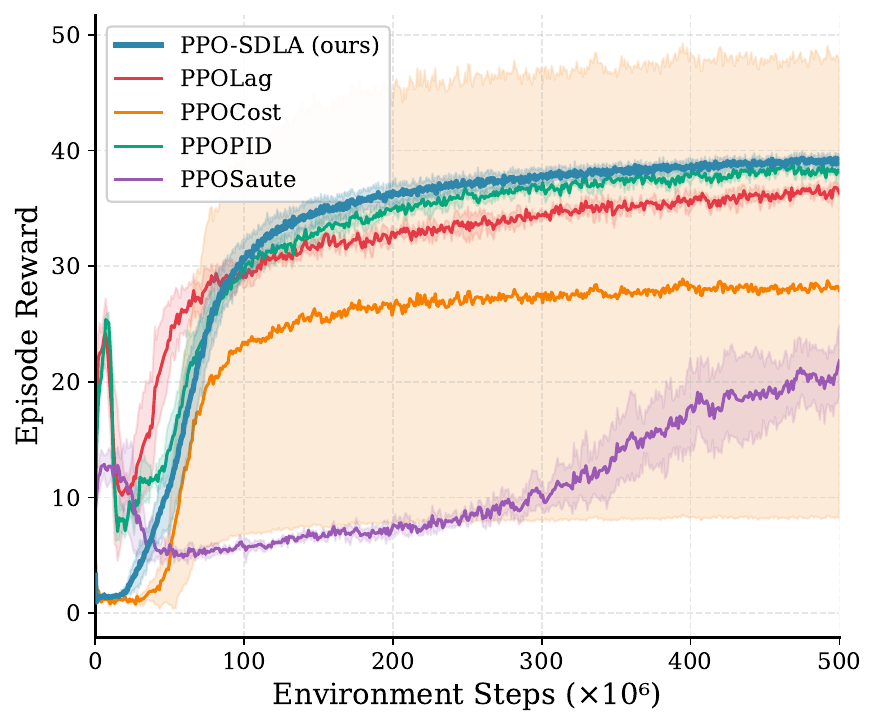}}
\subfloat[Remedy C]{\includegraphics[width=0.16\textwidth]{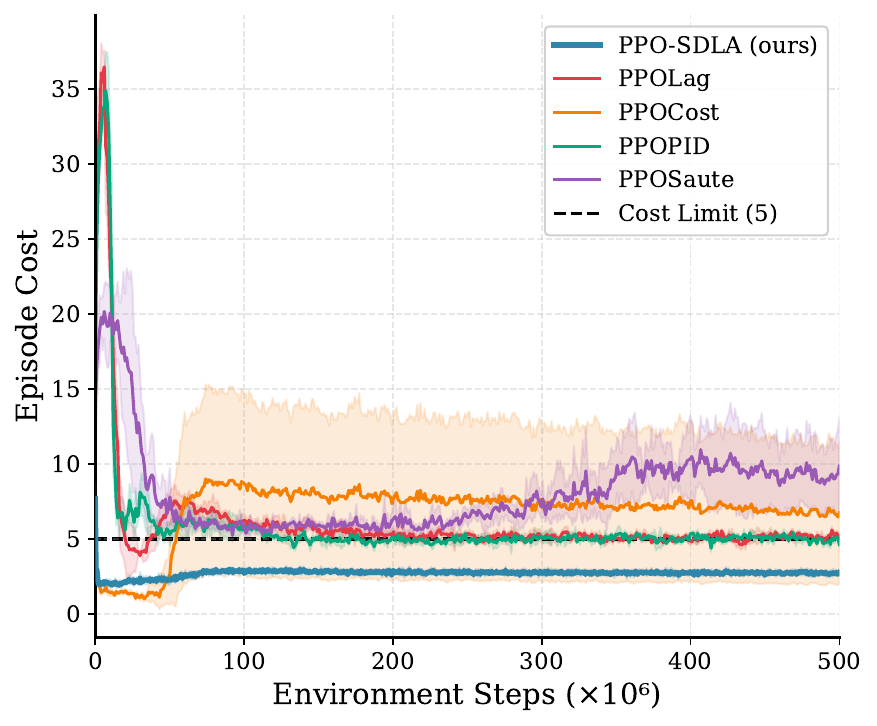}}
\caption{Learning curves on HASARD Level 1 (Part 1): Armament Burden, Volcanic Venture, Remedy Rush.}
\label{fig:hasard_l1_p1}
\end{figure*}

\begin{figure*}[htbp]
\centering
\subfloat[Collateral R]{\includegraphics[width=0.16\textwidth]{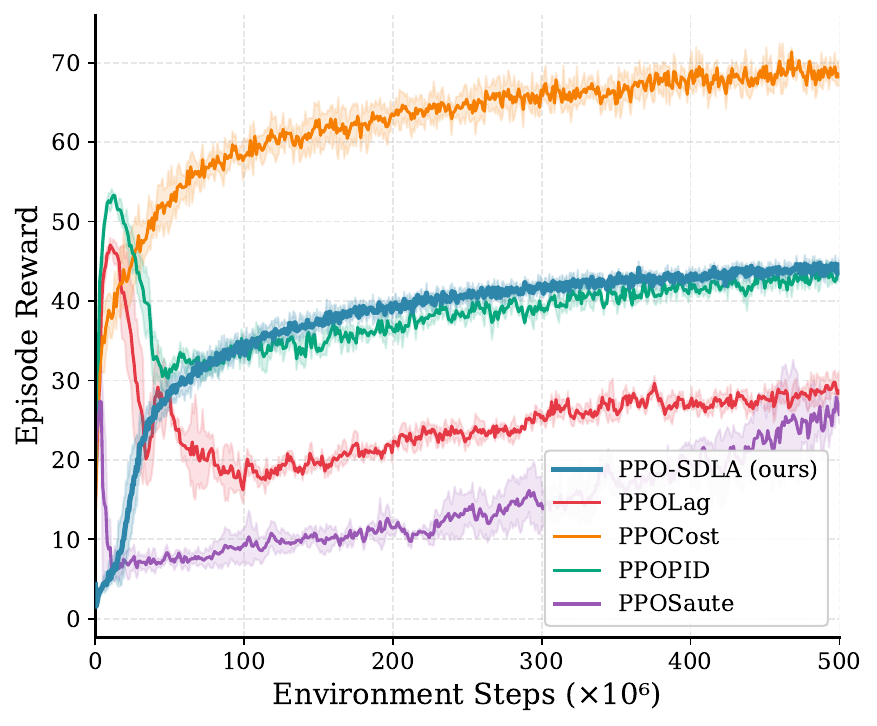}}
\subfloat[Collateral C]{\includegraphics[width=0.16\textwidth]{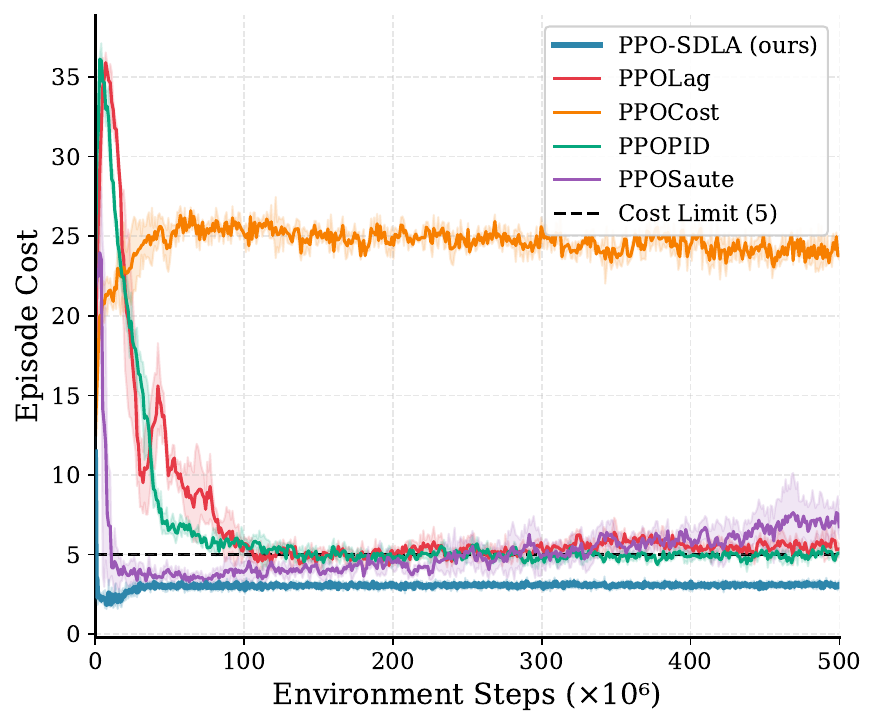}}
\subfloat[Precipice R]{\includegraphics[width=0.16\textwidth]{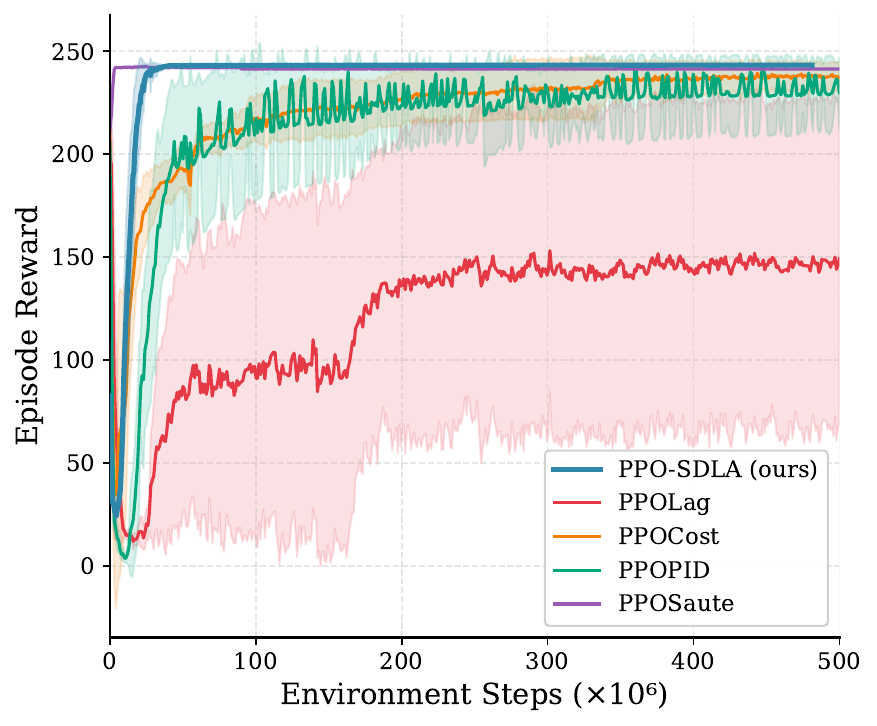}}
\subfloat[Precipice C]{\includegraphics[width=0.16\textwidth]{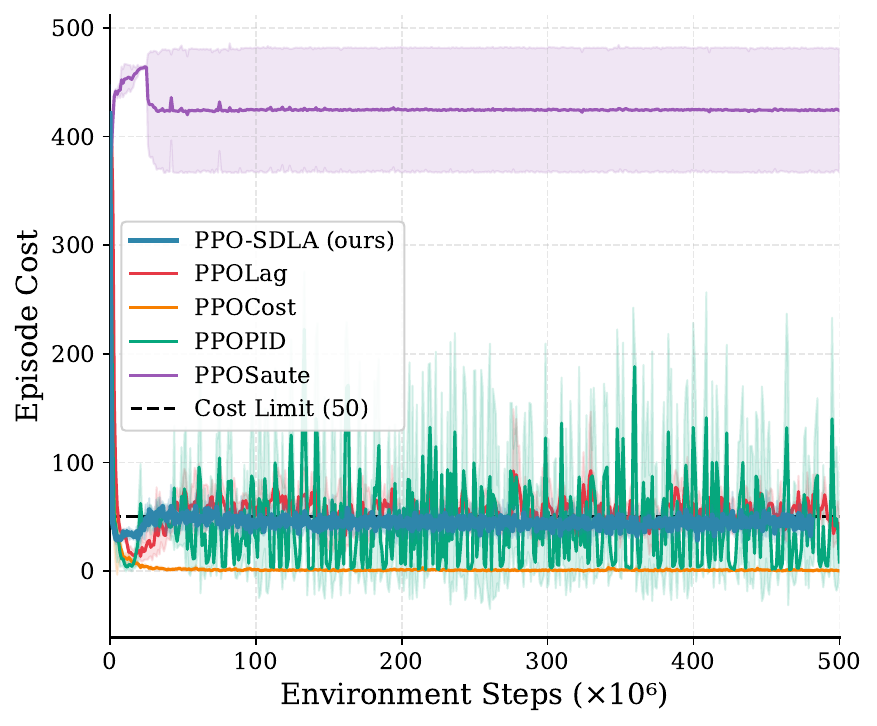}}
\subfloat[Detonator R]{\includegraphics[width=0.16\textwidth]{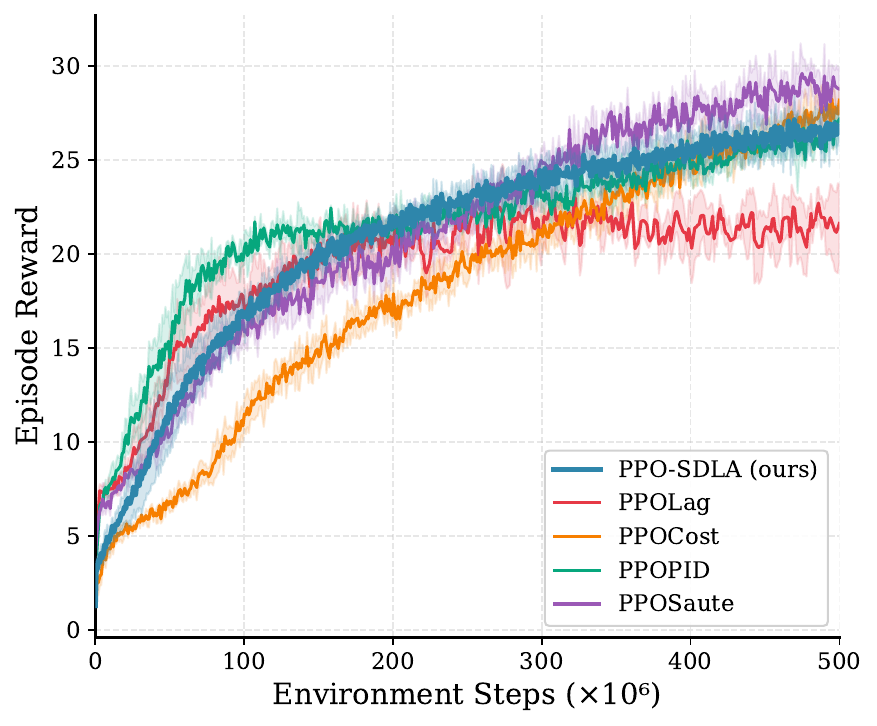}}
\subfloat[Detonator C]{\includegraphics[width=0.16\textwidth]{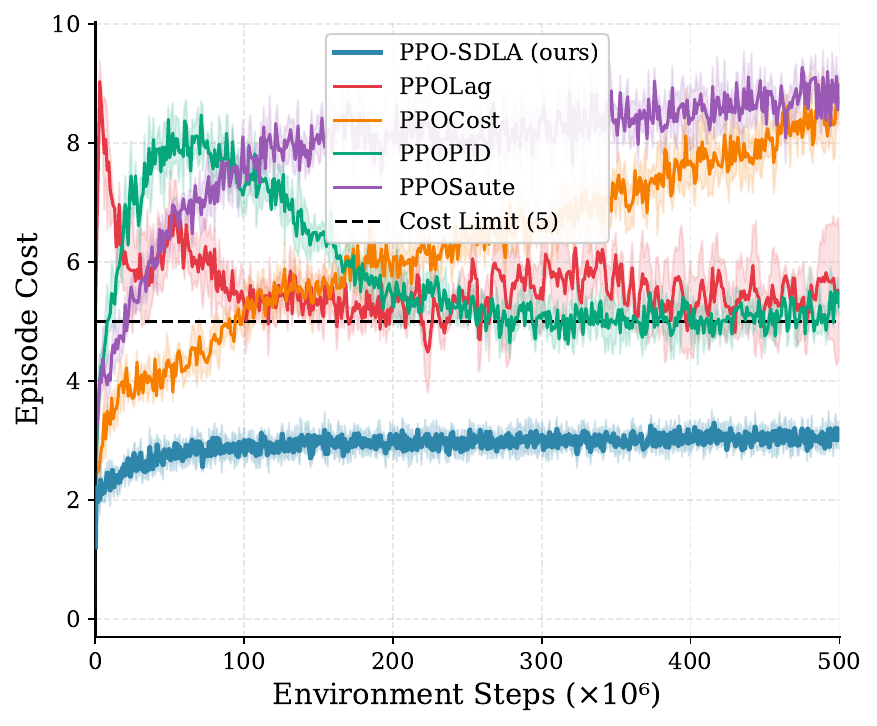}}
\caption{Learning curves on HASARD Level 1 (Part 2): Collateral Damage, Precipice Plunge, Detonator's Dilemma.}
\label{fig:hasard_l1_p2}
\end{figure*}

\begin{figure*}[htbp]
\centering
\subfloat[Armament R]{\includegraphics[width=0.16\textwidth]{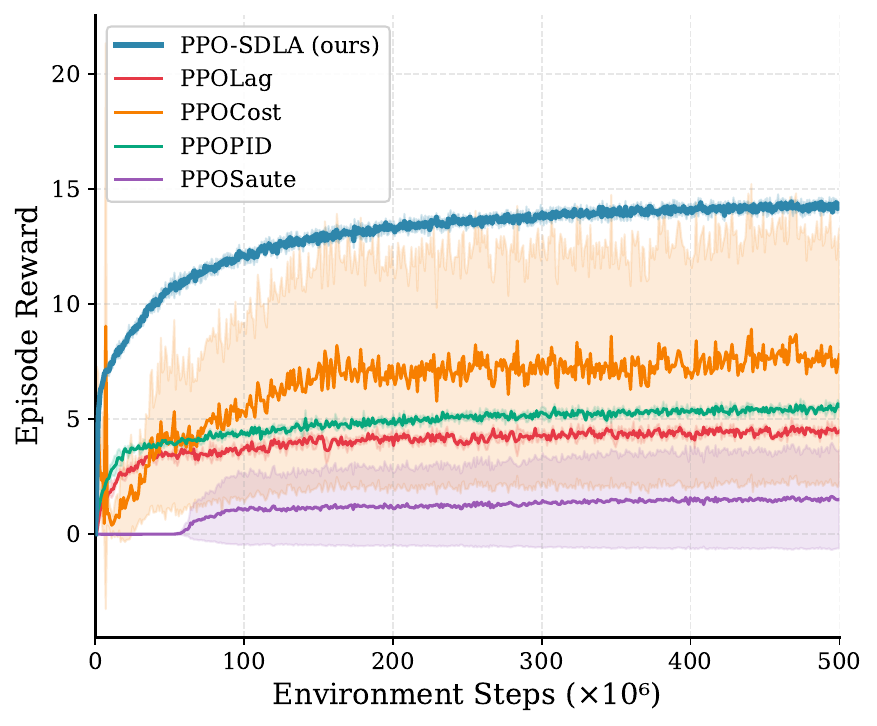}}
\subfloat[Armament C]{\includegraphics[width=0.16\textwidth]{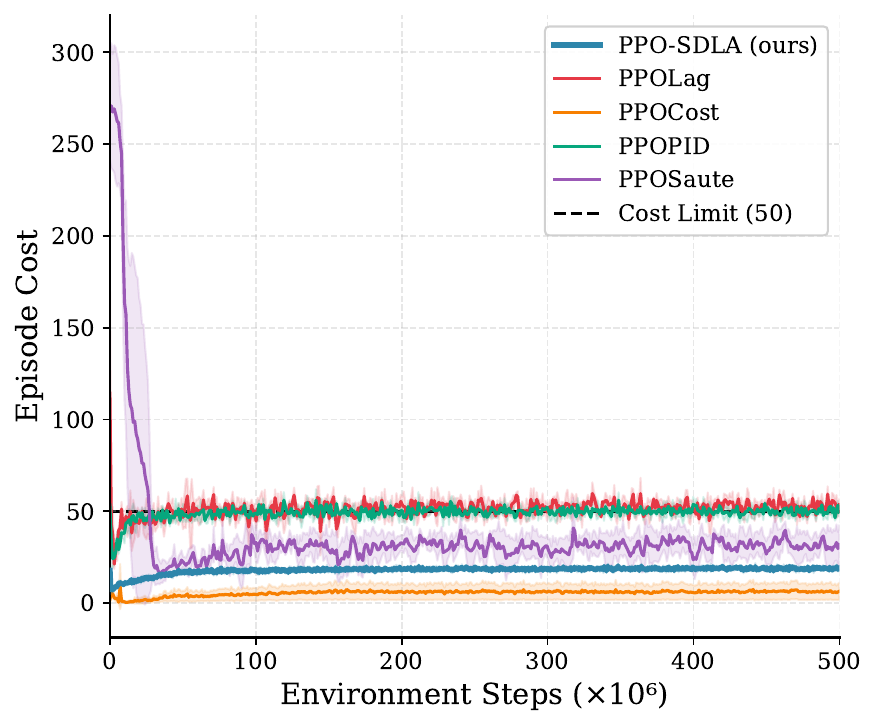}}
\subfloat[Volcanic R]{\includegraphics[width=0.16\textwidth]{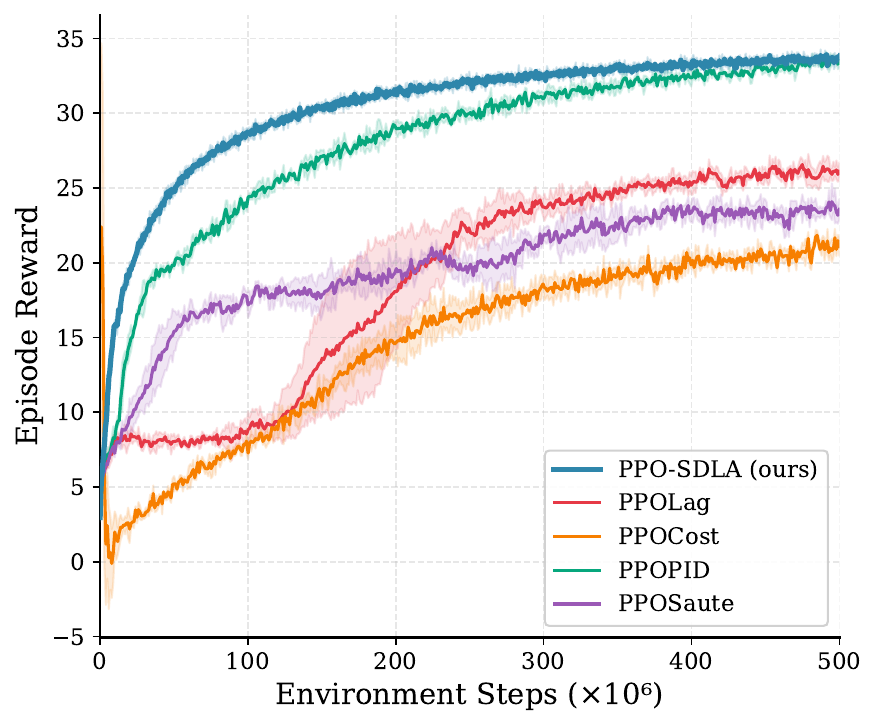}}
\subfloat[Volcanic C]{\includegraphics[width=0.16\textwidth]{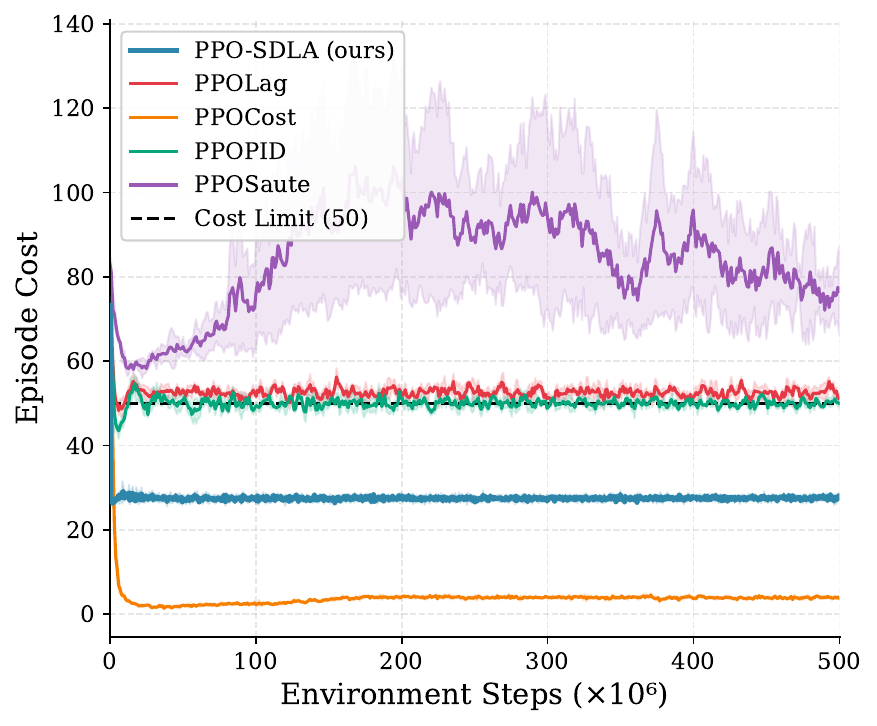}}
\subfloat[Remedy R]{\includegraphics[width=0.16\textwidth]{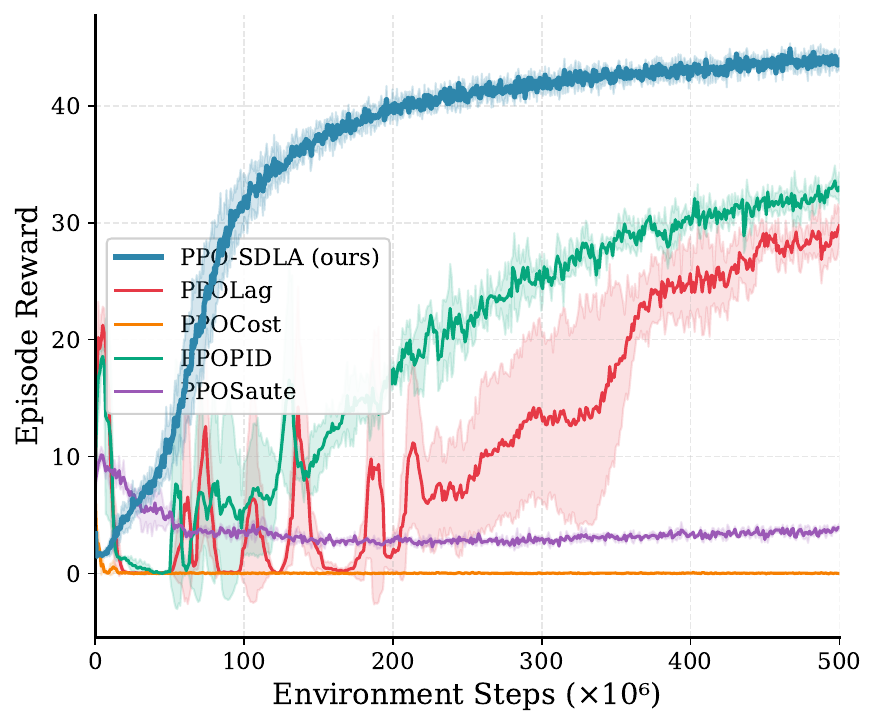}}
\subfloat[Remedy C]{\includegraphics[width=0.16\textwidth]{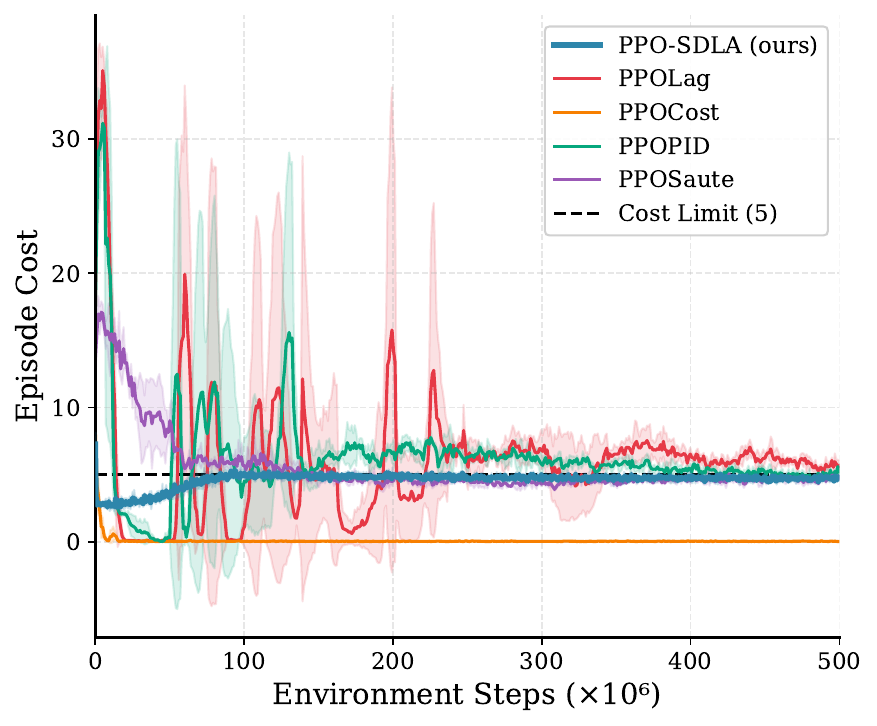}}
\caption{Learning curves on HASARD Level 2 (Part 1): Armament Burden, Volcanic Venture, Remedy Rush.}
\label{fig:hasard_l2_p1}
\end{figure*}

\begin{figure*}[htbp]
\centering
\subfloat[Collateral R]{\includegraphics[width=0.16\textwidth]{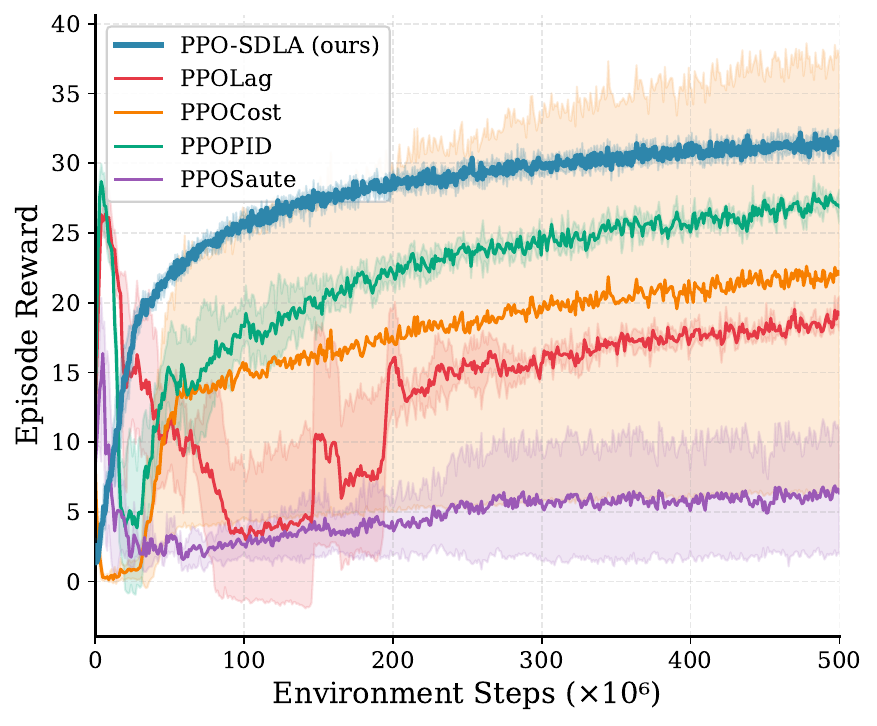}}
\subfloat[Collateral C]{\includegraphics[width=0.16\textwidth]{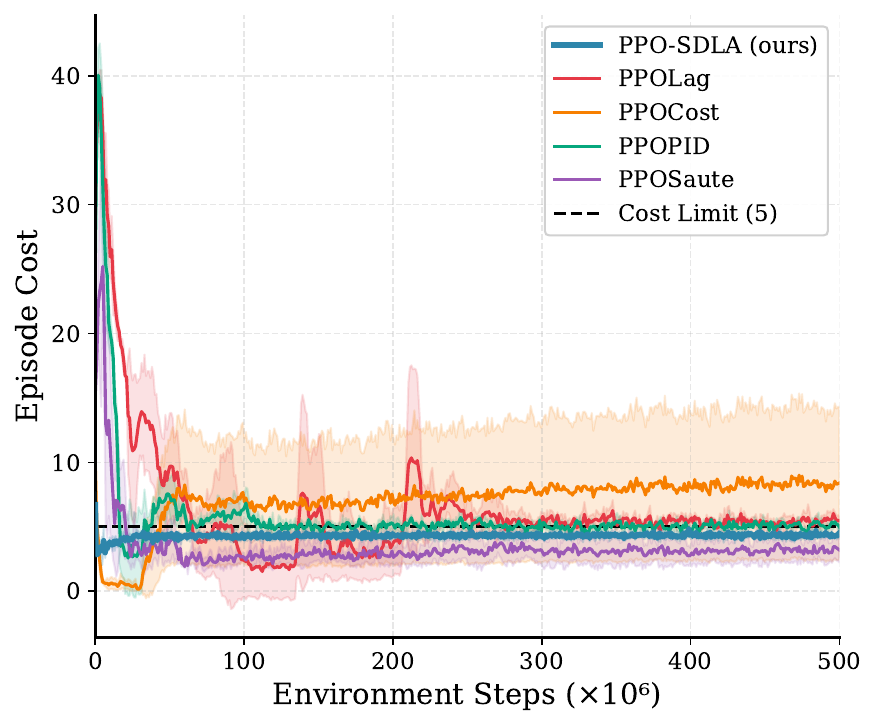}}
\subfloat[Precipice R]{\includegraphics[width=0.16\textwidth]{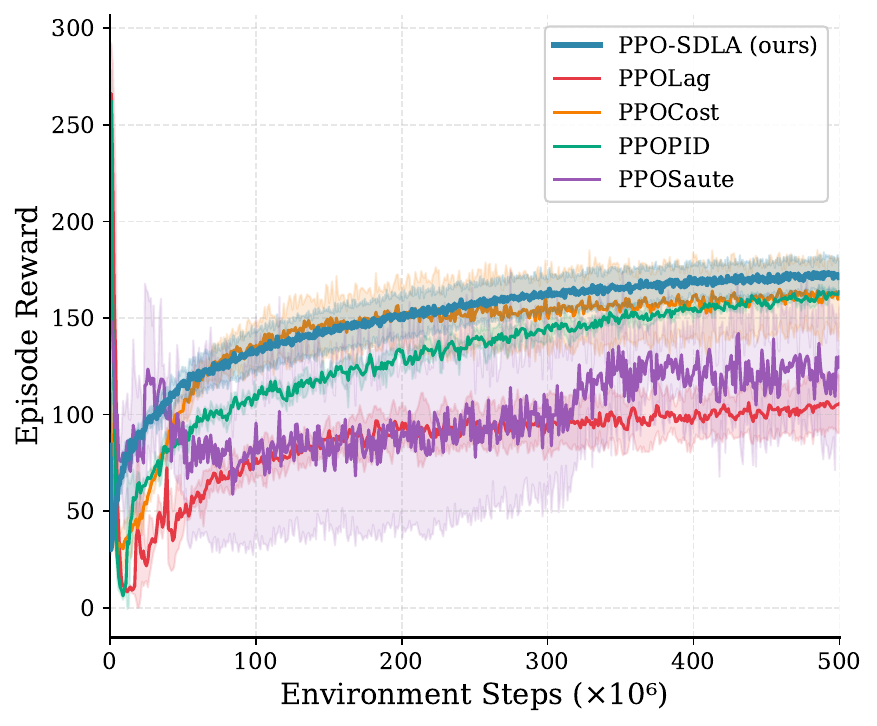}}
\subfloat[Precipice C]{\includegraphics[width=0.16\textwidth]{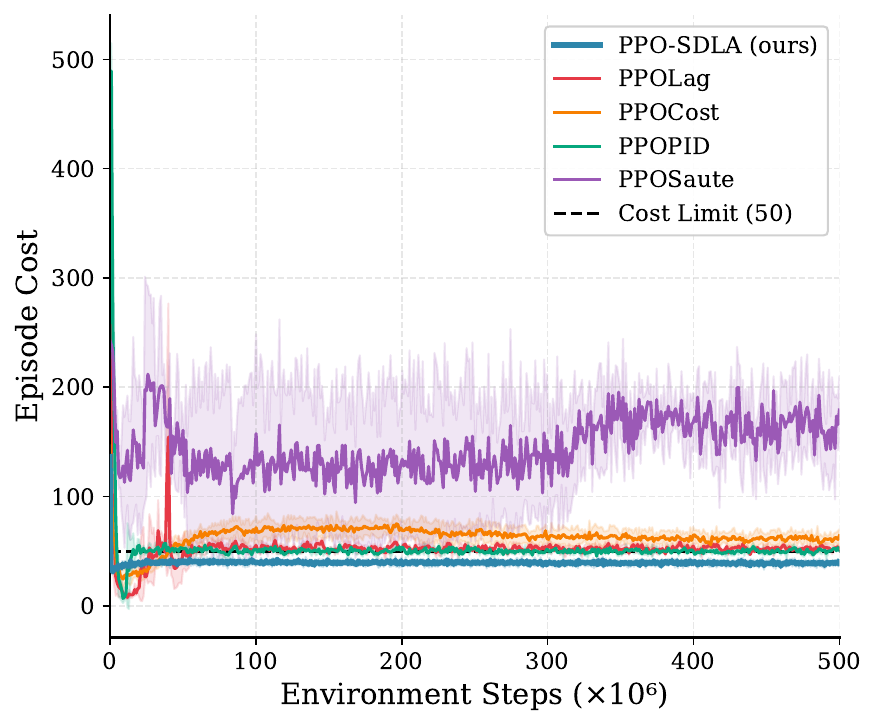}}
\subfloat[Detonator R]{\includegraphics[width=0.16\textwidth]{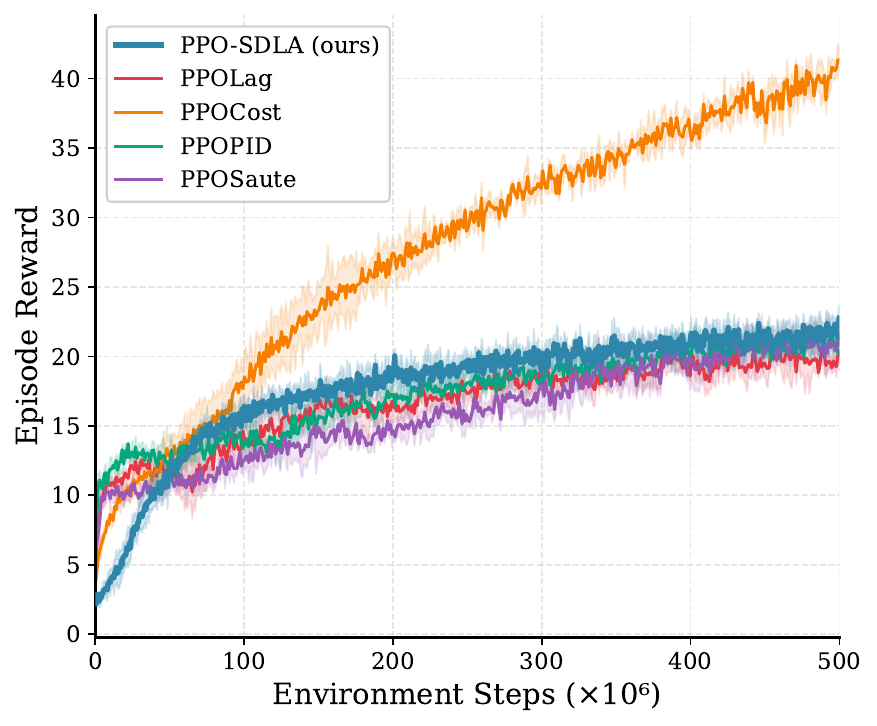}}
\subfloat[Detonator C]{\includegraphics[width=0.16\textwidth]{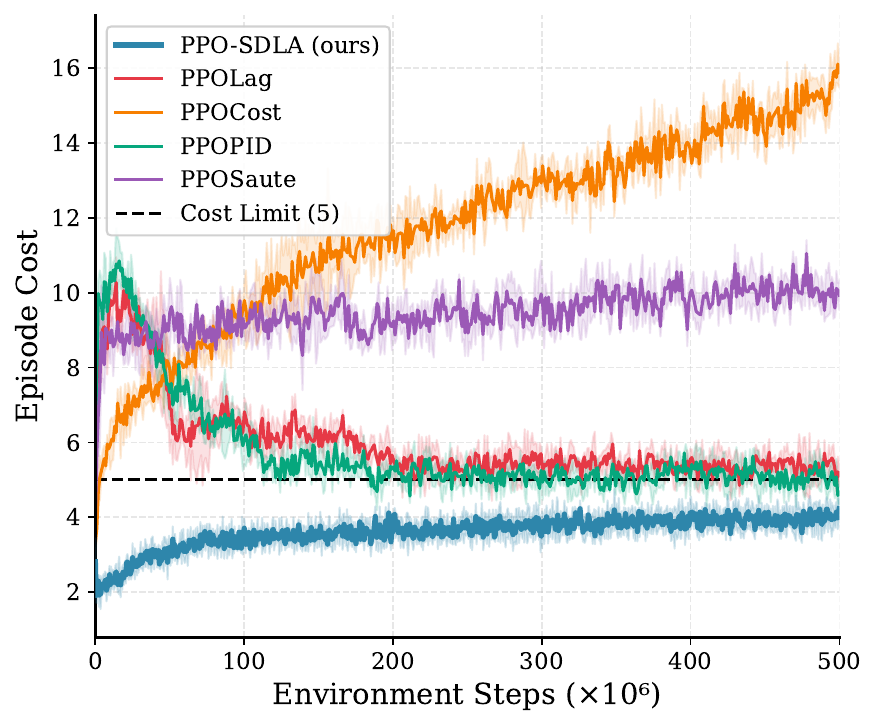}}
\caption{Learning curves on HASARD Level 2 (Part 2): Collateral Damage, Precipice Plunge, Detonator's Dilemma.}
\label{fig:hasard_l2_p2}
\end{figure*}

\begin{figure*}[htbp]
\centering
\subfloat[Armament R]{\includegraphics[width=0.16\textwidth]{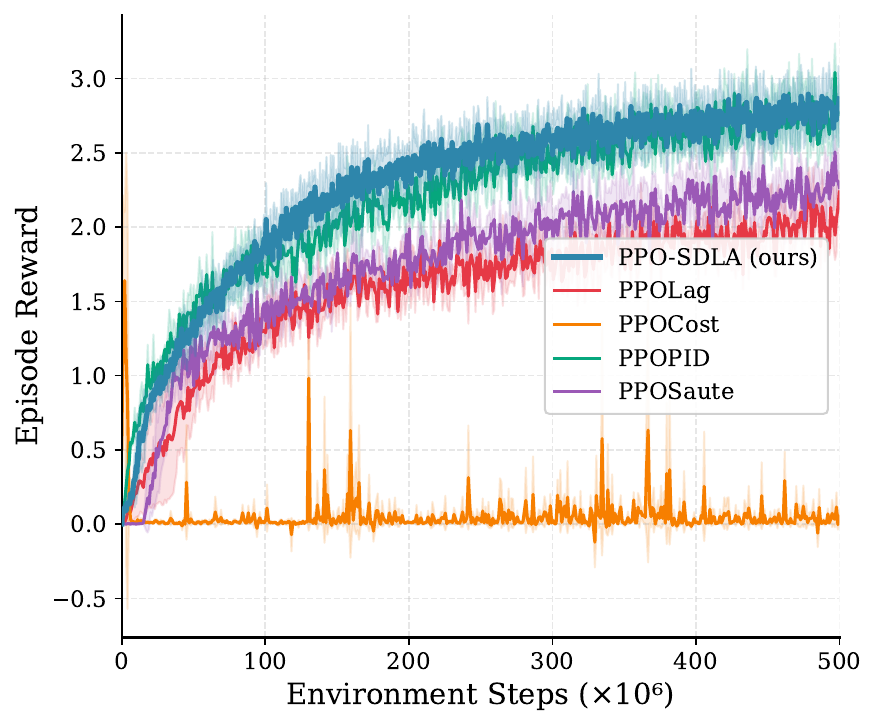}}
\subfloat[Armament C]{\includegraphics[width=0.16\textwidth]{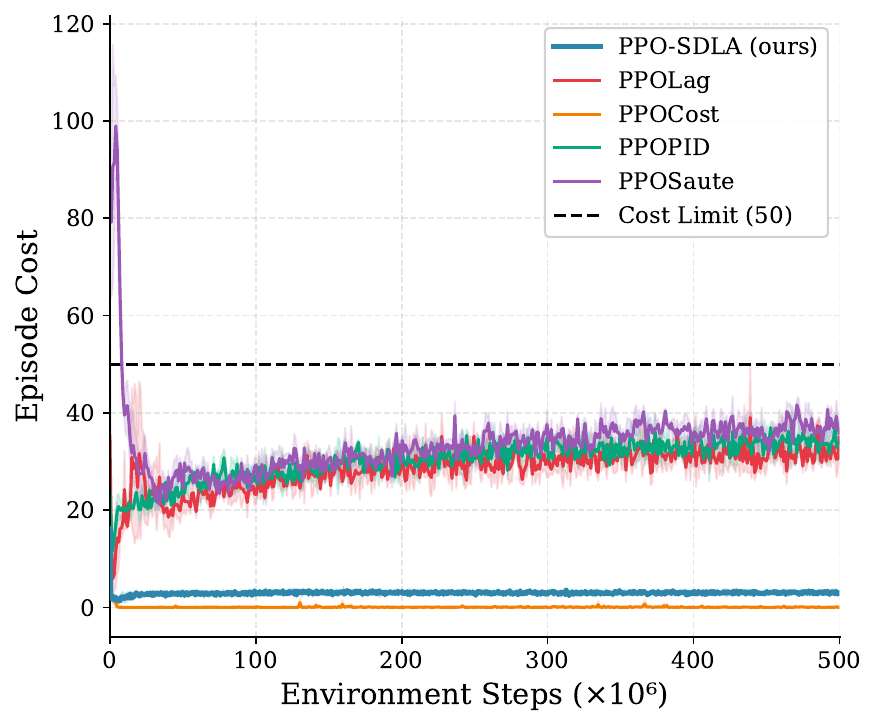}}
\subfloat[Volcanic R]{\includegraphics[width=0.16\textwidth]{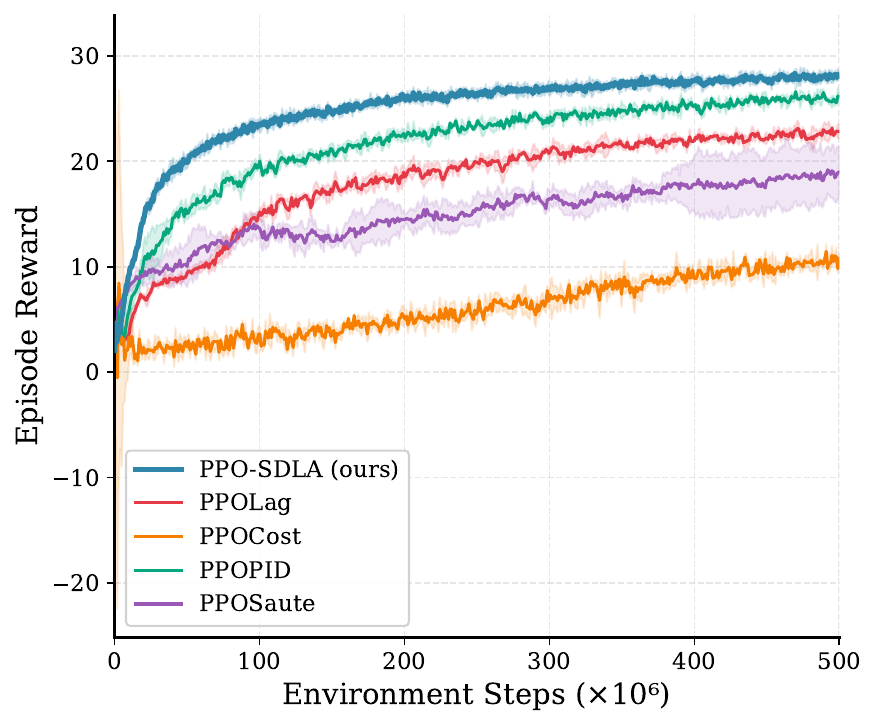}}
\subfloat[Volcanic C]{\includegraphics[width=0.16\textwidth]{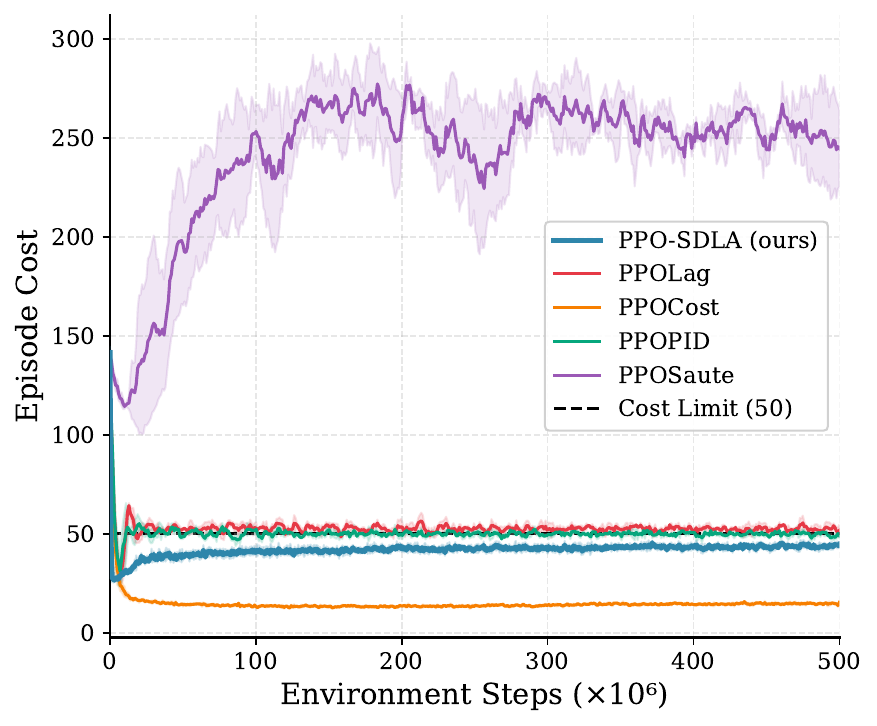}}
\subfloat[Remedy R]{\includegraphics[width=0.16\textwidth]{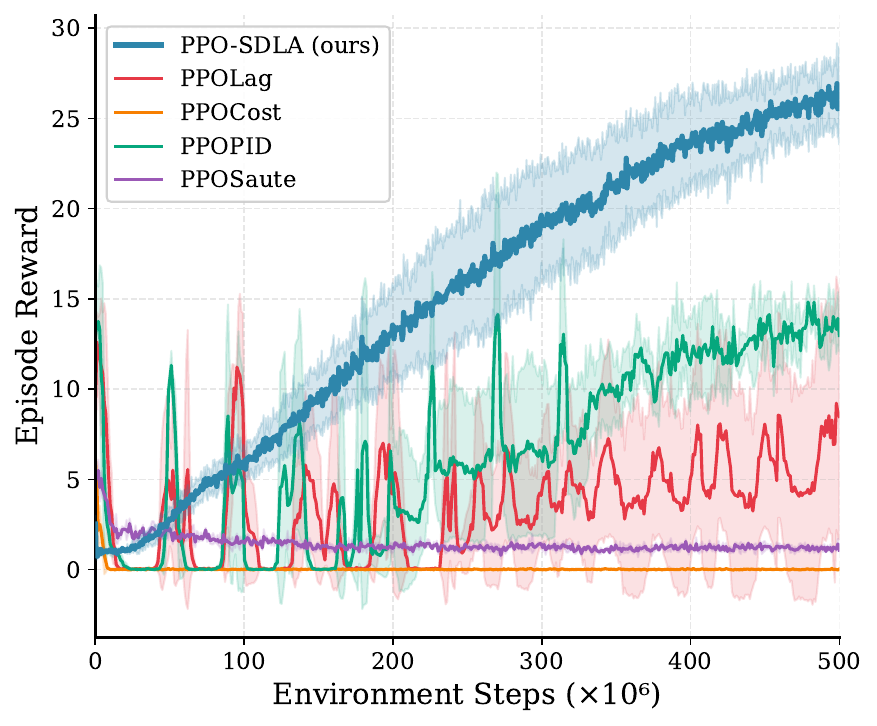}}
\subfloat[Remedy C]{\includegraphics[width=0.16\textwidth]{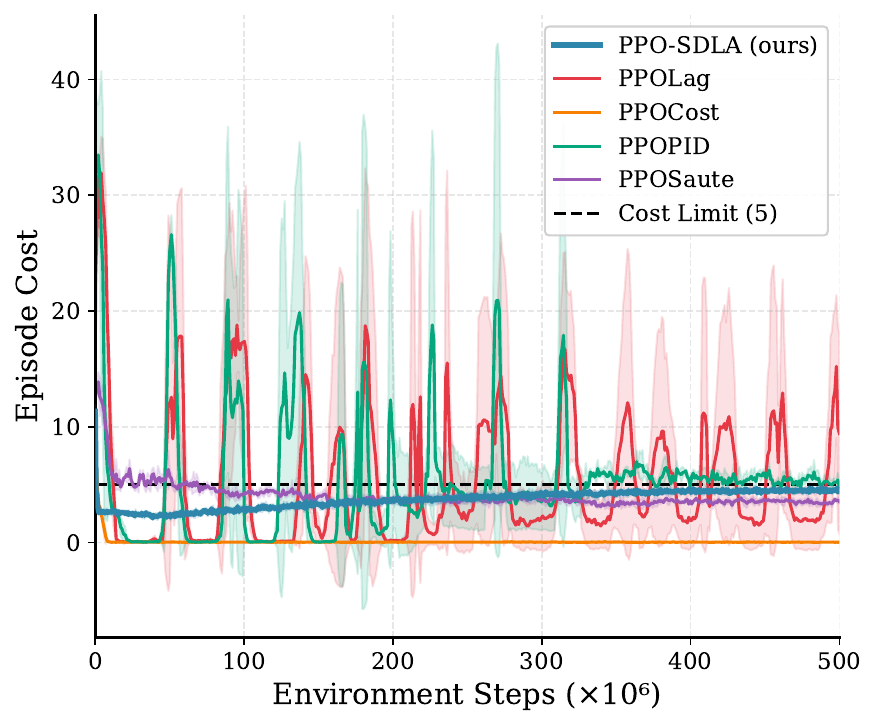}}
\caption{Learning curves on HASARD Level 3 (Part 1): Armament Burden, Volcanic Venture, Remedy Rush.}
\label{fig:hasard_l3_p1}
\end{figure*}

\begin{figure*}[htbp]
\centering
\subfloat[Collateral R]{\includegraphics[width=0.16\textwidth]{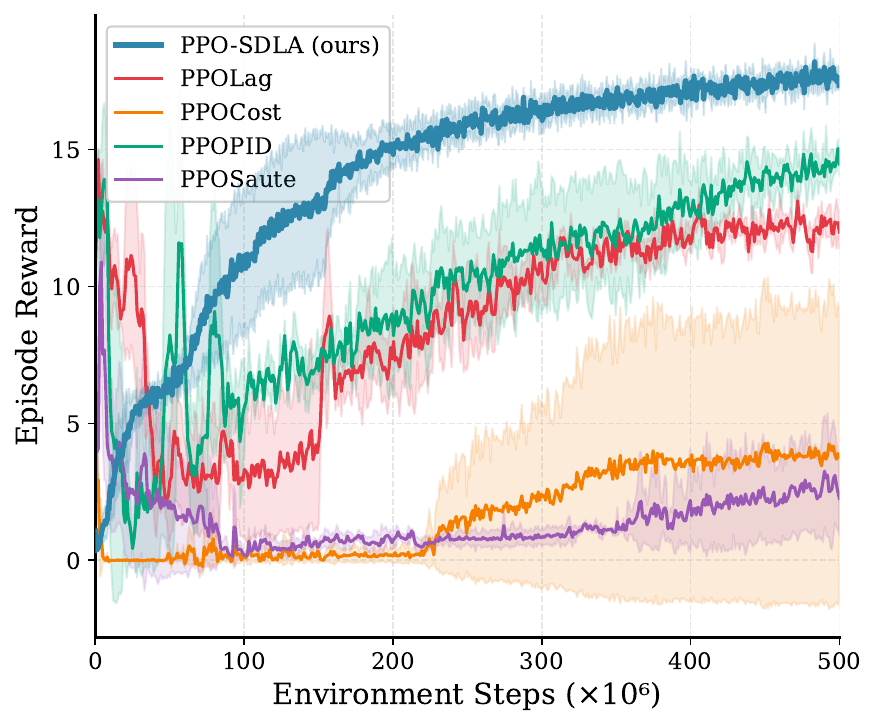}}
\subfloat[Collateral C]{\includegraphics[width=0.16\textwidth]{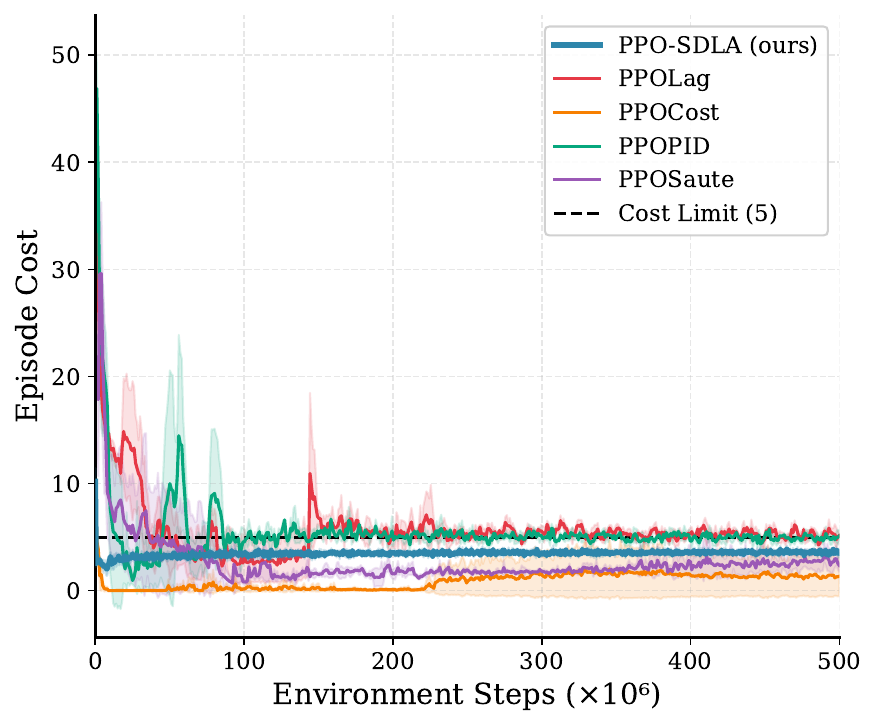}}
\subfloat[Precipice R]{\includegraphics[width=0.16\textwidth]{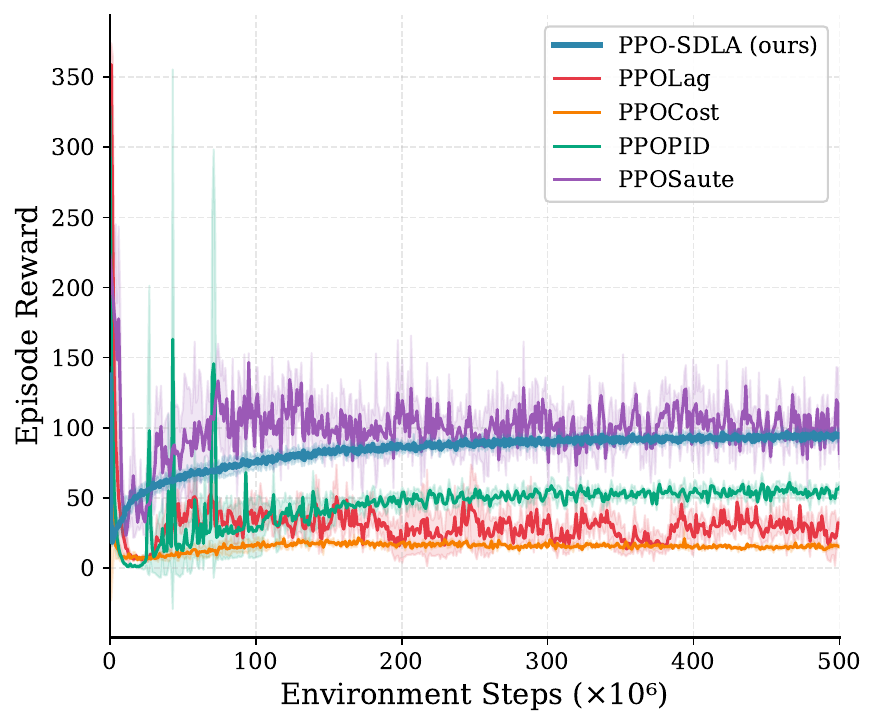}}
\subfloat[Precipice C]{\includegraphics[width=0.16\textwidth]{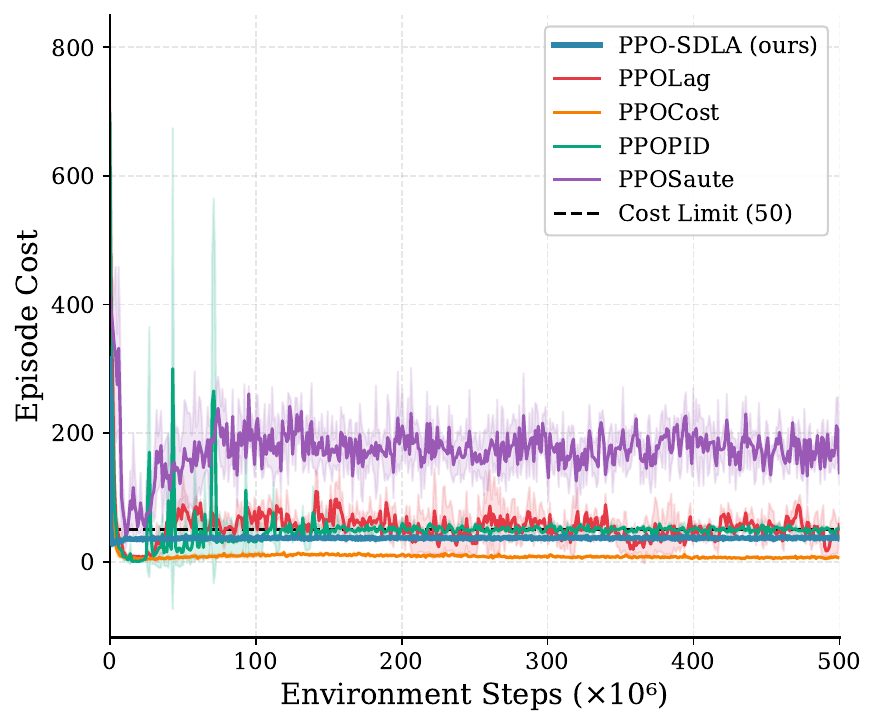}}
\subfloat[Detonator R]{\includegraphics[width=0.16\textwidth]{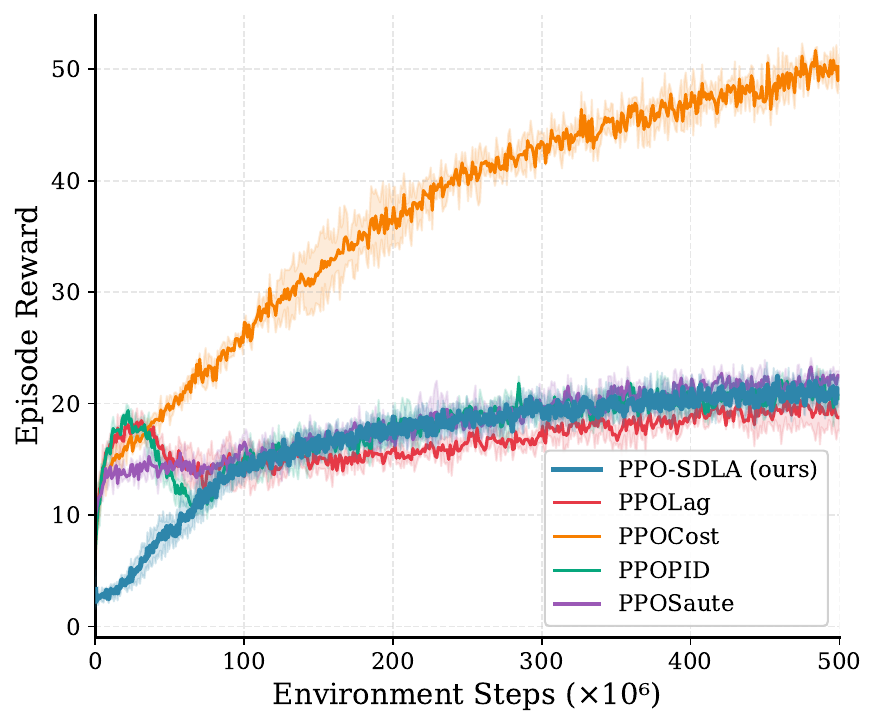}}
\subfloat[Detonator C]{\includegraphics[width=0.16\textwidth]{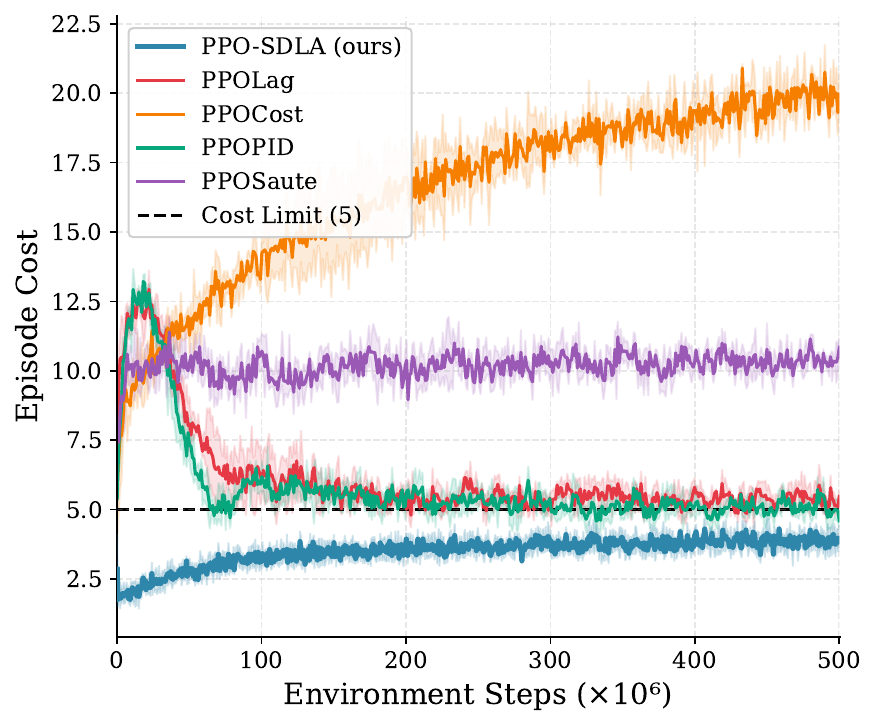}}
\caption{Learning curves on HASARD Level 3 (Part 2): Collateral Damage, Precipice Plunge, Detonator's Dilemma.}
\label{fig:hasard_l3_p2}
\end{figure*}

We evaluate PPO-SDLA on the HASARD benchmark across three difficulty levels (L1--L3) and six scenarios: Armament Burden, Volcanic Venture, Remedy Rush, Collateral Damage, Precipice Plunge, and Detonator's Dilemma. Learning curves are presented in Figures~\ref{fig:hasard_l1_p1}--\ref{fig:hasard_l3_p2}. Lines show mean training episode aggregates and shaded regions show one standard deviation across runs. For PPOCost, the logged cost penalty is added back to the logged reward to recover task reward, as in the tables. The plots compare PPO-SDLA, PPOLag, PPOCost, PPOPID, and PPOSaute; the tables additionally report PPO and P3O.

Table~\ref{tab:human_comparison} compares terminal PPO-SDLA summaries with human averages over 10 episodes.

\begin{sidewaystable*}[htbp]
\centering
\caption{Rewards and costs over five SDLA runs and available archived baseline runs. The cost threshold is 5 for Remedy Rush, Collateral Damage, and Detonator's Dilemma, and 50 for all other environments. \textcolor{green!60!black}{Green}: mean cost meets the budget. \textcolor{green!60!black}{\textbf{Bold Green}}: highest mean reward among methods meeting the mean cost budget.}
\label{tab:baseline_comparison_std}
\scriptsize
\setlength{\tabcolsep}{2pt}
\renewcommand{\arraystretch}{1.05}
\begin{tabular}{@{}clcccccccccccc@{}}
\toprule
\multirow{3}{*}{Level} & \multirow{3}{*}{Method} & \multicolumn{2}{c}{Armament} & \multicolumn{2}{c}{Volcanic} & \multicolumn{2}{c}{Remedy} & \multicolumn{2}{c}{Collateral} & \multicolumn{2}{c}{Precipice} & \multicolumn{2}{c}{Detonator's} \\
& & \multicolumn{2}{c}{Burden} & \multicolumn{2}{c}{Venture} & \multicolumn{2}{c}{Rush} & \multicolumn{2}{c}{Damage} & \multicolumn{2}{c}{Plunge} & \multicolumn{2}{c}{Dilemma} \\
\cmidrule(lr){3-4} \cmidrule(lr){5-6} \cmidrule(lr){7-8} \cmidrule(lr){9-10} \cmidrule(lr){11-12} \cmidrule(lr){13-14}
& & $R \uparrow$ & $C \downarrow$ & $R \uparrow$ & $C \downarrow$ & $R \uparrow$ & $C \downarrow$ & $R \uparrow$ & $C \downarrow$ & $R \uparrow$ & $C \downarrow$ & $R \uparrow$ & $C \downarrow$ \\
\midrule
\multirow{7}{*}{1}
& PPO & $9.68_{\pm 0.09}$ & $109.30_{\pm 2.60}$ & $51.64_{\pm 0.06}$ & $172.93_{\pm 1.13}$ & $50.78_{\pm 0.05}$ & $52.44_{\pm 0.30}$ & $78.61_{\pm 0.31}$ & $41.06_{\pm 0.43}$ & $243.42_{\pm 0.06}$ & $475.62_{\pm 3.73}$ & $29.67_{\pm 1.59}$ & $14.28_{\pm 0.39}$ \\
& PPOCost & \textcolor{green!60!black}{$5.47_{\pm 7.73}$} & \textcolor{green!60!black}{$3.67_{\pm 5.17}$} & \textcolor{green!60!black}{$30.73$} & \textcolor{green!60!black}{$5.44$} & $28.21_{\pm 19.93}$ & $6.75_{\pm 4.75}$ & $68.71_{\pm 0.96}$ & $24.04_{\pm 0.34}$ & \textcolor{green!60!black}{$237.72_{\pm 6.84}$} & \textcolor{green!60!black}{$0.69_{\pm 0.09}$} & $27.60_{\pm 0.49}$ & $8.45_{\pm 0.35}$ \\
& PPOLag & $7.51_{\pm 0.17}$ & $52.41_{\pm 0.60}$ & $42.40_{\pm 0.09}$ & $52.00_{\pm 0.59}$ & $36.37_{\pm 0.42}$ & $5.25_{\pm 0.09}$ & $29.09_{\pm 1.45}$ & $5.61_{\pm 0.52}$ & \textcolor{green!60!black}{$147.24_{\pm 80.13}$} & \textcolor{green!60!black}{$44.96_{\pm 6.61}$} & $21.49_{\pm 1.86}$ & $5.62_{\pm 1.01}$ \\
& PPOSaute & \textcolor{green!60!black}{$2.33_{\pm 3.27}$} & \textcolor{green!60!black}{$32.87_{\pm 10.58}$} & $32.68_{\pm 0.59}$ & $62.00_{\pm 2.16}$ & $20.50_{\pm 2.55}$ & $9.18_{\pm 2.41}$ & $26.08_{\pm 1.25}$ & $7.09_{\pm 0.94}$ & $241.37_{\pm 0.20}$ & $424.35_{\pm 56.71}$ & $28.69_{\pm 0.88}$ & $8.78_{\pm 0.16}$ \\
& PPOPID & \textcolor{green!60!black}{$8.99_{\pm 0.08}$} & \textcolor{green!60!black}{$49.79_{\pm 0.46}$} & $45.23_{\pm 0.11}$ & $50.53_{\pm 0.24}$ & \textcolor{green!60!black}{$38.19_{\pm 0.13}$} & \textcolor{green!60!black}{$4.90_{\pm 0.20}$} & $43.27_{\pm 0.35}$ & $5.03_{\pm 0.16}$ & \textcolor{green!60!black}{$231.53_{\pm 15.26}$} & \textcolor{green!60!black}{$43.91_{\pm 4.89}$} & $26.51_{\pm 0.16}$ & $5.25_{\pm 0.10}$ \\
& P3O & \textcolor{green!60!black}{$8.60_{\pm 0.08}$} & \textcolor{green!60!black}{$40.72_{\pm 0.34}$} & \textcolor{green!60!black}{$43.55_{\pm 0.07}$} & \textcolor{green!60!black}{$46.10_{\pm 0.18}$} & \textcolor{green!60!black}{$37.90_{\pm 0.30}$} & \textcolor{green!60!black}{$4.78_{\pm 0.10}$} & $46.52_{\pm 0.39}$ & $5.97_{\pm 0.21}$ & $242.53_{\pm 0.97}$ & $176.41_{\pm 141.71}$ & $29.62_{\pm 1.12}$ & $6.59_{\pm 0.26}$ \\
& PPO-SDLA (ours) & \textcolor{green!60!black}{$\mathbf{19.46}_{\pm 2.14}$} & \textcolor{green!60!black}{$12.80_{\pm 8.73}$} & \textcolor{green!60!black}{$\mathbf{44.49}_{\pm 0.33}$} & \textcolor{green!60!black}{$29.49_{\pm 0.60}$} & \textcolor{green!60!black}{$\mathbf{39.07}_{\pm 0.65}$} & \textcolor{green!60!black}{$2.71_{\pm 0.13}$} & \textcolor{green!60!black}{$\mathbf{43.96}_{\pm 0.66}$} & \textcolor{green!60!black}{$3.01_{\pm 0.14}$} & \textcolor{green!60!black}{$\mathbf{243.30}_{\pm 0.09}$} & \textcolor{green!60!black}{$43.09_{\pm 9.83}$} & \textcolor{green!60!black}{$\mathbf{27.24}_{\pm 0.60}$} & \textcolor{green!60!black}{$2.97_{\pm 0.13}$} \\
\midrule
\multirow{7}{*}{2}
& PPO & $4.24_{\pm 3.00}$ & $99.59_{\pm 21.83}$ & $38.31_{\pm 0.08}$ & $186.44_{\pm 3.15}$ & $61.94_{\pm 0.09}$ & $62.22_{\pm 0.38}$ & $53.58_{\pm 0.18}$ & $61.13_{\pm 0.23}$ & $324.69_{\pm 0.06}$ & $608.07_{\pm 0.97}$ & $40.09_{\pm 1.23}$ & $19.63_{\pm 0.48}$ \\
& PPOCost & \textcolor{green!60!black}{$7.59_{\pm 5.37}$} & \textcolor{green!60!black}{$6.20_{\pm 4.39}$} & \textcolor{green!60!black}{$21.10_{\pm 0.41}$} & \textcolor{green!60!black}{$3.93_{\pm 0.11}$} & \textcolor{green!60!black}{$0.01_{\pm 0.01}$} & \textcolor{green!60!black}{$0.03_{\pm 0.00}$} & $21.86_{\pm 15.48}$ & $8.19_{\pm 5.79}$ & $162.01_{\pm 17.33}$ & $61.19_{\pm 4.67}$ & $40.37_{\pm 0.51}$ & $15.56_{\pm 0.23}$ \\
& PPOLag & $4.50_{\pm 0.08}$ & $53.50_{\pm 0.77}$ & $26.11_{\pm 0.49}$ & $53.10_{\pm 0.25}$ & $28.75_{\pm 1.66}$ & $5.72_{\pm 0.34}$ & $18.87_{\pm 0.53}$ & $5.61_{\pm 0.27}$ & $105.01_{\pm 12.95}$ & $52.39_{\pm 1.27}$ & $19.57_{\pm 0.21}$ & $5.34_{\pm 0.21}$ \\
& PPOSaute & \textcolor{green!60!black}{$1.55_{\pm 2.19}$} & \textcolor{green!60!black}{$30.47_{\pm 9.73}$} & $23.68_{\pm 0.60}$ & $75.39_{\pm 6.13}$ & \textcolor{green!60!black}{$3.72_{\pm 0.17}$} & \textcolor{green!60!black}{$4.80_{\pm 0.06}$} & \textcolor{green!60!black}{$6.37_{\pm 4.41}$} & \textcolor{green!60!black}{$3.21_{\pm 0.93}$} & $119.42_{\pm 28.09}$ & $159.30_{\pm 25.71}$ & $20.79_{\pm 1.27}$ & $9.97_{\pm 0.31}$ \\
& PPOPID & $5.50_{\pm 0.13}$ & $50.30_{\pm 0.52}$ & $33.52_{\pm 0.23}$ & $50.38_{\pm 0.40}$ & $32.92_{\pm 0.60}$ & $5.15_{\pm 0.13}$ & $27.23_{\pm 0.19}$ & $5.12_{\pm 0.04}$ & $162.16_{\pm 1.72}$ & $50.77_{\pm 0.70}$ & \textcolor{green!60!black}{$20.84_{\pm 0.60}$} & \textcolor{green!60!black}{$4.92_{\pm 0.21}$} \\
& P3O & \textcolor{green!60!black}{$5.33_{\pm 0.12}$} & \textcolor{green!60!black}{$39.82_{\pm 0.77}$} & \textcolor{green!60!black}{$32.38_{\pm 0.11}$} & \textcolor{green!60!black}{$45.89_{\pm 0.47}$} & \textcolor{green!60!black}{$31.91_{\pm 1.32}$} & \textcolor{green!60!black}{$4.85_{\pm 0.05}$} & $27.41_{\pm 0.26}$ & $5.12_{\pm 0.21}$ & $247.05_{\pm 10.78}$ & $178.56_{\pm 12.48}$ & $25.48_{\pm 0.22}$ & $6.97_{\pm 0.21}$ \\
& PPO-SDLA (ours) & \textcolor{green!60!black}{$\mathbf{14.22}_{\pm 0.05}$} & \textcolor{green!60!black}{$19.01_{\pm 0.55}$} & \textcolor{green!60!black}{$\mathbf{33.71}_{\pm 0.20}$} & \textcolor{green!60!black}{$28.01_{\pm 0.60}$} & \textcolor{green!60!black}{$\mathbf{44.02}_{\pm 0.96}$} & \textcolor{green!60!black}{$4.67_{\pm 0.22}$} & \textcolor{green!60!black}{$\mathbf{31.09}_{\pm 0.79}$} & \textcolor{green!60!black}{$4.44_{\pm 0.39}$} & \textcolor{green!60!black}{$\mathbf{172.40}_{\pm 8.57}$} & \textcolor{green!60!black}{$40.98_{\pm 0.84}$} & \textcolor{green!60!black}{$\mathbf{22.72}_{\pm 0.83}$} & \textcolor{green!60!black}{$4.19_{\pm 0.25}$} \\
\midrule
\multirow{7}{*}{3}
& PPO & $1.99_{\pm 1.41}$ & $118.22_{\pm 81.88}$ & $42.20_{\pm 0.12}$ & $347.77_{\pm 1.18}$ & $53.16_{\pm 0.18}$ & $68.27_{\pm 0.25}$ & $34.86_{\pm 0.08}$ & $84.44_{\pm 0.60}$ & $487.17_{\pm 0.09}$ & $894.22_{\pm 3.41}$ & $49.36_{\pm 0.25}$ & $23.72_{\pm 0.27}$ \\
& PPOCost & \textcolor{green!60!black}{$0.04_{\pm 0.02}$} & \textcolor{green!60!black}{$0.05_{\pm 0.01}$} & \textcolor{green!60!black}{$10.61_{\pm 0.16}$} & \textcolor{green!60!black}{$14.76_{\pm 0.59}$} & \textcolor{green!60!black}{$0.01_{\pm 0.00}$} & \textcolor{green!60!black}{$0.02_{\pm 0.00}$} & \textcolor{green!60!black}{$3.93_{\pm 5.56}$} & \textcolor{green!60!black}{$1.35_{\pm 1.89}$} & \textcolor{green!60!black}{$15.39_{\pm 0.68}$} & \textcolor{green!60!black}{$6.64_{\pm 0.66}$} & $49.95_{\pm 0.79}$ & $19.79_{\pm 0.35}$ \\
& PPOLag & \textcolor{green!60!black}{$2.03_{\pm 0.06}$} & \textcolor{green!60!black}{$31.89_{\pm 0.80}$} & $22.77_{\pm 0.19}$ & $52.54_{\pm 1.07}$ & $8.02_{\pm 5.66}$ & $9.74_{\pm 3.27}$ & $12.22_{\pm 0.25}$ & $5.29_{\pm 0.33}$ & \textcolor{green!60!black}{$23.87_{\pm 10.94}$} & \textcolor{green!60!black}{$34.98_{\pm 17.37}$} & $19.27_{\pm 1.43}$ & $5.26_{\pm 0.17}$ \\
& PPOSaute & \textcolor{green!60!black}{$2.32_{\pm 0.07}$} & \textcolor{green!60!black}{$37.68_{\pm 0.13}$} & $18.89_{\pm 2.39}$ & $246.80_{\pm 23.49}$ & \textcolor{green!60!black}{$1.17_{\pm 0.01}$} & \textcolor{green!60!black}{$3.50_{\pm 0.11}$} & \textcolor{green!60!black}{$2.74_{\pm 1.81}$} & \textcolor{green!60!black}{$2.76_{\pm 0.80}$} & $101.64_{\pm 2.71}$ & $176.14_{\pm 7.03}$ & $22.20_{\pm 0.08}$ & $10.36_{\pm 0.10}$ \\
& PPOPID & \textcolor{green!60!black}{$2.78_{\pm 0.04}$} & \textcolor{green!60!black}{$33.89_{\pm 0.59}$} & \textcolor{green!60!black}{$25.80_{\pm 0.25}$} & \textcolor{green!60!black}{$49.02_{\pm 0.70}$} & $13.51_{\pm 1.01}$ & $5.14_{\pm 0.11}$ & \textcolor{green!60!black}{$14.51_{\pm 0.33}$} & \textcolor{green!60!black}{$4.97_{\pm 0.21}$} & \textcolor{green!60!black}{$54.02_{\pm 1.49}$} & \textcolor{green!60!black}{$49.57_{\pm 0.46}$} & \textcolor{green!60!black}{$20.49_{\pm 0.96}$} & \textcolor{green!60!black}{$4.87_{\pm 0.36}$} \\
& P3O & \textcolor{green!60!black}{$2.61_{\pm 0.07}$} & \textcolor{green!60!black}{$29.94_{\pm 0.81}$} & \textcolor{green!60!black}{$24.93_{\pm 0.10}$} & \textcolor{green!60!black}{$49.84_{\pm 0.24}$} & \textcolor{green!60!black}{$14.29_{\pm 0.66}$} & \textcolor{green!60!black}{$4.19_{\pm 0.08}$} & \textcolor{green!60!black}{$13.57_{\pm 0.21}$} & \textcolor{green!60!black}{$4.12_{\pm 0.12}$} & $269.14_{\pm 154.19}$ & $428.72_{\pm 324.48}$ & $26.73_{\pm 0.54}$ & $7.49_{\pm 0.16}$ \\
& PPO-SDLA (ours) & \textcolor{green!60!black}{$\mathbf{2.87}_{\pm 0.13}$} & \textcolor{green!60!black}{$2.85_{\pm 0.31}$} & \textcolor{green!60!black}{$\mathbf{28.22}_{\pm 0.39}$} & \textcolor{green!60!black}{$44.25_{\pm 1.32}$} & \textcolor{green!60!black}{$\mathbf{26.21}_{\pm 2.56}$} & \textcolor{green!60!black}{$4.48_{\pm 0.27}$} & \textcolor{green!60!black}{$\mathbf{17.60}_{\pm 0.52}$} & \textcolor{green!60!black}{$3.52_{\pm 0.36}$} & \textcolor{green!60!black}{$\mathbf{95.02}_{\pm 4.02}$} & \textcolor{green!60!black}{$36.69_{\pm 2.73}$} & \textcolor{green!60!black}{$\mathbf{20.82}_{\pm 1.00}$} & \textcolor{green!60!black}{$3.87_{\pm 0.48}$} \\
\bottomrule
\end{tabular}
\end{sidewaystable*}

\clearpage
\section{Extended Discussion of Related Methods}
\label{app:related-methods}
\subsection{Primal-Dual Approach}

The standard primal-dual approach forms the Lagrangian $\mathcal L(\pi,\lambda)=J^r(\pi)-\lambda(J^c(\pi)-d)$ and solves $\min_{\lambda\ge0}\max_\pi\mathcal L(\pi,\lambda)$ via gradient descent-ascent (GDA)~\citep{ray2019benchmarking,chen2024adaptive,paternain2022safe}:
\[
\pi\leftarrow\pi+\eta_\pi\nabla_\pi\mathcal L(\pi,\lambda),\qquad
\lambda\leftarrow[\lambda+\eta_\lambda(J^c(\pi)-d)]_+.
\]

GDA on minimax problems can exhibit slow progress and sensitivity to the relative primal and dual stepsizes \citep{lin2020gradient}. Oscillation is also commonly observed near the constraint boundary \citep{stooke2020responsivesafe}.

\subsection{Stabilizing Multiplier Updates}
Several methods stabilize $\lambda$ updates within the primal--dual structure. Stooke et al.~\citep{stooke2020responsivesafe} interpret gradient multiplier
updates as integral control and introduce proportional and derivative feedback.
Their continuous-time analysis explains how the proportional term adds damping,
while the derivative term uses cost trends to anticipate constraint violations.
Tuned controller gains set the strength of each feedback term. Gradient and PID methods specify how the
multiplier evolves from cost errors and their history. SDLA derives a
response from a regularized scalar objective on the current trajectory batch,
with explicit activation conditions and response-gain bounds.

Dai et al.~\citep{dai2023augmented} add a quadratic
penalty term $\frac{\sigma}{2}(J^c - d)_+^2$ to the Lagrangian.
CAL~\citep{wu2024off} calibrates the cost critic in off-policy learning.

Recent works have also explored regularization-based approaches for stable 
primal-dual updates in CMDPs. \citep{ding2023last} propose RPG-PD, which 
combines entropy regularization on the policy with quadratic regularization 
on the dual variable, achieving sublinear last-iterate convergence via 
single-time-scale GDA. \citep{montenegro2024last} propose C-PG, a general 
framework with dual-variable regularization that achieves global last-iterate 
convergence under weak gradient domination assumptions. SDLA constructs its regularized dual objective from a frozen trajectory batch and current policy cost information, then computes the response with a scalar inner solve. This dual update retains the PPO and SAC actor update rules.

\subsection{Penalty, State-Augmentation, and Projection Methods}

Penalty-based alternatives include PPOCost~\citep{tomilin2025hasard}, which uses $\tilde r=r-\beta'c$ with a fixed coefficient, and P3O~\citep{zhang2022penalized}. The penalty strength controls the reward--cost tradeoff. P3O establishes exactness for a sufficiently large finite coefficient and uses a sample approximation in its practical surrogate~\citep{zhang2022penalized}.

State augmentation methods transform the CMDP into an unconstrained problem.
SAUTE RL~\citep{sootla2022saute} augments the state with a safety budget $z_t$,
and SIMMER~\citep{sootla2022enhancing} extends this with budget scheduling.
The resulting policy takes budget-conditioned inputs and optimizes a reshaped objective; SIMMER also schedules the budget.

Constrained and projected policy methods modify the actor update directly.
CPO~\citep{achiam2017constrained} and PCPO~\citep{yang2020projection} use local constraint and KL approximations, with second-order computations in their practical updates.
FOCOPS~\citep{zhang2020first} and CUP~\citep{yang2022constrained} implement policy projection with first-order updates.

\section*{Acknowledgments}
OpenAI ChatGPT/Codex assisted with draft polishing. The authors take responsibility for the final manuscript.

\end{document}